\documentclass{article} 
\usepackage{iclr2026_conference}
\iclrfinalcopy
\usepackage{times}
\usepackage{booktabs}
\usepackage{algorithmic}
\usepackage[utf8]{inputenc} 
\usepackage[T1]{fontenc}    
\usepackage{url}            
\usepackage{booktabs}       
\usepackage{amsfonts}       
\usepackage{nicefrac}       
\usepackage{microtype}      
\usepackage{soul}
\usepackage{graphicx}
\usepackage{amsmath}
\usepackage{outlines}
\usepackage{xurl}

\usepackage{amsmath,amsfonts,bm}

\def\eqref#1{equation~\ref{#1}}

\def\1{\bm{1}}

\DeclareMathAlphabet{\mathsfit}{\encodingdefault}{\sfdefault}{m}{sl}
\SetMathAlphabet{\mathsfit}{bold}{\encodingdefault}{\sfdefault}{bx}{n}

\usepackage{multirow}
\usepackage{paralist}

\usepackage{multicol}
\usepackage{diagbox}

\usepackage[ruled,noend]{algorithm2e}

\newtheorem{theorem}{Theorem}
\SetCommentSty{mycommfont}

\usepackage{here}

\usepackage{amsmath,amssymb,amsfonts,amsbsy,amsfonts,latexsym}
\usepackage{makecell}
\usepackage{xcolor}
\usepackage{colortbl}

\usepackage{tabularx,colortbl,xcolor}
\usepackage[normalem]{ulem}
\useunder{\uline}{\ul}{}

\usepackage{enumitem}

\usepackage{xparse}

\SetKwInput{KwInput}{Input}
\SetKwInput{KwRequire}{Require}

\usepackage{longtable}

\NewDocumentCommand{\var}{O{s} m O{}}{%
  \ensuremath{#1_{#2}^{#3}}
}
\usepackage{siunitx}

\newcommand{\commentout}[1]{}

\definecolor{light-gray}{gray}{0.80}

\usepackage{amsthm}

\usepackage{amsthm}

\newtheorem{mytheorem}{Theorem}
\newtheorem{assumption}[mytheorem]{Assumption}
\newtheorem{lemma}[mytheorem]{Lemma}
\newtheorem{remk}[mytheorem]{Remark}

\definecolor{myblue}{rgb}{0,0.2,0.8}
\usepackage{hyperref}
\usepackage{xcolor}
\usepackage{listings}

\definecolor{upforestgreen}{rgb}{0.6, 0.8, 0.2}

\usepackage{chngcntr}
\usepackage{adjustbox}
\usepackage{wrapfig}
\usepackage{arydshln}
\usepackage{tcolorbox}
\usepackage{amsthm,amsmath,amssymb}
\usepackage{amsfonts,dsfont}
\usepackage{mathtools}

\newcommand*\samethanks[1][\value{footnote}]{\footnotemark[#1]}

\definecolor{mygreen}{HTML}{009901}
\definecolor{myred}{HTML}{A52A2A}

\author{
\textbf{Zhongzhu Zhou}$^{1,3}$, 
\textbf{Fengxiang Bie}$^{1}$, 
\textbf{Ziyan Chen}$^{1}$, 
\textbf{Zhenyu Zhang}$^{4}$, 
\textbf{Yibo Yang}$^{2}$, \\
\textbf{Junxiong Wang}$^{3}$,
\textbf{Ben Athiwaratkun}$^{3}$, 
\textbf{Xiaoxia Wu}$^{3}$\thanks{Equal advising.}\thanks{Corresponding author: \texttt{shirely@together.ai}, Code is available at \url{https://github.com/FutureMLS-Lab/CARE}.}, 
\textbf{Shuaiwen Leon Song}$^{1,3}$\samethanks[1] \\
$^{1}$ University of Sydney,
$^{2}$ King Abdullah University of Science and Technology, \\
$^{3}$ Together AI,
$^{4}$ University of Texas at Austin
}

\begin{document}

\title{CARE: Covariance-Aware and Rank-Enhanced Decomposition for Enabling Multi-Head Latent Attention}

\maketitle

\begin{abstract}

\noindent Converting pretrained attention modules such as \emph{grouped-query attention} (GQA) into \emph{multi-head latent attention} (MLA) can improve expressivity without increasing KV-cache cost, making it attractive for efficient inference. However, many practical conversion baselines rely on weight-only low-rank approximations (e.g., SVD-style initializations) and uniform rank allocation. They focus on minimizing the difference between weight matrices rather than on how those weights affect input activations, ignore the covariance structure of activations, and enforce uniform rank across layers—causing activation drift and degraded attention fidelity. To address these issues, we propose CARE, a \textbf{\textit{C}ovariance-\textit{A}ware, \textit{R}ank-\textit{E}nhanced} MLA conversion pipeline under a fixed KV width. CARE introduces three key steps: (i) \emph{activation-preserving factorization}, which aligns the approximation with the actual input activations rather than just the weights; (ii) \emph{adjusted-rank allocation}, which spreads a fixed KV budget across layers by giving more capacity to layers that need it most; and (iii) \emph{KV-parity mapping}, which reparameterizes the converted $K$ and $V$ to fit the MLA format while keeping the KV-cache size unchanged.
Our method outperforms a uniform-rank SVD baseline on Qwen3-4B/30B-A3B-Instruct-2507 and Llama-3.1-8B/70B-Instruct, reducing one-shot perplexity by up to $215\times$ and improving mean accuracy by up to $1.70\times$ at matched KV budgets.
With a brief post-SVD “healing” fine-tune, we fully recover the original model's accuracy.

\end{abstract}

\section{Introduction}~\label{sec:intro}
\noindent Large Language Models (LLMs) deliver impressive capabilities but at high inference cost, with the key–value (KV) cache in self-attention emerging as a primary memory and bandwidth bottleneck~\citep{vaswani2017attention,kwon2023vllm}. In the standard multi-head attention (MHA) formulation, each head materializes and caches its own keys and values at every decoding step, causing the KV footprint to grow linearly with sequence length and head count. To alleviate this, architecture variants such as multi-query attention (MQA), which shares a single $K,V$ across all heads, and grouped-query attention (GQA), which shares $K,V$ within head groups, have been adopted at scale to shrink KV cache size
~\citep{shazeer2019mqa,ainslie2023gqa,touvron2023llama2,jiang2023mistral,chowdhery2022palm,shoeybi2019megatron}. While effective, these variants reduce the number of distinct key/value projections, which can limit attention expressivity and introduce quality regressions when compression is pushed aggressively.

A more recent line of work reframes the KV-cache problem as one of \emph{learned low-rank representation}~\citep{wang2020linformer,xiong2021nystromformer}. Multi-Head Latent Attention (MLA) compresses keys and values into low-dimensional \emph{latent} vectors, caches only these latents, and restores expressivity with lightweight up-down projections at compute time~\citep{deepseekv2}. In practice, MLA can dramatically reduce KV size while preserving or even improving task accuracy by trading memory and communication for modest extra floating-point-operations (FLOPs) in the projections~\citep{deepseekv2,guo2025deepseek, liu2024deepseek, geens2025hardwaremla}. Despite these advantages, the ecosystem is dominated by pretrained MHA/GQA checkpoints~\citep{touvron2023llama2,jiang2023mistral,yang2024qwen2}. Retraining large models from scratch under MLA is expensive, so a natural question arises: 

\begin{center}
   Can we \emph{convert} strong, pretrained MHA/GQA models into MLA \emph{post-hoc}, without increasing the KV budget and without incurring large performance loss?
\end{center}

Recent work has explored converting traditional attention (MHA/GQA) into multi-head latent attention (MLA) under fixed KV width. \emph{TransMLA}~\citep{meng2025transmla} demonstrates that every GQA layer admits an equivalent MLA parameterization and proposes a practical post-training mapping followed by light finetuning. \emph{MHA2MLA}~\citep{ji2025mha2mla} generalizes to MHA$\to$MLA by addressing positional encoding mismatches (e.g., partial RoPE adjustments) and initializing $W_K,W_V$ with low-rank joint SVD before efficient recovery~\citep{su2021roformer}. \emph{X-EcoMLA}~\citep{li2025xecoMLA} constructs MLA from pretrained MHA via structured SVD decomposition and cross-layer distillation–based parameter recycling for efficient latent projection initialization. 
In parallel, general-purpose SVD-based compression and approximation methods (e.g., SVD-LLM V2~\citep{wang2024svd,wang2025svdllmv2} and activation-aware SVD (ASVD)~\citep{yuan2023asvd}) improve truncation and orientation beyond naïve weight SVD, while cache-centric baselines such as Palu~\citep{chang2024palu} reduce KV memory via low-rank KV-cache projection. Together, these works establish MLA as a promising post-hoc target and highlight \emph{low-rank factorization} as central to preserving pretrained knowledge under KV-constrained reparameterizations~\citep{hu2022lora,denil2013predicting,denton2014exploiting,sainath2013lowrank,eckart1936approximation}.

However, direct SVD initialization has two key shortcomings. 
First, it minimizes error in weight space ($\|W-\hat W\|$) rather than activation space ($\|XW - X\hat W\|$), ignoring how the projection actually operates during decoding~\citep{hassibi1993optimal,wang2024svd,yuan2023asvd}. This mismatch induces attention-logit drift even when the weight approximation is accurate. 
Second, it enforces a uniform rank across layers, neglecting differences in spectral structure. Layers with fast spectral decay are over-compressed, while those with slower decay are under-compressed, leading to fidelity loss and heavier reliance on post-conversion finetuning.
\begin{figure}[t]
\centering
\includegraphics[width=\linewidth]{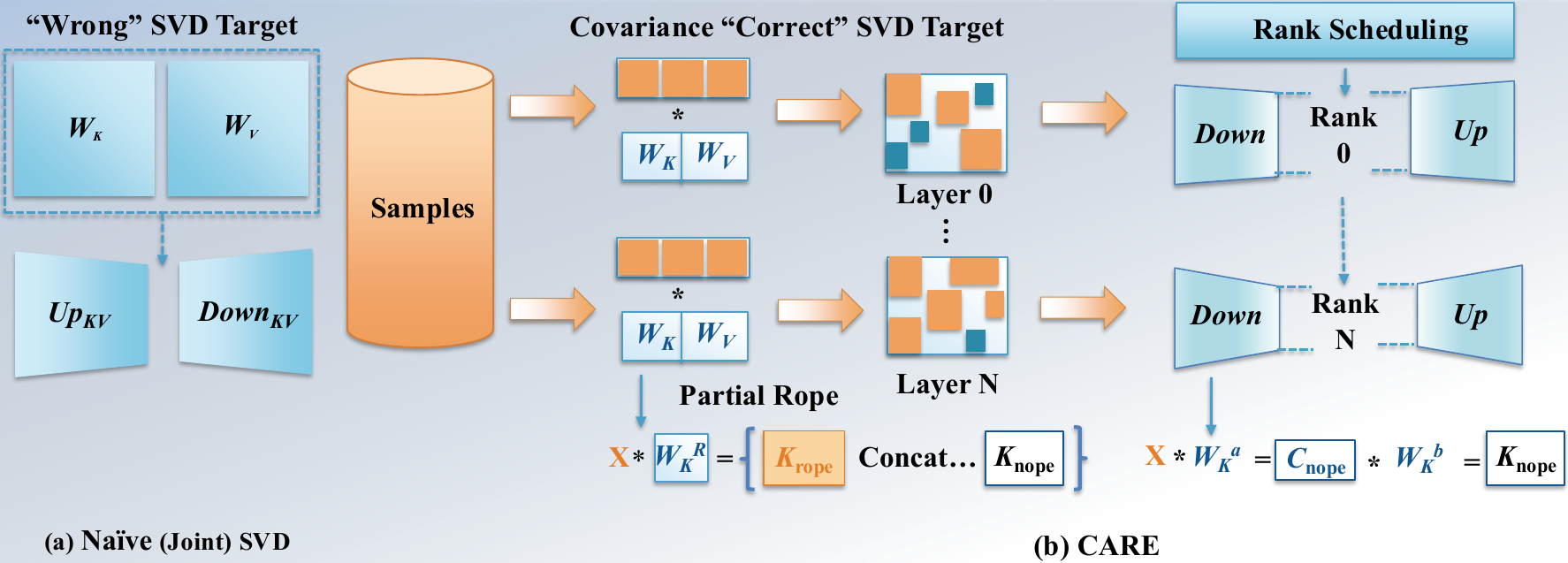}
\vspace{-0.5em}
\caption{\textbf{(a) Naive MLA transfer}: jointly factorize $W_K^{(g)}$ and $W_V^{(g)}$ by SVD and truncate to a uniform per-layer rank, optimizing $\|W-\hat W\|_F$ while ignoring layerwise heterogeneity. \textbf{(b) CARE}: estimate activation covariance $C$, factorize $\sqrt{C}W$, unwhiten via $\sqrt{C}^{-1}$ to initialize MLA factors, and use the singular spectrum of $\sqrt{C}W$ for global dynamicrank scheduling under KV parity. This preserves activation geometry and yields a stronger one-shot initialization with less healing.}
\label{fig:care-pipeline}\vspace{-0.5em}
\end{figure}

To address the above two shortcomings, we propose \textsc{Care}, a \textbf{C}ovariance-\textbf{A}ware, \textbf{R}ank-\textbf{E}nhanced conversion pipeline, as shown in Fig.~\ref{fig:care-pipeline}. 
First, \textsc{Care} makes the decomposition \emph{activation-aware}: rather than applying vanilla SVD to $W$, 
we solve a whitened approximation problem by applying SVD to $\sqrt{C}W$ and then unwhitening 
to obtain $\hat W$ \footnote{Here `whiten' means to use $\sqrt{C}W$ instead of $W$ as 
our target to compress, and `unwhiten' means the inverse map which recovers the compressed $W$.}, 
where $C$ summarizes input activation covariance estimated from a modest calibration set. 
This ensures that dominant activation directions are preserved and substantially reduces attention-logit error before any finetuning. 
Second, \textsc{Care} is \emph{rank-adaptive}: it distributes a fixed KV budget across layers and heads based on their singular spectra, allocating higher rank to spectrally complex matrices and lower rank to intrinsically low-rank ones, akin in spirit to budgeted, importance-aware adapter methods \citep{zhang2023adalora,valipour2023dylora,hu2022lora, wang2025dobi}.
This budgeted, importance-aware scheduling maintains fidelity under the KV constraint while reducing reliance on post-conversion finetuning.

\paragraph{Contributions.}
\begin{itemize}
\item \textbf{Activation-aware Initialization.} We propose a covariance-aware factorization that minimizes activation error $\|XW - X\hat W\|$ (rather than weight error), implemented via SVD on a whitened operator and subsequent unwhitening. This preserves attention logits more faithfully at equal KV budget to initialize MLA weights.
\item \textbf{Rank-adaptive scheduling under fixed KV width.} We introduce a singular-value-guided allocation that distributes rank unevenly across layers/heads and the $\{K,V\}$ matrices, matching spectral difficulty and improving zero-shot fidelity compared with uniform ranks.
\item \textbf{KV-parity mapping and practical pipeline.} We derive a KV-parity reparameterization for MLA conversion and integrate the above techniques into a practical conversion pipeline
CARE-converted models exhibit lower activation error and improved task quality over naive (joint) SVD baselines at equal KV cost, while requiring less data to recover residual gaps.
\end{itemize}

\section{Naïve (Joint) SVD is not enough for MLA Transfer}

Multi-head latent attention (MLA) transfer is often initialized with singular value decomposition (SVD), either per matrix or via a joint factorization across related matrices (e.g., $W_K$, $W_V$). While convenient, this practice implicitly optimizes weight-space error $\lVert W-\hat W\rVert_F$ and assumes that the spectrum alone reveals task importance. In this section we show both assumptions break down in practice and, consequently, naïve (joint) SVD is an unreliable recipe for high-fidelity MLA transfer under a fixed KV budget.

\begin{figure}[h]\centering
\includegraphics[width=\linewidth]{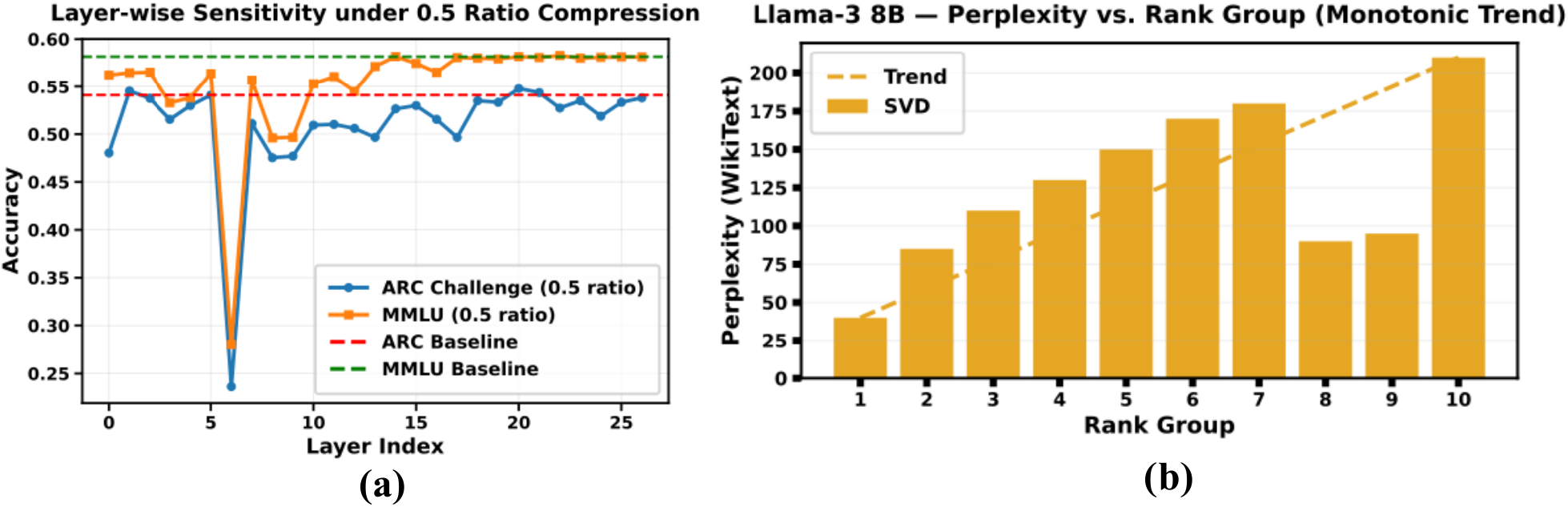}
\vspace{-0.5em}
\caption{(a) Accuracy under 50\% rank reduction applied one layer at a time in DeepSeek-V2-Lite, measured on ARC Challenge and MMLU. Sensitivity is strongly layer-dependent. (b) WikiText perplexity under grouped truncation of GQA attention in Llama-3-8B (layers 30--32). Singular-value groups are ordered by magnitude; the resulting non-monotone degradation shows that singular values alone are an imperfect proxy for MLA conversion quality.}
\label{fig:mot}\vspace{-0.5em}
\end{figure}

\paragraph{Observation 1: Accuracy-preserving rank is \emph{not} uniform across layers.}
Fig.~\ref{fig:mot} (a) exhibits pronounced layer-wise heterogeneity when we halve the rank of every layer: some layers tolerate aggressive reduction with negligible loss, whereas others incur sharp drops on ARC and MMLU.
A one-size-fits-all policy (uniform pruning or a fixed ratio per layer) either over-compresses fragile layers—degrading task performance—or under-compresses robust layers—wasting KV budget. 

\paragraph{Observation 2: Singular values are poor proxies for accuracy importance.}
A common heuristic treats singular values as importance scores, expecting that truncating smaller values should least affect accuracy. We directly test this with a brute-force ablation: given $W=U\Sigma V^\top$, we set the $i$-th singular value to zero and reconstruct $\bar W^{(i)} = U\,\mathrm{diag}(\sigma_1,\ldots,0,\ldots,\sigma_r)\,V^\top$, treating $V^T$ as compress of MLA and $U$ as expand of MLA. As shown in Fig.~\ref{fig:mot} (b), the link between singular-value magnitude and downstream accuracy is non-monotonic. We conjecture the root cause is mismatch of objectives and statistics: vanilla SVD minimizes weight error, not activation-space error $\lVert XW-X\hat W\rVert$ under the true (anisotropic) input distribution.

Thus, naïve (joint) SVD isn’t enough: (i) the rank needed to keep accuracy varies by layer, and (ii) singular values alone don’t reflect accuracy importance of initialization.

\section{CARE: Covariance-Aware and Rank-Enhanced MLA Conversion} \label{sec:method}

We propose \textbf{CARE}, a post-hoc conversion pipeline that maps a pretrained MHA or GQA layer to an MLA layer \emph{at the same KV budget}, while explicitly minimizing \emph{activation} error and \emph{adapting ranks} to spectral difficulty. CARE revisits low-rank factorization through the lens of activation statistics and integrates covariance-weighted SVD and non-uniform rank allocation into a single, practical procedure compatible with multi-models' backbones.

\paragraph{Notation.}
Given a fixed layer with input activations $X\!\in\!\mathbb{R}^{T\times D}$  with length $T$ and embedding dimension $D$. The multi-head attention in our setup contains $n_h$ heads of size $d_h$, where we assume the output space corresponds with input space, formally $n_h d_h = D$. We denote $g_h$ to be the number of GQA groups, where for this layer:
\(
Q = X W_Q,\;
K = X W_K^{(g)}\!\in\!\mathbb{R}^{T\times (g_h d_h)},\;
V = XW_V^{(g)}\!\in\!\mathbb{R}^{T\times (g_h d_h)}
\)
with $g_h < n_h$, with $W_{\{\cdot\}}^{(g)}$ represents the weight matrix under the GQA setup.
By contrast, an MLA layer with latent rank $r$ uses $K=(XW_K^a)W_K^b$, $V=(XW_V^a)W_V^b$, where $W_{\{\cdot\}}^a\!\in\!\mathbb{R}^{D\times r}$, $W_{\{\cdot\}}^b\!\in\!\mathbb{R}^{r\times (n_h d_h)}$. Only the latent $XW_{\{\cdot\}}^a \in \mathbb{R}^{T\times r}$ is cached, with $W_{\{\cdot\}}^b\!\in\!\mathbb{R}^{r\times (n_h d_h)}$ used to recover KV matrix.

\subsection{Preliminary: Covariance for input activations}
\label{subsec:cov}
Let $X_b^{(l)}\!\in\!\mathbb{R}^{T_b\times D}$ be the $b^{th}$ batch of the domain activations with length $T_b$ at some layer $l$ (note that the length of tokens remains the same across layers, but can be different across batches). These batches of domain activations are used to calculate the extent of preserved rank in Sec.~\ref{subsec:ranksched} and initialize the trainable parameters later in Sec.~\ref{subsec:rethinking}.

We then define $C^{(l)}$, the covariance matrix over all the $N$ batches at layer $l$, as follows:
\[
\label{eq: covariance}
C^{(l)} = \tfrac{1}{N}\sum_{b = 1}^N (X_b^{(l)})^\top X_b^{(l)}.
\]
Note that even if we denote the above $C^{(l)}$ to be covariance matrix, it is a slight different from the definition of actual covariance: we do not apply centralize and normalize operation here (see App.~\ref{covariance def} for formal definition).

\subsection{Adjusted-Rank Scheduling Across Layers}
\label{subsec:ranksched}
 Due to heterogeneous key/value spectra across different layers, the retained rank of each layer using MLA are supposed to be different. Let $\mathcal{W}=\{\widetilde{W}_K^{(l)},\widetilde{W}_V^{(l)}\}_{l=1}^{L}$ represent the pretrained KV weights from Sec.~\ref{subsec:kvparity}, where $L$ represents the number of layers. Note that $\mathcal{W}$ contains $2L$ weight matrices, each with dimension $D \times n_hd_h$ (recall $D=n_hd_h $). 

Given a total key-rank budget $R_{\text{tot}}^{(K)}$ (same for value), we score the next rank by its \emph{normalized residual reduction}. Let
\[
\rho_{K,m}^{(l)} \;=\; \frac{\bigl(\sigma_{K,m}^{(l)}\bigr)^2}{\sum_{i=1}^{R_K^{(l)}} \bigl(\sigma_{K,i}^{(l)}\bigr)^2},
\]
where $\sigma_{K,m}^{(l)}$ (same for $V$) is the $m^{th}$-largest singular value of $\sqrt{C^{(l)}}\widetilde{W}_K^{(l)}$, $C^{(l)}$ is the covariance matrix defined in Sec.~\ref{eq: covariance}, and $R_K^{(l)}$ is the rank of $\sqrt{C^{(l)}}\widetilde{W}_K^{(l)}$. For a layer currently assigned rank $r_K^{(l)}$, we define the priority of allocating one more rank as
\begin{equation}
s_K^{(l)}\!\left(r_K^{(l)}\right)
\;=\;
\frac{\rho_{K,r_K^{(l)}+1}^{(l)}}{1-\sum_{m=1}^{r_K^{(l)}} \rho_{K,m}^{(l)}}
\;=\;
\frac{\bigl(\sigma_{K,r_K^{(l)}+1}^{(l)}\bigr)^2}{\sum_{m=r_K^{(l)}+1}^{R_K^{(l)}} \bigl(\sigma_{K,m}^{(l)}\bigr)^2}.
\label{eq:waterfill}
\end{equation}
Appendix~\ref{thm_proof} proves that the Frobenius residual after rank-$r$ truncation equals the squared tail energy; normalizing by this residual makes layers with different spectral scales comparable. We allocate the budget $R_{\text{tot}}^{(K)}$ by greedy water-filling: starting from $r_K^{(l)}\!=\! C$ for all $l$ where $C$ is some constant, we repeatedly assign one rank to $l^{\star}\!=\!\arg\max_l s_K^{(l)}(r_K^{(l)})$ until the budget is exhausted; $V$ is handled identically with its own budget.

\subsection{Rethinking SVD with Activation Covariance}
\label{subsec:rethinking}
A naive rank lowering attempts to minimize Frobenius error $\|W-\widehat{W}\|_F$, where $\widehat{W}$ represents the de-ranked matrix for compressing, and the original pretrained weight matrix to compress, $W := W_{\{\cdot\}}^{(l)}\in \mathcal{W}$, lies in layer $l$ . For inference fidelity, we propose the relevant objective to be minimizing the empirical \emph{activation error}. Focusing on compressing the certain weight matrix $W$, we denote $r$ to be the target compressed rank calculated in Sec.~\ref{subsec:ranksched}. Here $\{X_b := X_b^{(l)}\in \mathbb{R}^{T_b \times D}\}_{b=1}^N$ represents the small domain activation batches to compute the low-rank decomposition at layer $l$, where $N$ is the total number of batches, $T_b$ is the length of each batch and $C:=C^{(l)}$ is the covariance matrix defined in Eq.~\ref{eq: covariance}.

Formally, we try to minimize the following:
\begin{equation}
\min_{\mathrm{rank}(\widehat{W}) \le r} \frac{1}{N}\sum_{b = 1}^N\|X_bW - X_b\widehat{W}\|_F^2.
\label{eq:actobj}
\end{equation}

This optimization formalizes how well $\widehat{W}$ preserves $X_bW$ on relevant inputs $X_b$. From the definition of $C$, one can show that $\frac{1}{N}\sum_{b = 1}^N\|X_bW - X_b\widehat{W}\|_F^2 = \|\,\sqrt{C}(W-\widehat{W})\,\|_F^2$; 
see App.~\ref{derivation} for the proof.

For $K$ and $V$ separately, we compute
\begin{equation}
\sqrt{C}\,W\;=\; U\,\Sigma\,V^\top,\text{then }
\widehat{W} \;=\; \sqrt{C}^{-1}\,U_r\,\Sigma_r\,V_r^\top,
\label{eq:care-svd}
\end{equation}
with $U_r,\Sigma_r,V_r$ the top-$r$ components of $U,\Sigma,V$. In practice, we use a shrinkage $\sqrt{C}_\lambda=(1-\alpha)\sqrt{C}+\alpha\lambda I$ to ensure $C$ is invertible, with $\alpha \in (0,1)$ and $\lambda>0$.

Then we initialize the trainable parameters $W^{a}\!\in\!\mathbb{R}^{D\times r}$ and $W^{b}\!\in\!\mathbb{R}^{r\times (n_h d_h)}$ s.t.\ $W^{a}W^{b}$ equals the compressed matrix. We map SVD factors to MLA by
\begin{equation}
W^{a} \;\leftarrow\; \sqrt{C}^{-1}\,U_r\,\Sigma_r,\qquad
W^{b} \;\leftarrow\; V_r^\top,
\label{eq:splitab}
\end{equation}
so that $W^{a}W^{b}=\widehat{W}$. The cached latent $XW^{a} \in \mathbb{R}^{T \times r}$ in MLA spans the principal activation subspace, where $X$ is the actual input at layer $l$.

\subsection{100\% MLA Conversion by Healing}

Based on our initialization of down-and-up matrices $W^{a}$, $W^{b}$ and number of $T$ tokens, we attempt to encode positional information in the MLA-like attention mechanism. Given layer $l$, let $Q_t=X_t W_Q$ be the usual query at step $t$, and let $K_{C,t}=(X_t W^{a}_K) W^{b}_K$ and $V_{C,t}=(X_t W^{a}_V) W^{b}_V$ be the MLA generated keys/values. Following the decoupled RoPE design in \citep{deepseekv2}, we bring a small RoPE channel of width $d_r$ by introducing \emph{new trainable} matrices~\footnote{TransMLA~\cite{meng2025transmla} and X-EcoMLA~\cite{li2025xecoMLA} both describe how to add such a RoPE channel.}:
\[
W^{R}_Q \in \mathbb{R}^{D\times (n_h d_r)}, \qquad
W^{R}_K \in \mathbb{R}^{D\times d_r},
\]
where $W^{R}_K$ maps \emph{directly} from the activation $X_t$ to a shared RoPE key of width $d_r$ (DeepSeek-like decoupled RoPE). Let $\mathcal{R}_t\in\mathbb{R}^{(n_h d_r)\times(n_h d_r)}$ denote the standard block-diagonal RoPE rotation at step $t$ (one $d_r\times d_r$ rotation per head), and let $R_t\in\mathbb{R}^{d_r\times d_r}$ denote the per-head rotation applied to the shared RoPE key. We compute:
\[
Q_{R,t} \;=\; (X_t W^{R}_Q)\,\mathcal{R}_t,
\qquad
K_{R,t} \;=\; \mathrm{repeat}\!\big(X_t\,W^{R}_K\,R_t\big),
\]
where \texttt{repeat} replicates the shared RoPE key across heads, yielding $K_{R,t}\in\mathbb{R}^{1\times (n_h d_r)}$. We then concatenate channels and compute attention:
\[
Q_t^{\star}=\big[\,Q_t\;;\;Q_{R,t}\,\big], \quad
K_t^{\star}=\big[\,K_{C,t}\;;\;K_{R,t}\,\big], \quad
A_t=\mathrm{Softmax}\!\left(\frac{Q_t^{\star}(K_t^{\star})^\top}{\sqrt{d_h+d_r}}\right),\quad
O_t=A_t\,V_{C,t},
\]
where $A_t$ denotes attention weights and $O_t$ denotes the layer output.
\paragraph{Caching and a compact ``join'' form.}
To preserve KV-cache efficiency, we cache only the KV-side latents $X_t W^{a}_K$ and $X_t W^{a}_V$, along with the \emph{small} shared RoPE key $(X_t W^{R}_K)R_t$ (repeated across heads when forming $K_{R,t}$). Queries (including $Q_{R,t}$) are computed on-the-fly.
Equivalently, the MLA-generated KV can be written in a compact joined form:
\begin{equation}
\label{eq:join_kv}
\operatorname{Concat}\!\left(K_{C,t},\,V_{C,t}\right)
\;=\;
\operatorname{Concat}\!\left(X_t W^{a}_K,\,X_t W^{a}_V\right)\,W_{\mathrm{join}},
\end{equation}
where \(\operatorname{Concat}(\cdot,\cdot)\) concatenates along the last (feature) dimension and \(W_{\mathrm{join}}=\mathrm{blkdiag}(W^{b}_K,\,W^{b}_V)\in\mathbb{R}^{2r\times 2D}\) jointly represents the two bilinear maps \(W^{a}_K W^{b}_K\) and \(W^{a}_V W^{b}_V\).

We penalize the low-rank decomposition by its cross-entropy classification error and KL-divergence imitation error, namely the loss functions:
\begin{align}
\mathcal{L}_{\mathrm{CE}}
&=\;-\tfrac{1}{T}\sum_{t=1}^{T}\log p^{\mathbb{S}}(x_{t+1}\mid x_{\le t}),
\label{eq:ce_app}\\
\mathcal{L}_{\mathrm{KD}}
&=\;\tfrac{1}{T}\sum_{t=1}^{T} 
\mathrm{KL}\!\left(
\mathrm{softmax}(z_{t}^{\mathbb{T}}/\tau)\,\big\|\,\mathrm{softmax}(z_{t}^{\mathbb{S}}/\tau)
\right),
\label{eq:kd_app}\\
\mathcal{L}
&=\;
\mathcal{L}_{\mathrm{CE}}
+\beta\,\tau^{2}\,\mathcal{L}_{\mathrm{KD}}.
\label{eq:total_app}
\end{align}
Here $p^{\mathbb{S}}(x_{t+1}\mid x_{\le t})$ is the student next-token probability under the converted MLA layer (with RoPE adapters $W^{R}_Q,W^{R}_K$ and MLA factors $W^a,W^b$), while $z_{t}^{\mathbb{T}}$ and $z_{t}^{\mathbb{S}}$ are teacher and student logits. We use $p^{\mathbb{S}}(x_{t+1}\mid x_{\le t}) = \mathrm{softmax}(\tfrac{z_{t}^{\mathbb{S}}}{\tau})_{x_{t+1}}$ with temperature $\tau$ and weight $\beta$~\citep{hinton2015distilling}.

\section{Experimental Results} \label{sec:results}

We evaluate whether \textbf{CARE} improves MLA migration under a fixed KV-cache budget (\emph{KV-parity}). We report one-shot and post-healing perplexity \& accuracy, ablation study, long-context retrieval (NiH), and system efficiency.

\noindent\textbf{Setup.} We match cached KV state per token under the same global budget (Tab.~\ref{tab:1}). CARE estimates $C=\mathrm{Cov}[X]$ on a small calibration set and factors $\sqrt{C}W$ with shrinkage $\sqrt{C}\leftarrow(1-\lambda)\sqrt{C}+\lambda I$; We use LM Harness~\citep{eval-harness} with a matched heal budget; full \emph{100\% MLA restoration} (TransMLA/MHA2MLA) is evaluated separately in Sec.~\ref{subsec:data}.
\paragraph{Original, baselines, CARE variants, and datasets.}
\label{subsec:baselines}
\textbf{GQA (source).} Unmodified grouped-query attention.
We compare against KV-compression baselines:
\textbf{Palu~\citep{chang2024palu}.} a low-rank projection baseline for KV-cache reduction under a fixed cache budget;
\textbf{ASVD~\citep{yuan2023asvd}.} activation-aware SVD applied to $(W_K,W_V)$; and
\textbf{SVD-LLM V2~\citep{wang2024svd,wang2025svdllmv2}.} truncation-aware SVD-style factorization applied to $(W_K,W_V)$.
We also include conversion baselines \textbf{TransMLA, MHA2MLA~\citep{meng2025transmla,ji2025mha2mla}.} We report their SVD-style initialization in the one-shot setting, while full 100\% MLA restoration is evaluated in Sec.~\ref{subsec:data}.
\textbf{CARE-U (uniform-rank).} Covariance-aware factorization with a uniform rank across layers.
\textbf{CARE-E (adjusted-rank).} The same covariance-aware factorization, but with our adjusted-rank allocation.
We evaluate on LM Harness tasks: Wikitext2 (Wiki)~\citep{WIKITEXT2}, ARC Challenge (ARC), ARC Easy (ARE)~\citep{clark2018think}, HellaSwag~\citep{zellers2019hellaswag}, PIQA~\citep{bisk2020piqa}, MMLU~\citep{hendrycks2020measuring}, OpenBookQA (OBQA)~\citep{mihaylov2018can}, RACE (RA)~\citep{lai2017race}, and WinoGrande (WG)~\citep{sakaguchi2021winogrande}. Hyperparameters are in App.~\ref{hyper}.

\subsection{One-shot Results}
We report \emph{one-shot}, direct KV-compression performance under KV-parity before partial-RoPE and healing on \emph{Alpaca}~\cite{ALPACA} calibration dataset; Tab.~\ref{tab:1} summarizes the main results for Llama3.1-8B and Qwen3-4B-Instruct-2507. App.~\ref{app:detailed-zero-shot} reports additional one-shot results on other models and calibration datasets.
\begingroup
  \setlength{\tabcolsep}{3.2pt}
  \renewcommand{\arraystretch}{1.08}
  \scriptsize
  \setlength{\LTleft}{0pt}
  \setlength{\LTright}{0pt}
  \begin{longtable}{cccc<{\,($\downarrow$)}ccccccccc}
    \caption{One-shot comparison on Llama3.1-8B and Qwen3-4B-Instruct-2507 against original and baselines on multiple tasks. Calibration samples: 256. Sequence length: 2048. Calibration dataset: alpaca. Higher is better for Accuracy (\%) (ACC.) ($\uparrow$) and lower is better for Perplexity (PPL.) ($\downarrow$).}\phantomsection\label{tab:1} \\
    \toprule
    \multirow{2}{*}{\textbf{Rank}} & \multirow{2}{*}{\textbf{KV Save}} & \multirow{2}{*}{\textbf{Methods}} &
    \multicolumn{1}{c}{\textbf{PPL} ($\downarrow$)} &
    \multicolumn{9}{c}{\textbf{ACC} ($\uparrow$)} \\
    \cmidrule(lr){4-4} \cmidrule(lr){5-13}
    & & &
    \textbf{ARC} &
    \textbf{ARE} &
    \textbf{HellaSwag} &
    \textbf{PIQA} &
    \textbf{MMLU} &
    \textbf{OBQA} &
    \textbf{RA} &
    \textbf{WG} &
    \textbf{AVG} \\
    \midrule
    \endfirsthead
    \toprule
    \multirow{2}{*}{\textbf{Rank}} & \multirow{2}{*}{\textbf{KV Save}} & \multirow{2}{*}{\textbf{Methods}} &
    \multicolumn{1}{c}{\textbf{PPL} ($\downarrow$)} &
    \multicolumn{9}{c}{\textbf{ACC} ($\uparrow$)} \\
    \cmidrule(lr){4-4} \cmidrule(lr){5-13}
    & & &
    \textbf{ARC} &
    \textbf{ARE} &
    \textbf{HellaSwag} &
    \textbf{PIQA} &
    \textbf{MMLU} &
    \textbf{OBQA} &
    \textbf{RA} &
    \textbf{WG} &
    \textbf{AVG} \\
    \midrule
    \endhead
    \rowcolor{myblue!14}
    \multicolumn{13}{c}{\textbf{\textcolor{myblue}{Llama-3.1-8B-Instruct}}} \\
    \midrule
    \multirow{7}{*}{\textbf{64}} & \multirow{7}{*}{\textbf{93.75}}
    & \texttt{\bfseries GQA (Original)}           & \bfseries 7.21 & \bfseries 50.34 & \bfseries 80.18 & \bfseries 60.15 & \bfseries 79.65 & \bfseries 48.05 & \bfseries 34.80 & \bfseries 40.10 & \bfseries 72.69 & \bfseries 58.24 \\
      && \texttt{Palu(SVD)}        & 2260.60 &  25.77 &  27.53 &  26.50 &  52.18 &  24.26 &  27.80 &  20.86 &  50.04 &  31.87 \\
      && \texttt{MHA2MLA}                 & 284863.91 &  25.94 &  23.82 &  26.34 &  50.92 &  25.62 &  27.80 &  22.11 &  50.59 &  31.64 \\
      && \cellcolor{myblue!4}\textbf{\texttt{CARE-U (OURS)}}   & \cellcolor{myblue!4}\bfseries 983.55 & \cellcolor{myblue!4}\bfseries 23.89 & \cellcolor{myblue!4}\bfseries 30.51 & \cellcolor{myblue!4}\bfseries 26.82 & \cellcolor{myblue!4}\bfseries 54.73 & \cellcolor{myblue!4}\bfseries 23.18 & \cellcolor{myblue!4}\bfseries 26.00 & \cellcolor{myblue!4}\bfseries 20.96 & \cellcolor{myblue!4}\bfseries 49.01 & \cellcolor{myblue!4}\bfseries 31.89 \\
      && \cellcolor{myblue!8}\textbf{\texttt{CARE-E (OURS)}}   & \cellcolor{myblue!8}\bfseries 983.03 & \cellcolor{myblue!8}\bfseries 23.63 & \cellcolor{myblue!8}\bfseries 31.31 & \cellcolor{myblue!8}\bfseries 27.49 & \cellcolor{myblue!8}\bfseries 54.62 & \cellcolor{myblue!8}\bfseries 22.98 & \cellcolor{myblue!8}\bfseries 27.40 & \cellcolor{myblue!8}\bfseries 21.15 & \cellcolor{myblue!8}\bfseries 50.36 & \cellcolor{myblue!8}\bfseries 32.37 \\
      && \texttt{ASVD}        & 2525.33 &  23.81 &  26.68 &  26.68 &  52.18 &  22.97 &  27.80 &  20.86 &  50.99 &  31.50 \\
      && \texttt{SVD-LLM V2}        & 967.04 &  23.63 &  30.72 &  26.75 &  54.90 &  23.22 &  26.40 &  20.86 &  48.93 &  31.93 \\
    \midrule
    \multirow{6}{*}{\textbf{128}} & \multirow{6}{*}{\textbf{87.50}}
    & \texttt{Palu(SVD)}        & 3046.58 &  23.89 &  27.02 &  26.62 &  50.38 &  23.14 &  24.80 &  22.49 &  49.25 &  30.95 \\
      && \texttt{MHA2MLA}                 & 15028.91 &  25.00 &  24.12 &  26.58 &  52.18 &  23.82 &  29.20 &  22.11 &  51.07 &  31.76 \\
      && \cellcolor{myblue!4}\textbf{\texttt{CARE-U (OURS)}}   & \cellcolor{myblue!4}\bfseries 398.91 & \cellcolor{myblue!4}\bfseries 26.71 & \cellcolor{myblue!4}\bfseries 41.96 & \cellcolor{myblue!4}\bfseries 33.64 & \cellcolor{myblue!4}\bfseries 60.99 & \cellcolor{myblue!4}\bfseries 26.06 & \cellcolor{myblue!4}\bfseries 27.00 & \cellcolor{myblue!4}\bfseries 24.21 & \cellcolor{myblue!4}\bfseries 53.20 & \cellcolor{myblue!4}\bfseries 36.72 \\
      && \cellcolor{myblue!8}\textbf{\texttt{CARE-E (OURS)}}   & \cellcolor{myblue!8}\bfseries 353.74 & \cellcolor{myblue!8}\bfseries 27.30 & \cellcolor{myblue!8}\bfseries 42.47 & \cellcolor{myblue!8}\bfseries 37.96 & \cellcolor{myblue!8}\bfseries 62.46 & \cellcolor{myblue!8}\bfseries 29.38 & \cellcolor{myblue!8}\bfseries 28.20 & \cellcolor{myblue!8}\bfseries 26.12 & \cellcolor{myblue!8}\bfseries 54.85 & \cellcolor{myblue!8}\bfseries 38.59 \\
      && \texttt{ASVD}         & 1675.54 &  25.00 &  27.99 &  26.61 &  52.39 &  23.04 &  26.60 &  21.82 &  48.46 &  31.49 \\
      && \texttt{SVD-LLM V2}   & 386.23 &  27.56 &  42.38 &  34.06 &  61.04 &  26.17 &  27.80 &  24.21 &  53.99 &  37.15 \\
    \midrule
    \multirow{6}{*}{\textbf{256}} & \multirow{6}{*}{\textbf{75.00}}
    & \texttt{Palu(SVD)}        & 537.57 &  22.61 &  31.10 &  27.93 &  55.22 &  22.82 &  27.00 &  22.68 &  51.70 &  32.63 \\
      && \texttt{MHA2MLA}                 & 1633.65 &  27.47 &  25.63 &  26.50 &  52.39 &  23.10 &  28.20 &  22.49 &  50.83 &  32.08 \\
      && \cellcolor{myblue!4}\textbf{\texttt{CARE-U (OURS)}}   & \cellcolor{myblue!4}\bfseries 48.35 & \cellcolor{myblue!4}\bfseries 36.09 & \cellcolor{myblue!4}\bfseries 60.19 & \cellcolor{myblue!4}\bfseries 55.75 & \cellcolor{myblue!4}\bfseries 71.87 & \cellcolor{myblue!4}\bfseries 45.64 & \cellcolor{myblue!4}\bfseries 33.20 & \cellcolor{myblue!4}\bfseries 35.22 & \cellcolor{myblue!4}\bfseries 62.04 & \cellcolor{myblue!4}\bfseries 50.00 \\
      && \cellcolor{myblue!8}\textbf{\texttt{CARE-E (OURS)}}   & \cellcolor{myblue!8}\bfseries 49.43 & \cellcolor{myblue!8}\bfseries 38.48 & \cellcolor{myblue!8}\bfseries 60.06 & \cellcolor{myblue!8}\bfseries 60.54 & \cellcolor{myblue!8}\bfseries 72.42 & \cellcolor{myblue!8}\bfseries 54.29 & \cellcolor{myblue!8}\bfseries 33.60 & \cellcolor{myblue!8}\bfseries 35.50 & \cellcolor{myblue!8}\bfseries 66.06 & \cellcolor{myblue!8}\bfseries 52.62 \\
      && \texttt{ASVD}         & 312.86 &  21.76 &  32.03 &  29.52 &  55.22 &  23.05 &  25.20 &  24.59 &  52.64 &  33.00 \\
      && \texttt{SVD-LLM V2}   & 47.28 &  36.26 &  60.69 &  56.39 &  72.20 &  46.57 &  33.40 &  34.83 &  60.69 &  50.13 \\
    \midrule
    \multirow{6}{*}{\textbf{512}}  & \multirow{6}{*}{\textbf{50.00}}
    & \texttt{Palu(SVD)}        & 45.40 &  28.16 &  45.96 &  43.37 &  64.15 &  24.98 &  30.80 &  23.83 &  53.43 &  39.33 \\
      && \texttt{MHA2MLA}                 & 220.29 &  25.94 &  40.95 &  39.27 &  61.37 &  25.54 &  26.60 &  26.79 &  56.59 &  37.88 \\
      && \cellcolor{myblue!4}\textbf{\texttt{CARE-U (OURS)}}   & \cellcolor{myblue!4}\bfseries 9.64 & \cellcolor{myblue!4}\bfseries 52.73 & \cellcolor{myblue!4}\bfseries 76.30 & \cellcolor{myblue!4}\bfseries 73.98 & \cellcolor{myblue!4}\bfseries 78.73 & \cellcolor{myblue!4}\bfseries 62.17 & \cellcolor{myblue!4}\bfseries 40.60 & \cellcolor{myblue!4}\bfseries 41.53 & \cellcolor{myblue!4}\bfseries 72.61 & \cellcolor{myblue!4}\bfseries 62.33 \\
      && \cellcolor{myblue!8}\textbf{\texttt{CARE-E (OURS)}} & \cellcolor{myblue!8}\bfseries 19.50 & \cellcolor{myblue!8}\bfseries 42.41 & \cellcolor{myblue!8}\bfseries 64.69 & \cellcolor{myblue!8}\bfseries 68.96 & \cellcolor{myblue!8}\bfseries 75.46 & \cellcolor{myblue!8}\bfseries 63.98 & \cellcolor{myblue!8}\bfseries 37.00 & \cellcolor{myblue!8}\bfseries 38.47 & \cellcolor{myblue!8}\bfseries 71.59 & \cellcolor{myblue!8}\bfseries 57.82 \\
      && \texttt{ASVD}         & 12.02 &  46.33 &  69.11 &  70.75 &  76.17 &  41.80 &  36.60 &  33.97 &  66.85 &  55.20 \\
      && \texttt{SVD-LLM V2}   & 9.63 &  52.39 &  76.68 &  73.57 &  78.45 &  62.31 &  40.40 &  41.63 &  72.22 &  62.21 \\
    \midrule
    \rowcolor{myblue!14}
    \multicolumn{13}{c}{\textbf{\textcolor{myblue}{Qwen3-4B-Instruct-2507}}} \\
    \midrule
    \multirow{7}{*}{\textbf{64}}
    & \multirow{7}{*}{\textbf{93.75}}
    & \texttt{\bfseries GQA (Original)}           & \bfseries 10.04 & \bfseries 55.89 & \bfseries 83.12 & \bfseries 52.65 & \bfseries 76.01 & \bfseries 73.37 & \bfseries 32.00 & \bfseries 41.24 & \bfseries 68.11 & \bfseries 60.30 \\
      && \texttt{Palu(SVD)}        & 56922.11 & 25.34 & 26.77 & 25.73 & 50.44 & 24.31 & 28.40 & 22.11 & 50.91 & 31.75 \\
      && \texttt{MHA2MLA}                 & 21850.16 & 25.68 & 25.46 & 26.03 & 51.96 & 22.92 & 27.00 & 22.01 & 50.20 & 31.41 \\
      && \cellcolor{myblue!4}\textbf{\texttt{CARE-U (OURS)}}   & \cellcolor{myblue!4}\bfseries 905.74 & \cellcolor{myblue!4}\bfseries 23.46 & \cellcolor{myblue!4}\bfseries 29.00 & \cellcolor{myblue!4}\bfseries 27.61 & \cellcolor{myblue!4}\bfseries 55.22 & \cellcolor{myblue!4}\bfseries 23.03 & \cellcolor{myblue!4}\bfseries 26.60 & \cellcolor{myblue!4}\bfseries 23.06 & \cellcolor{myblue!4}\bfseries 50.59 & \cellcolor{myblue!4}\bfseries 32.32 \\
      && \cellcolor{myblue!8}\textbf{\texttt{CARE-E (OURS)}}   & \cellcolor{myblue!8}\bfseries 730.93 & \cellcolor{myblue!8}\bfseries 23.29 & \cellcolor{myblue!8}\bfseries 30.77 & \cellcolor{myblue!8}\bfseries 28.33 & \cellcolor{myblue!8}\bfseries 55.22 & \cellcolor{myblue!8}\bfseries 22.95 & \cellcolor{myblue!8}\bfseries 24.80 & \cellcolor{myblue!8}\bfseries 22.78 & \cellcolor{myblue!8}\bfseries 51.54 & \cellcolor{myblue!8}\bfseries 32.46 \\
      && \texttt{ASVD}        & 6683.95 & 27.39 & 25.00 & 25.71 & 50.38 & 24.08 & 26.60 & 22.11 & 50.43 & 31.46 \\
      && \texttt{SVD-LLM V2}        & 894.94 & 23.12 & 28.58 & 27.66 & 55.11 & 23.02 & 27.20 & 23.44 & 51.54 & 32.46 \\
    \midrule
    \multirow{6}{*}{\textbf{128}} & \multirow{6}{*}{\textbf{87.50}}
    & \texttt{Palu(SVD)}        & 22048.79 & 26.02 & 26.18 & 26.29 & 51.09 & 24.49 & 25.40 & 21.05 & 49.72 & 31.28 \\
      && \texttt{MHA2MLA}                 & 52683.47 & 23.81 & 26.56 & 26.74 & 52.18 & 24.74 & 28.20 & 22.68 & 49.72 & 31.83 \\
      && \cellcolor{myblue!4}\textbf{\texttt{CARE-U (OURS)}}   & \cellcolor{myblue!4}\bfseries 111.17 & \cellcolor{myblue!4}\bfseries 27.47 & \cellcolor{myblue!4}\bfseries 39.69 & \cellcolor{myblue!4}\bfseries 37.52 & \cellcolor{myblue!4}\bfseries 60.50 & \cellcolor{myblue!4}\bfseries 27.03 & \cellcolor{myblue!4}\bfseries 27.60 & \cellcolor{myblue!4}\bfseries 26.99 & \cellcolor{myblue!4}\bfseries 53.59 & \cellcolor{myblue!4}\bfseries 37.55 \\
      && \cellcolor{myblue!8}\textbf{\texttt{CARE-E (OURS)}}   & \cellcolor{myblue!8}\bfseries 102.38 & \cellcolor{myblue!8}\bfseries 30.29 & \cellcolor{myblue!8}\bfseries 44.82 & \cellcolor{myblue!8}\bfseries 42.54 & \cellcolor{myblue!8}\bfseries 63.76 & \cellcolor{myblue!8}\bfseries 30.52 & \cellcolor{myblue!8}\bfseries 29.40 & \cellcolor{myblue!8}\bfseries 28.71 & \cellcolor{myblue!8}\bfseries 54.54 & \cellcolor{myblue!8}\bfseries 40.57 \\
      && \texttt{ASVD}         & 1682.84 & 22.95 & 30.01 & 29.29 & 52.34 & 23.42 & 27.00 & 23.54 & 50.12 & 32.33 \\
      && \texttt{SVD-LLM V2}   & 116.76 & 27.05 & 40.03 & 37.02 & 61.04 & 26.69 & 26.80 & 25.74 & 53.28 & 37.21 \\
    \midrule
    \multirow{6}{*}{\textbf{256}} & \multirow{6}{*}{\textbf{75.00}}
    & \texttt{Palu(SVD)}        & 2561.97 & 26.11 & 29.46 & 29.92 & 52.88 & 24.48 & 28.40 & 23.64 & 51.70 & 33.32 \\
      && \texttt{MHA2MLA}                 & 44509.79 & 22.18 & 28.49 & 28.92 & 52.45 & 23.00 & 24.60 & 24.59 & 51.30 & 31.94 \\
      && \cellcolor{myblue!4}\textbf{\texttt{CARE-U (OURS)}}   & \cellcolor{myblue!4}\bfseries 22.08 & \cellcolor{myblue!4}\bfseries 46.42 & \cellcolor{myblue!4}\bfseries 68.90 & \cellcolor{myblue!4}\bfseries 59.16 & \cellcolor{myblue!4}\bfseries 71.55 & \cellcolor{myblue!4}\bfseries 54.76 & \cellcolor{myblue!4}\bfseries 36.40 & \cellcolor{myblue!4}\bfseries 35.02 & \cellcolor{myblue!4}\bfseries 62.43 & \cellcolor{myblue!4}\bfseries 54.33 \\
      && \cellcolor{myblue!8}\textbf{\texttt{CARE-E (OURS)}}   & \cellcolor{myblue!8}\bfseries 28.84 & \cellcolor{myblue!8}\bfseries 41.30 & \cellcolor{myblue!8}\bfseries 59.22 & \cellcolor{myblue!8}\bfseries 56.53 & \cellcolor{myblue!8}\bfseries 69.37 & \cellcolor{myblue!8}\bfseries 53.50 & \cellcolor{myblue!8}\bfseries 35.20 & \cellcolor{myblue!8}\bfseries 32.82 & \cellcolor{myblue!8}\bfseries 61.88 & \cellcolor{myblue!8}\bfseries 51.23 \\
      && \texttt{ASVD}         & 63.15 & 32.76 & 46.84 & 47.45 & 63.93 & 26.38 & 30.80 & 30.05 & 52.01 & 41.28 \\
      && \texttt{SVD-LLM V2}   & 22.88 & 44.54 & 67.17 & 57.77 & 70.78 & 52.81 & 35.60 & 34.83 & 61.48 & 53.12 \\
    \midrule
    \multirow{6}{*}{\textbf{512}}  & \multirow{6}{*}{\textbf{50.00}}
    & \texttt{Palu(SVD)}        & 33.97 & 35.58 & 47.64 & 50.44 & 65.18 & 27.85 & 32.80 & 30.24 & 52.64 & 42.80 \\
      && \texttt{MHA2MLA}                 & 100.99 & 27.05 & 41.08 & 37.97 & 59.19 & 29.14 & 27.20 & 29.47 & 54.06 & 38.15 \\
      && \cellcolor{myblue!4}\textbf{\texttt{CARE-U (OURS)}}   & \cellcolor{myblue!4}\bfseries 12.03 & \cellcolor{myblue!4}\bfseries 54.95 & \cellcolor{myblue!4}\bfseries 77.23 & \cellcolor{myblue!4}\bfseries 69.24 & \cellcolor{myblue!4}\bfseries 76.22 & \cellcolor{myblue!4}\bfseries 67.46 & \cellcolor{myblue!4}\bfseries 40.00 & \cellcolor{myblue!4}\bfseries 39.43 & \cellcolor{myblue!4}\bfseries 68.43 & \cellcolor{myblue!4}\bfseries 61.62 \\
      && \cellcolor{myblue!8}\textbf{\texttt{CARE-E (OURS)}}   & \cellcolor{myblue!8}\bfseries 15.91 & \cellcolor{myblue!8}\bfseries 49.23 & \cellcolor{myblue!8}\bfseries 70.88 & \cellcolor{myblue!8}\bfseries 64.13 & \cellcolor{myblue!8}\bfseries 72.80 & \cellcolor{myblue!8}\bfseries 64.16 & \cellcolor{myblue!8}\bfseries 36.20 & \cellcolor{myblue!8}\bfseries 36.56 & \cellcolor{myblue!8}\bfseries 64.56 & \cellcolor{myblue!8}\bfseries 57.31 \\
      && \texttt{ASVD}         & 15.49 & 47.61 & 66.54 & 67.49 & 73.01 & 56.56 & 35.60 & 35.22 & 62.75 & 55.60 \\
      && \texttt{SVD-LLM V2}   & 11.88 & 54.61 & 77.44 & 68.53 & 75.68 & 67.65 & 39.80 & 38.18 & 67.25 & 61.14 \\
    \bottomrule
  \end{longtable}
\endgroup

\begin{figure}[H]\centering
  \begin{minipage}[t]{0.49\linewidth}
  \centering
  \includegraphics[width=\linewidth,trim={0.2cm 0 0.2cm 0},clip]{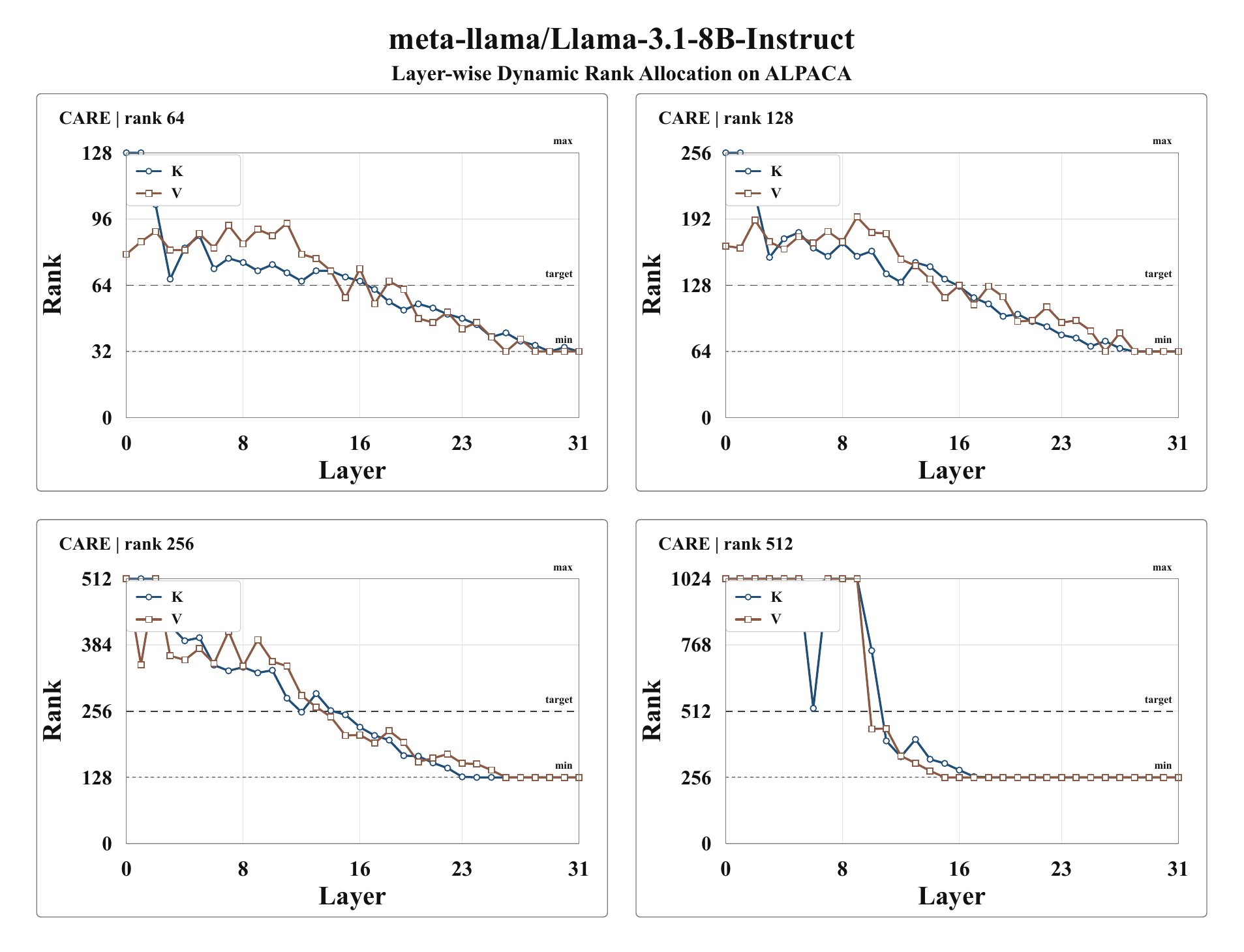}
  \end{minipage}\hspace{0.01\linewidth}
  \begin{minipage}[t]{0.49\linewidth}
  \centering
  \includegraphics[width=\linewidth,trim={0.2cm 0 0.2cm 0},clip]{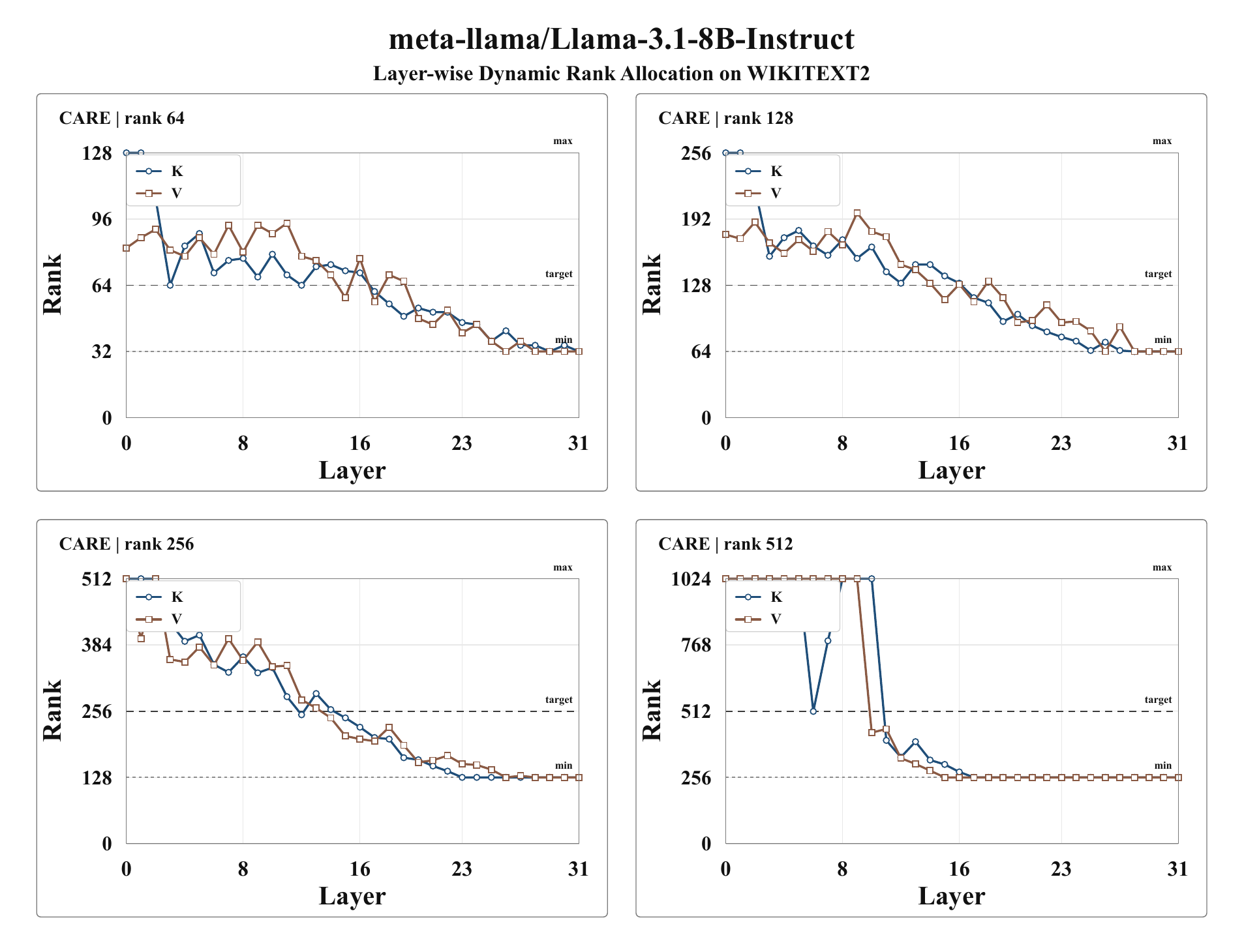}
  \end{minipage}
  
  \vspace{0.15em}
  
  \begin{minipage}[t]{0.49\linewidth}
  \centering
  \includegraphics[width=\linewidth,trim={0.2cm 0 0.2cm 0},clip]{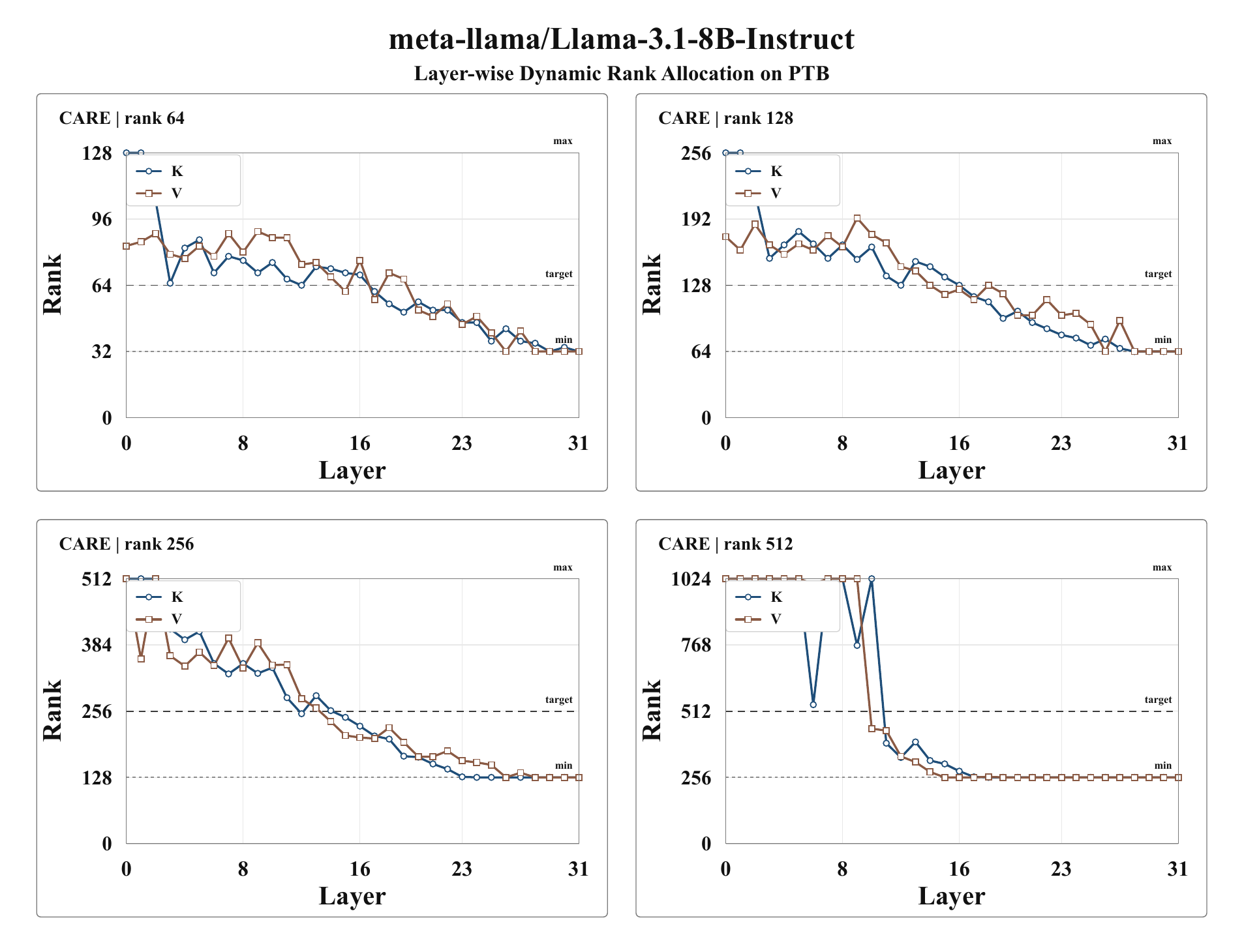}
  \end{minipage}\hspace{0.01\linewidth}
  \begin{minipage}[t]{0.49\linewidth}
  \centering
  \includegraphics[width=\linewidth,trim={0.2cm 0 0.2cm 0},clip]{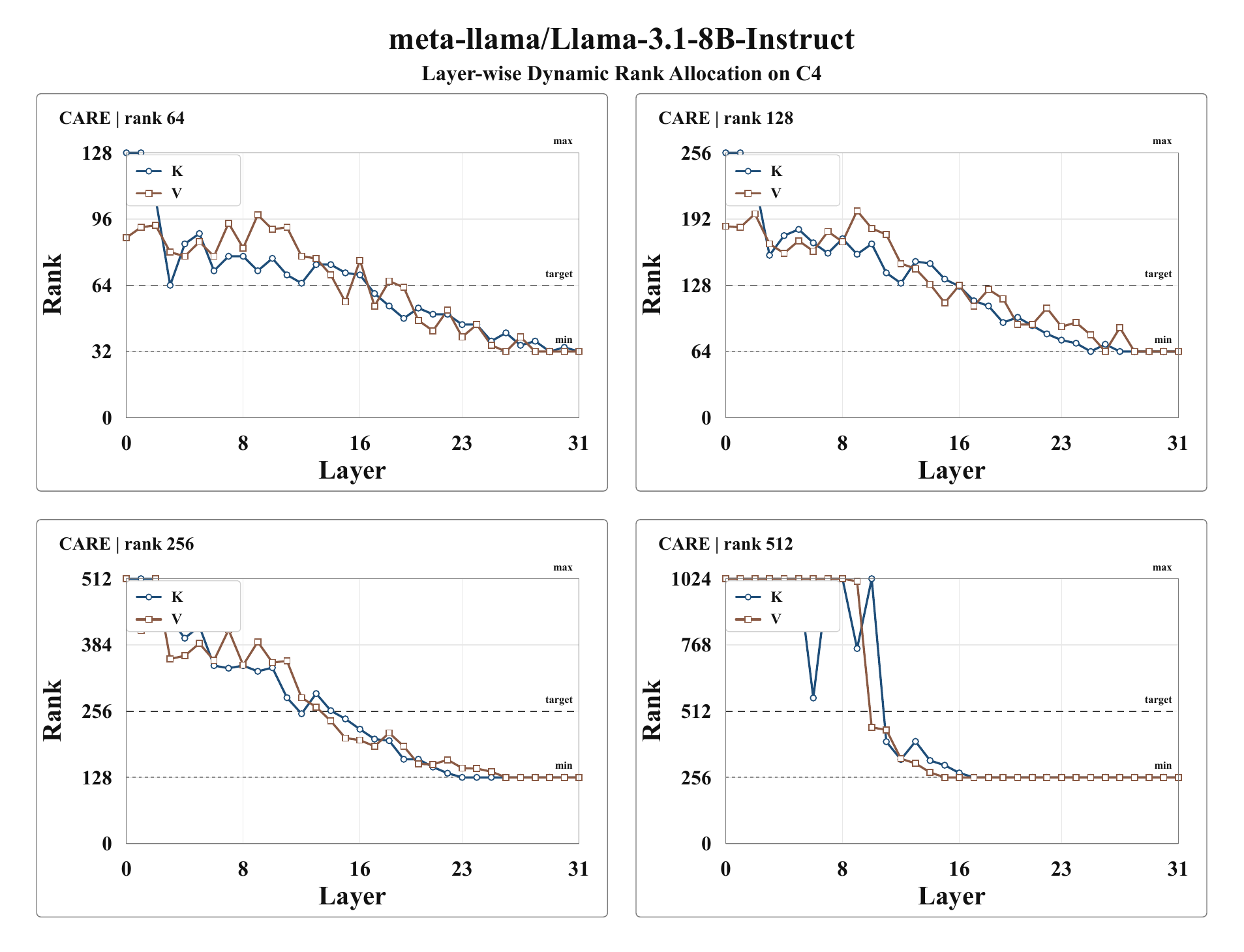}
  \end{minipage}
  \vspace{-0.5em}
  \caption{Covariance-aware rank profiles across calibration corpora (Alpaca, WikiText2, PTB, C4) for Llama-3.1-8B-Instruct at target ranks 64, 128, 256, and 512. Across all target budgets, both $W_K$ and $W_V$ show a depth-dependent increase—small in early layers, rising through mid layers—with stronger late-layer growth for $W_V$. The consistency across corpora suggests a model-intrinsic trend.}
  \label{fig:ranks}\vspace{-0.5em}
\end{figure}

At very low rank, one-shot MLA remains challenging for all methods, but regarding perplexity, CARE already improves substantially over the naive Palu(SVD) or MHA2MLA initialization. 
As the rank budget increases to 256/512, the advantage of covariance-aware initialization becomes much clearer, especially on Llama-3.1-8B, Qwen3-4B where CARE-E gives the strongest overall accuracy-PPL trade-off. 
We provide additional one-shot long-context retrieval results and qualitative examples in App.~\ref{sec:nih} and App.~\ref{app:examples}, respectively.

\subsection{Ablation Studies}
\label{exp:as}
\subsubsection{Energy Distribution vs. Covariance}
\label{exp:as:en}

Using the covariance-aware energy in Sec.~\ref{sec:method}, we obtain highly consistent rank profiles across C4~\citep{C4}, 
Alpaca~\citep{ALPACA}, WikiText2~\citep{WIKITEXT2}, and PTB~\citep{PTB}. As shown in Fig.~\ref{fig:ranks}, rank is front-loaded: 
across target ranks 64, 128, 256, and 512, both $W_K$ and $W_V$ receive much larger ranks in the first layers and then decrease steadily with depth, approaching the minimum rank in later layers. 
This indicates that early layers are more accuracy-critical, while deeper layers can be compressed more aggressively. 
The same trend also appears on other models (From 1.5B - 70B) (App.~\ref{app:other-ranks}), suggesting that the profile is largely \emph{model-intrinsic} rather than tied to a particular calibration corpus.

\subsubsection{Accuracy Impact of Covariance}
\label{exp:as:cov}
\label{exp:as:var}

\begin{wrapfigure}{r}{0.4\columnwidth}
  \centering
  \includegraphics[width=\linewidth,height=0.24\textheight,keepaspectratio]{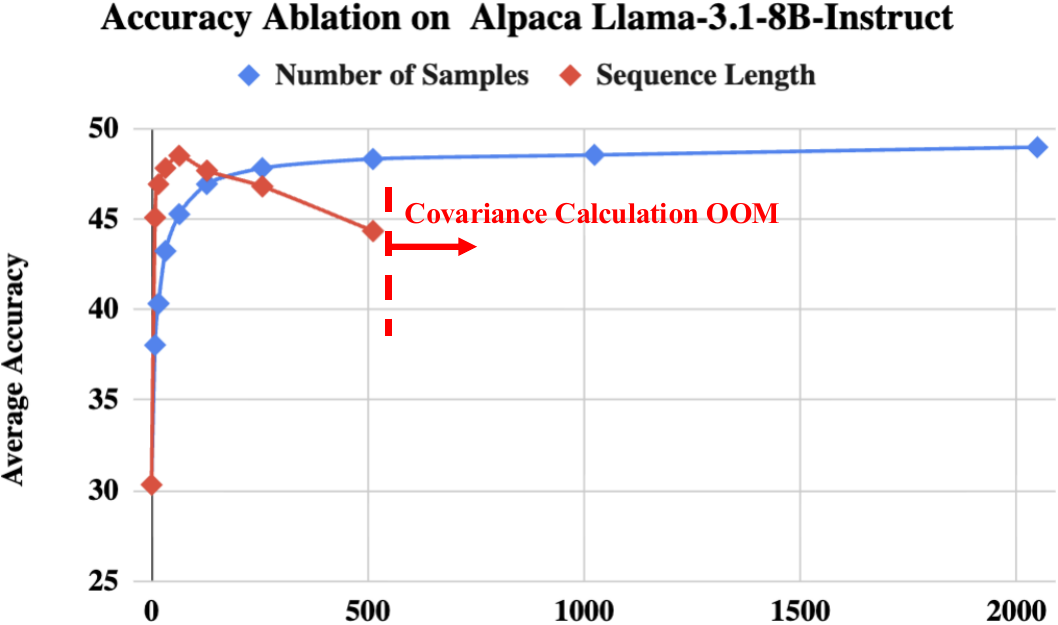}
  \caption{One-shot accuracy versus calibration samples and sequence length across eight benchmarks. Sequence length is fixed at 256 unless otherwise noted; the red curve denotes fixed 256 samples with varying sequence length. OOM occurs beyond length $=512$ during covariance computation.}
  \label{fig:sample-seqlen}
  \vspace{-1.2\baselineskip}
\end{wrapfigure}

Using Llama-3.1-8B-Instruct, we vary (i) calibration samples, (ii) sequence length, and (iii) calibration corpus.
Fig.~\ref{fig:sample-seqlen} shows that CARE is already strong with small calibration budgets: performance saturates beyond 512 samples, and longer sequences tend to overfit limited calibration data. We therefore use 256 samples with a short sequence length (32) as the default trade-off.
Across corpora (\textsc{Alpaca}, \textsc{C4}, \textsc{WikiText2}, \textsc{PTB}, \textsc{ARC}, \textsc{ARE}, \textsc{MMLU}), one-shot accuracy changes are modest on average (Tab.~\ref{tab:2}); domain-aligned corpora give mostly local gains. We default to \textsc{Alpaca} for broad coverage and stable generalization.

Additional ablations are deferred to the appendix for readability. App.~\ref{app:shrinkage} sweeps the shrinkage coefficient, showing that CARE is not sensitive to the exact regularization magnitude and supporting our default $\alpha=0.01$. App.~\ref{exp:as:shift} studies cross-domain calibration under distribution shift, App.~\ref{exp:as:mixcov} examines task-weighted covariance mixing, and App.~\ref{app:sqrtc-vs-c} compares the default $\sqrt{C}$ formulation against using $C$ directly under matched settings. Although the derivation is based on $\sqrt{C}$, the appendix shows that $\sqrt{C}$ is more consistent than using $C$ directly, especially beyond the lowest ranks.

\subsection{Recovery 100\% MLA with small SFT budgets}
\label{subsec:data}

\paragraph{Setup (healing baselines).}
We compare CARE against Palu(SVD) and TransMLA under matched healing budgets at MLA rank 512.
Our hybrid \texttt{TransMLA + CARE(E) Init} keeps the same 100\% MLA restoration and healing pipeline as TransMLA,
but when TransMLA performs the KV low-rank mapping, we replace its original initialization with the covariance-aware $\sqrt{C}W$ SVD used by CARE(E).
All subsequent restoration and healing stages are unchanged. CARE-based 100 use \texttt{alpaca-256-256} calibration (256 samples, sequence length 256),
TransMLA uses \texttt{wiki-256-256} (256 samples, sequence length 256), and Palu(SVD) does not require a separate calibration corpus. We report the one-shot initialization (0B) and healed checkpoints after 1B and 3B tokens on the same LM Harness suite.

\begin{table*}[!htb]
  \centering
  \setlength{\tabcolsep}{5pt}
  \renewcommand{\arraystretch}{1.08}
  \scriptsize
  \centering
  \caption{\small Healed Llama3.1-8B-Instruct comparison at MLA rank 512 across token budgets.}
  \label{tab:3}
  \begin{adjustbox}{max width=0.92\textwidth}
  \begin{tabular}{c|c|c|c|c|c|c|c|c|c|c|c|c}
    \toprule
    \multirow{1}{*}{\textbf{Rank}} &
    \multirow{1}{*}{\textbf{Methods}} &
    \multicolumn{1}{c|}{\textbf{Calibration}} &
    \multicolumn{1}{c|}{\textbf{Token}} &
    \multicolumn{1}{c|}{\textbf{ARC} ($\uparrow$)} &
    \multicolumn{1}{c|}{\textbf{ARE} ($\uparrow$)} &
    \multicolumn{1}{c|}{\textbf{HellaSwag} ($\uparrow$)}  &
    \multicolumn{1}{c|}{\textbf{PIQA} ($\uparrow$)}  &
    \multicolumn{1}{c|}{\textbf{MMLU} ($\uparrow$)}  &
    \multicolumn{1}{c|}{\textbf{OBQA} ($\uparrow$)}  &
    \multicolumn{1}{c|}{\textbf{RA} ($\uparrow$)}  &
    \multicolumn{1}{c|}{\textbf{WG} ($\uparrow$)}  &
    \multicolumn{1}{c}{\textbf{AVG} ($\uparrow$)}  \\
    \midrule
    \textbf{--}
      & \texttt{\bfseries GQA (original)} & \texttt{N/A} & \texttt{N/A} & \bfseries 50.34 & \bfseries 80.18 & \bfseries 60.15 & \bfseries 79.65 & \bfseries 48.05 & \bfseries 34.80 & \bfseries 40.10 & \bfseries 72.69 & \bfseries 58.24 \\
    \midrule
    \multirow{10}{*}{\textbf{512}}
      & \texttt{Palu(SVD)} & \texttt{N/A} & 0B & 26.02 & 50.97 & 37.43 & 64.15 & 27.89 & 19.40 & 26.60 & 57.54 & 38.75 \\
      & \texttt{Palu(SVD)} & \texttt{N/A} & 1B & 33.45 & 62.38 & 45.55 & 72.34 & 50.57 & 23.00 & 31.54 & 61.78 & 47.58 \\
      & \texttt{Palu(SVD)} & \texttt{N/A} & 3B & 44.56 & 74.96 & 52.20 & 76.63 & 61.08 & 30.40 & 45.15 & 65.42 & 56.30 \\
      & \cellcolor{myblue!4}\textbf{\texttt{CARE (OURS) ONE-SHOT}} & \cellcolor{myblue!4}\texttt{alpaca-256-32} & \cellcolor{myblue!4}0B & \cellcolor{myblue!4}\bfseries 52.73 & \cellcolor{myblue!4}\bfseries 76.30 & \cellcolor{myblue!4}\bfseries 73.98 & \cellcolor{myblue!4}\bfseries 78.73 & \cellcolor{myblue!4}\bfseries 62.17 & \cellcolor{myblue!4}\bfseries 40.60 & \cellcolor{myblue!4}\bfseries 41.53 & \cellcolor{myblue!4}\bfseries 72.61 & \cellcolor{myblue!4}\bfseries 62.33 \\
      & \cellcolor{myblue!8}\textbf{\texttt{TransMLA + CARE(E) Init (OURS, 100\% MLA Restore)}} & \cellcolor{myblue!8}\texttt{alpaca-256-256} & \cellcolor{myblue!8}0B & \cellcolor{myblue!8}\bfseries 45.05 & \cellcolor{myblue!8}\bfseries 69.02 & \cellcolor{myblue!8}\bfseries 68.98 & \cellcolor{myblue!8}\bfseries 76.06 & \cellcolor{myblue!8}\bfseries 51.16 & \cellcolor{myblue!8}\bfseries 37.60 & \cellcolor{myblue!8}\bfseries 39.43 & \cellcolor{myblue!8}\bfseries 68.98 & \cellcolor{myblue!8}\bfseries 57.04 \\
      & \cellcolor{myblue!8}\textbf{\texttt{TransMLA + CARE(E) Init (OURS, 100\% MLA Restore)}} & \cellcolor{myblue!8}\texttt{alpaca-256-256} & \cellcolor{myblue!8}1B & \cellcolor{myblue!8}\bfseries 52.25 & \cellcolor{myblue!8}\bfseries 82.33 & \cellcolor{myblue!8}\bfseries 62.47 & \cellcolor{myblue!8}\bfseries 80.21 & \cellcolor{myblue!8}\bfseries 70.31 & \cellcolor{myblue!8}\bfseries 32.90 & \cellcolor{myblue!8}\bfseries 45.11 & \cellcolor{myblue!8}\bfseries 75.13 & \cellcolor{myblue!8}\bfseries 62.59 \\
      & \cellcolor{myblue!8}\textbf{\texttt{TransMLA + CARE(E) Init (OURS, 100\% MLA Restore)}} & \cellcolor{myblue!8}\texttt{alpaca-256-256} & \cellcolor{myblue!8}3B & \cellcolor{myblue!8}\bfseries 51.75 & \cellcolor{myblue!8}\bfseries 80.73 & \cellcolor{myblue!8}\bfseries 64.45 & \cellcolor{myblue!8}\bfseries 83.23 & \cellcolor{myblue!8}\bfseries 71.57 & \cellcolor{myblue!8}\bfseries 34.00 & \cellcolor{myblue!8}\bfseries 46.33 & \cellcolor{myblue!8}\bfseries 74.09 & \cellcolor{myblue!8}\bfseries 63.27 \\
      & \texttt{TransMLA} & \texttt{wiki-256-256} & 0B & 43.17 & 66.12 & 68.25 & 74.81 & 48.89 & 38.80 & 37.70 & 66.46 & 55.52 \\
      & \texttt{TransMLA} & \texttt{wiki-256-256} & 1B & 53.04 & 81.07 & 58.75 & 81.04 & 69.13 & 32.00 & 44.09 & 71.74 & 61.36 \\
      & \texttt{TransMLA} & \texttt{wiki-256-256} & 3B & 53.77 & 82.34 & 56.44 & 80.70 & 70.23 & 33.30 & 45.61 & 72.47 & 61.86 \\
    \bottomrule
  \end{tabular}
  \end{adjustbox}
\end{table*}

We sweep small-scale SFT budgets to quantify post-conversion recovery; 
results are shown in Tab.~\ref{tab:3}. 
CARE(E) already starts from the strongest one-shot initialization, 
and this advantage largely remains after healing. At 1B and 3B tokens, 
\texttt{TransMLA + CARE(E) Init} reaches 62.59 and 63.27 average accuracy, 
outperforming TransMLA (61.36/61.86) and Palu(SVD) (47.58/56.30) at matched budgets. 
While both Palu(SVD) and TransMLA improve substantially with more tokens, 
CARE attains the best overall recovery with less reliance on long healing runs; 
optimization objectives, datasets, and hyperparameters in App.~\ref{hyper}. A system-level KV-cache efficiency analysis is given in App.~\ref{app:efficiency}.

\section{Related Work}\label{sec:related}
\paragraph{Conversion from Traditional Attention to MLA.}
Standard multi-head attention (MHA) underpins modern LLMs but induces a KV cache that scales linearly with sequence length and head width, creating a memory and bandwidth bottleneck at inference time~\citep{vaswani2017attention,kwon2023vllm}. Grouped-Query Attention (GQA) reduces KV heads by sharing keys/values across query groups, lowering KV memory while sacrificing expressiveness~\citep{ainslie2023gqa,shazeer2019mqa}. Multi-Head Latent Attention (MLA) addresses KV memory by caching low-dimensional latents with lightweight up/down projections~\citep{deepseekv2}. 
Beyond MLA that trains from scratch, several \emph{post-hoc} pathways demonstrate conversion feasibility: \emph{TransMLA} gives a theory-based reduction from GQA to MLA at corresponding KV budget~\citep{meng2025transmla}, \emph{MHA2MLA} focuses on practical alignment via partial RoPE and joint-SVD initialization~\citep{ji2025mha2mla}, and \emph{X\mbox{-}EcoMLA} explores distillation-based upcycling of pretrained attention into MLA for extreme KV compression~\citep{li2025xecoMLA}. Also, following \emph{MambaInLlama}~\citep{wang2024mamba}, \emph{Zebra-Llama} composes efficient hybrids to improve inference efficiency and can be paired with MLA-style KV reductions in deployed systems~\citep{yang2025zebrallama}.

\paragraph{SVD inspirations.}
Naive SVD truncation minimizes $\|W - W_r\|_F$ and ranks by raw singular values, which need not correlate with downstream loss~\citep{eckart1936approximation,LeCun1990OBD,Hassibi1993OBS,Dong2019HAWQ,frantar2022gptq,krzanowski2000principles}. Recent works refine this in LLMs: \emph{FWSVD} (weighted by Fisher information)~\citep{hua2022tfwsvd}, \emph{SVD\mbox{-}LLM} (truncation-aware whitening + sequential low-rank updates) and its \emph{V2} variant (improved truncation/rank selection)~\citep{wang2024svd,wang2025svdllmv2}. \emph{SoCo} learns a diagonal reweighting to optimize the singular spectrum directly for compression rather than trusting singular-value magnitudes~\citep{li2025soco}, while \emph{Dobi\mbox{-}SVD} introduces a differentiable SVD that targets activation-side truncation and efficient reconstruction~\citep{wang2025dobi}. Together with architecture-aware conversions~\citep{meng2025transmla,ji2025mha2mla}, these SVD-oriented techniques motivate \emph{data/curvature-aware} orientations and \emph{non-uniform rank} allocation as core tools for preserving pretrained knowledge.
Relatedly, \emph{CorDA} leverages context/activation statistics to guide decomposition adaptation for parameter-efficient fine-tuning, further supporting the value of covariance-aware orientations beyond weight-only SVD~\citep{yang2024coda}.

\paragraph{SVD for cache compression.}
Orthogonally, some methods compress the \emph{cache itself}. \emph{Palu} compresses KV-cache with low-rank projection, reconstructing full $K,V$ on the fly with efficient rank search and 
quantization interoperability~\citep{chang2024palu}. \emph{ReALLM} combines low-rank components with vector-quantized latents under a unified recipe~\citep{reallm2025}. 
These methods can be stacked with MLA-style conversions or used standalone to further lower KV memory.

For more related works, please refer to App.~\ref{other_related}.
\section{Conclusions}\label{sec:conclusions}

We proposed \textbf{CARE}, a \emph{Covariance-Aware, Rank-Enhanced} procedure for migrating traditional attention to MLA under a fixed KV-parity. 
CARE replaces naïve weight-only SVD with a covariance-weighted factorization and assigns per-layer ranks via an energy-driven, water-filling schedule. 
Empirically, CARE preserves MLA’s identical KV footprint while delivering lower zero-shot perplexity and higher accuracy before healing, and better final performance than competing baselines under the same post-conversion tuning budget. 
It is also more robust under aggressive rank reduction, providing a stronger starting point for brief SFT to recover the original model's accuracy. 

\newpage
\section{Ethics Statement}
We have read and will comply with the ICLR Code of Ethics. Our study involves no human subjects, personally identifiable information, or user-generated content. All datasets are standard, publicly available benchmarks used under their respective licenses; we do not collect or infer demographic attributes. The work focuses on model architecture/optimization and does not introduce capabilities intended for surveillance, profiling, or other harmful use. We identify no foreseeable risks related to privacy, security, fairness, or legal/regulatory compliance, and no IRB/ethics approval was required. To support transparency, we will release code, configuration files, and clear instructions to reproduce all results. All findings are reported honestly without fabrication or inappropriate manipulation. The authors declare no conflicts of interest and no external sponsorship that could bias the work.

\section{Reproducibility Statement}
We provide an open-source repository (\url{https://github.com/FutureMLS-Lab/CARE}). The repo includes exact configuration files for all experiments in Tab.~\ref{tab:1}--\ref{tab:3} and Fig.~\ref{fig:ranks}--\ref{fig:sample-seqlen}, 
scripts to download and verify datasets, deterministic preprocessing, fixed random seeds, and environment specifications (Conda with pinned versions). 
Algorithmic details appear in Sec.~\ref{sec:method}; dataset descriptions and evaluation setup are given in Sec.~\ref{sec:results}; hyperparameters, hardware, optimizer and scheduler configurations are listed in App.~\ref{hyper}. 
Detailed zero-shot tables for all models and calibration corpora are in App.~\ref{app:detailed-zero-shot}; additional rank profiles (1.5B--70B, including MoE) in App.~\ref{app:other-ranks}; 
ablations on shrinkage (App.~\ref{app:shrinkage}), distribution shift (App.~\ref{exp:as:shift}), covariance mixing (App.~\ref{exp:as:mixcov}), and $\sqrt{C}$ vs.\ $C$ (App.~\ref{app:sqrtc-vs-c}); 
long-context NiH evaluation in App.~\ref{sec:nih}; and KV-cache efficiency analysis in App.~\ref{app:efficiency}. Running scripts in the repository recreates reported metrics and regenerates all plots and logs. 
The repository is released under the \textit{Apache License 2.0}.

{
\bibliographystyle{iclr2026_conference}
\bibliography{ref.bib}

@inproceedings{vaswani2017attention,
  title     = {Attention Is All You Need},
  author    = {Vaswani, Ashish and Shazeer, Noam and Parmar, Niki and Uszkoreit, Jakob and Jones, Llion and Gomez, Aidan N. and Kaiser, {\L}ukasz and Polosukhin, Illia},
  booktitle = {Advances in Neural Information Processing Systems (NeurIPS)},
  year      = {2017}
}

@inproceedings{shazeer2019mqa,
  title     = {Fast Transformer Decoding: One Write-Head is All You Need},
  author    = {Shazeer, Noam},
  booktitle = {NeurIPS Workshop on Efficient Natural Language and Speech Processing},
  year      = {2019},
  note      = {Multi-Query Attention (MQA)}
}

@article{wang2024mamba,
  title={The mamba in the llama: Distilling and accelerating hybrid models},
  author={Wang, Junxiong and Paliotta, Daniele and May, Avner and Rush, Alexander and Dao, Tri},
  journal={Advances in Neural Information Processing Systems},
  volume={37},
  pages={62432--62457},
  year={2024}
}

@inproceedings{ainslie2023gqa,
  title     = {Training Generalized Multi-Query Transformer Models from Multi-Head Checkpoints},
  author    = {Ainslie, Joshua and Lee-Thorp, James and De Jong, Michiel and Zemlyanskiy, Yury and Lebr{\'o}n, Federico and Sanghai, Sumit},
  booktitle = {Proceedings of the 2023 Conference on Empirical Methods in Natural Language Processing (EMNLP)},
  year      = {2023}
}

@misc{deepseekv2,
  title        = {DeepSeek-V2: Multi-Head Latent Attention for Economical Inference},
  author       = {{DeepSeek-AI Team}},
  year         = {2024},
  howpublished = {Technical Report},
  url          = {https://github.com/deepseek-ai}
}

@misc{meng2025transmla,
  title        = {TransMLA: Multi-Head Latent Attention Is All You Need},
  author       = {Meng, Fanxu and Yao, Zengwei and Zhang, Muhan},
  year         = {2025},
  howpublished = {Technical Report},
  institution  = {Peking University},
  url          = {https://github.com/fxmeng/TransMLA}
}

@misc{ji2025mha2mla,
  title        = {Towards Economical Inference: Enabling Multi-Head Latent Attention in Transformer-based LLMs},
  author       = {Ji, Tao and Guo, Bin and Wu, Yuanbin and Guo, Qipeng and Shen, Lixing and Chen, Zhan and Qiu, Xipeng and Zhang, Qi and Gui, Tao},
  year         = {2025},
  howpublished = {Technical Report},
  institution  = {Fudan University},
  url          = {https://github.com/JT-Ushio/MHA2MLA}
}

@inproceedings{su2021roformer,
  title     = {RoFormer: Enhanced Transformer with Rotary Position Embedding},
  author    = {Su, Jianlin and Lu, Yu and Pan, Shengfeng and Wen, Bo and Liu, Yunfeng},
  booktitle = {Findings of the Association for Computational Linguistics (ACL Findings)},
  year      = {2021}
}

@inproceedings{hu2022lora,
  title     = {LoRA: Low-Rank Adaptation of Large Language Models},
  author    = {Hu, Edward J. and Shen, Yelong and Wallis, Phillip and Allen-Zhu, Zeyuan and Li, Yuanzhi and Wang, Shean and Wang, Lu and Chen, Weizhu},
  booktitle = {International Conference on Learning Representations (ICLR)},
  year      = {2022}
}

@inproceedings{valipour2023dylora,
  title     = {DyLoRA: Parameter-Efficient Tuning of Pre-trained Models using Dynamic Search-Free Low-Rank Adaptation},
  author    = {Valipour, Mojtaba and Rezagholizadeh, Mehdi and Kobyzev, Ivan and Ghodsi, Ali},
  booktitle = {Proceedings of the 17th Conference of the European Chapter of the ACL (EACL)},
  year      = {2023}
}

@inproceedings{zhang2023adalora,
  title     = {AdaLoRA: Adaptive Budget Allocation for Parameter-Efficient Fine-Tuning},
  author    = {Zhang, Qingru and Chen, Minshuo and Bukharin, Alexander and Karampatziakis, Nikos and He, Pengcheng and Cheng, Yu and Chen, Weizhu and Zhao, Tuo},
  booktitle = {International Conference on Learning Representations (ICLR)},
  year      = {2023}
}

@inproceedings{yang2024coda,
  title     = {CorDA: Context-Oriented Decomposition Adaptation of Large Language Models for Task-Aware Parameter-Efficient Fine-tuning},
  author    = {Yibo Yang and Xiaojie Li and Zhongzhu Zhou and Shuaiwen Leon Song and Jianlong Wu and Liqiang Nie and Bernard Ghanem},
  booktitle = {Advances in Neural Information Processing Systems 37 (NeurIPS)},
  year      = {2024}
}

@article{li2025soco,
  title   = {Optimizing Singular Spectrum for Large Language Model Compression},
  author  = {Dengjie Li and Tiancheng Shen and Yao Zhou and Baisong Yang and Zhongying Liu and Masheng Yang and Bernard Ghanem and Yibo Yang and Yujie Zhong and Ming-Hsuan Yang},
  journal = {CoRR},
  volume  = {abs/2502.15092},
  year    = {2025},
  url     = {https://arxiv.org/abs/2502.15092}
}

@misc{frantar2022gptq,
  title         = {GPTQ: Accurate Post-Training Quantization for Generative Pre-trained Transformers},
  author        = {Elias Frantar and Saleh Ashkboos and Torsten Hoefler and Dan Alistarh},
  year          = {2022},
  eprint        = {2210.17323},
  archivePrefix = {arXiv},
  primaryClass  = {cs.LG},
  note          = {ICLR 2023}
}

@article{eckart1936approximation,
  title   = {The Approximation of One Matrix by Another of Lower Rank},
  author  = {Eckart, Carl and Young, Gale},
  journal = {Psychometrika},
  year    = {1936},
  volume  = {1},
  number  = {3},
  pages   = {211--218}
}

@book{krzanowski2000principles,
  title     = {Principles of Multivariate Analysis: A User's Perspective},
  author    = {Krzanowski, W. J.},
  publisher = {Oxford University Press},
  year      = {2000}
}

@inproceedings{hinton2015distilling,
  title     = {Distilling the Knowledge in a Neural Network},
  author    = {Hinton, Geoffrey and Vinyals, Oriol and Dean, Jeff},
  booktitle = {NeurIPS Deep Learning and Representation Learning Workshop},
  year      = {2015}
}

@article{liu2024deepseek,   title={Deepseek-v3 technical report},   author={Liu, Aixin and Feng, Bei and Xue, Bing and Wang, Bingxuan and Wu, Bochao and Lu, Chengda and Zhao, Chenggang and Deng, Chengqi and Zhang, Chenyu and Ruan, Chong and others},   journal={arXiv preprint arXiv:2412.19437},   year={2024} }

@article{guo2025deepseek,
  title={Deepseek-r1: Incentivizing reasoning capability in llms via reinforcement learning},
  author={Guo, Daya and Yang, Dejian and Zhang, Haowei and Song, Junxiao and Zhang, Ruoyu and Xu, Runxin and Zhu, Qihao and Ma, Shirong and Wang, Peiyi and Bi, Xiao and others},
  journal={arXiv preprint arXiv:2501.12948},
  year={2025}
}

@inproceedings{hassibi1993optimal,
  title={Optimal brain surgeon and general network pruning},
  author={Hassibi, Babak and Stork, David G and Wolff, Gregory J},
  booktitle={IEEE international conference on neural networks},
  pages={293--299},
  year={1993},
  organization={IEEE}
}

@article{wang2024svd,
  title={Svd-llm: Truncation-aware singular value decomposition for large language model compression},
  author={Wang, Xin and Zheng, Yu and Wan, Zhongwei and Zhang, Mi},
  journal={arXiv preprint arXiv:2403.07378},
  year={2024}
}

@article{wang2025dobi,
  title={Dobi-SVD: Differentiable SVD for LLM Compression and Some New Perspectives},
  author={Wang, Qinsi and Ke, Jinghan and Tomizuka, Masayoshi and Chen, Yiran and Keutzer, Kurt and Xu, Chenfeng},
  journal={arXiv preprint arXiv:2502.02723},
  year={2025}
}

@article{yuan2023asvd,
  title={Asvd: Activation-aware singular value decomposition for compressing large language models},
  author={Yuan, Zhihang and Shang, Yuzhang and Song, Yue and Wu, Qiang and Yan, Yan and Sun, Guangyu},
  journal={arXiv preprint arXiv:2312.05821},
  year={2023}
}

@article{Su2021RoPE,
  title     = {RoFormer: Enhanced Transformer with Rotary Position Embedding},
  author    = {Su, Jianlin and Lu, Yu and Pan, Shengfeng and Moudgil, Ahmed and others},
  journal   = {arXiv preprint arXiv:2104.09864},
  year      = {2021}
}

@inproceedings{LeCun1990OBD,
  title     = {Optimal Brain Damage},
  author    = {LeCun, Yann and Denker, John S. and Solla, Sara A.},
  booktitle = {Advances in Neural Information Processing Systems (NeurIPS)},
  year      = {1990}
}

@inproceedings{Hassibi1993OBS,
  title     = {Second Order Derivatives for Network Pruning: Optimal Brain Surgeon},
  author    = {Hassibi, Babak and Stork, David G.},
  booktitle = {Advances in Neural Information Processing Systems (NeurIPS)},
  year      = {1993}
}

@inproceedings{Dong2019HAWQ,
  title     = {HAWQ: Hessian Aware Quantization of Neural Networks with Mixed-Precision},
  author    = {Dong, Zhen and Yao, Zhewei and Gholami, Amir and Mahoney, Michael W. and Keutzer, Kurt},
  booktitle = {Proceedings of the IEEE/CVF International Conference on Computer Vision (ICCV)},
  year      = {2019}
}

@article{kwon2023vllm,
  title   = {Efficient Memory Management for Large Language Model Serving with PagedAttention},
  author  = {Kwon, Woosuk and Li, Zhuohan and Zhuang, Siyuan and Sheng, Ying and Zheng, Lianmin and Yu, Cody Hao and Gonzalez, Joseph E. and Zhang, Hao and Stoica, Ion},
  journal = {arXiv preprint arXiv:2309.06180},
  year    = {2023},
  url     = {https://arxiv.org/abs/2309.06180}
}

@article{touvron2023llama2,
  title   = {Llama 2: Open Foundation and Fine-Tuned Chat Models},
  author  = {Touvron, Hugo and Martin, Louis and Stone, Kevin and Albert, Peter and Almahairi, Amjad and Babaei, Yasmine and others},
  journal = {arXiv preprint arXiv:2307.09288},
  year    = {2023},
  url     = {https://arxiv.org/abs/2307.09288}
}

@article{jiang2023mistral,
  title   = {Mistral 7B},
  author  = {Jiang, Albert Q. and Sablayrolles, Alexandre and Mensch, Arthur and Bamford, Chris and Chaplot, Devendra Singh and de las Casas, Diego and Bressand, Florian and Lengyel, Gianna and Lample, Guillaume and Saulnier, Lucile and Lavaud, L{\'e}lio Renard and Lachaux, Marie-Anne and Stock, Pierre and Le Scao, Teven and Lavril, Thibaut and Wang, Thomas and Lacroix, Timoth{\'e}e and El Sayed, William},
  journal = {arXiv preprint arXiv:2310.06825},
  year    = {2023},
  url     = {https://arxiv.org/abs/2310.06825}
}

@article{shoeybi2019megatron,
  title   = {Megatron-LM: Training Multi-Billion Parameter Language Models Using Model Parallelism},
  author  = {Shoeybi, Mohammad and Patwary, Mostofa and Puri, Raul and LeGresley, Patrick and Casper, Jared and Catanzaro, Bryan},
  journal = {arXiv preprint arXiv:1909.08053},
  year    = {2019},
  url     = {https://arxiv.org/abs/1909.08053}
}

@article{chowdhery2022palm,
  title   = {PaLM: Scaling Language Modeling with Pathways},
  author  = {Chowdhery, Aakanksha and others},
  journal = {arXiv preprint arXiv:2204.02311},
  year    = {2022},
  url     = {https://arxiv.org/abs/2204.02311}
}

@article{wang2020linformer,
  title   = {Linformer: Self-Attention with Linear Complexity},
  author  = {Wang, Sinong and Li, Belinda Z. and Khabsa, Madian and Fang, Han and Ma, Hao},
  journal = {arXiv preprint arXiv:2006.04768},
  year    = {2020},
  url     = {https://arxiv.org/abs/2006.04768}
}

@article{xiong2021nystromformer,
  title   = {Nystr{\"o}mformer: A Nystr{\"o}m-Based Algorithm for Approximating Self-Attention},
  author  = {Xiong, Yunyang and Zeng, Zhanpeng and Chakraborty, Rudrasis and Tan, Mingxing and Fung, Glenn and Li, Yin and Singh, Vikas},
  journal = {arXiv preprint arXiv:2102.03902},
  year    = {2021},
  url     = {https://arxiv.org/abs/2102.03902}
}

@article{yang2024qwen2,
  title   = {Qwen2 Technical Report},
  author  = {Yang, An and Yang, Baosong and Hui, Binyuan and Zheng, Bo and Yu, Bowen and Zhou, Chang and others},
  journal = {arXiv preprint arXiv:2407.10671},
  year    = {2024},
  url     = {https://arxiv.org/abs/2407.10671}
}

@article{geens2025hardwaremla,
  title   = {Hardware-Centric Analysis of DeepSeek's Multi-Head Latent Attention},
  author  = {Geens, Robin and Verhelst, Marian},
  journal = {arXiv preprint arXiv:2506.02523},
  year    = {2025},
  doi     = {10.48550/arXiv.2506.02523},
  url     = {https://arxiv.org/abs/2506.02523}
}

@inproceedings{denil2013predicting,
  title     = {Predicting Parameters in Deep Learning},
  author    = {Denil, Misha and Shakibi, Babak and Dinh, Laurent and Ranzato, Marc'Aurelio and de Freitas, Nando},
  booktitle = {Advances in Neural Information Processing Systems 26 (NeurIPS 2013)},
  year      = {2013},
  url       = {https://papers.nips.cc/paper/5025-predicting-parameters-in-deep-learning}
}

@inproceedings{denton2014exploiting,
  title     = {Exploiting Linear Structure Within Convolutional Networks for Efficient Evaluation},
  author    = {Denton, Emily L. and Zaremba, Wojciech and Bruna, Joan and LeCun, Yann and Fergus, Rob},
  booktitle = {Advances in Neural Information Processing Systems 27 (NeurIPS 2014)},
  year      = {2014},
  url       = {https://papers.nips.cc/paper/5544-exploiting-linear-structure-within-convolutional-networks-for-efficient-evaluation.pdf}
}

@inproceedings{sainath2013lowrank,
  title     = {Low-rank Matrix Factorization for Deep Neural Network Training with High-dimensional Output Targets},
  author    = {Sainath, Tara N. and Kingsbury, Brian and Sindhwani, Vikas and Arisoy, Ebru and Ramabhadran, Bhuvana},
  booktitle = {Proc. IEEE Int. Conf. on Acoustics, Speech and Signal Processing (ICASSP)},
  address   = {Vancouver, Canada},
  pages     = {6655--6659},
  publisher = {IEEE},
  year      = {2013},
  doi       = {10.1109/ICASSP.2013.6638949}
}

@article{sakaguchi2021winogrande,
  title={Winogrande: An adversarial winograd schema challenge at scale},
  author={Sakaguchi, Keisuke and Bras, Ronan Le and Bhagavatula, Chandra and Choi, Yejin},
  journal={Communications of the ACM},
  volume={64},
  number={9},
  pages={99--106},
  year={2021},
  publisher={ACM New York, NY, USA}
}

@article{C4,
  title={Exploring the limits of transfer learning with a unified text-to-text transformer},
  author={Raffel, Colin and Shazeer, Noam and Roberts, Adam and Lee, Katherine and Narang, Sharan and Matena, Michael and Zhou, Yanqi and Li, Wei and Liu, Peter J},
  journal={Journal of machine learning research},
  volume={21},
  number={140},
  pages={1--67},
  year={2020}
}

@misc{ALPACA,
  author = {Rohan Taori and Ishaan Gulrajani and Tianyi Zhang and Yann Dubois and Xuechen Li and Carlos Guestrin and Percy Liang and Tatsunori B. Hashimoto },
  title = {Stanford Alpaca: An Instruction-following LLaMA model},
  year = {2023},
  publisher = {GitHub},
  journal = {GitHub repository},
  howpublished = {\url{https://github.com/tatsu-lab/stanford_alpaca}},
}

@article{PTB, title = "Building a Large Annotated Corpus of {E}nglish: The {P}enn {T}reebank", author = "Marcus, Mitchell P. and Santorini, Beatrice and Marcinkiewicz, Mary Ann", journal = "Computational Linguistics", volume = "19", number = "2", year = "1993", url = "https://www.aclweb.org/anthology/J93-2004", pages = "313--330", }

@misc{WIKITEXT2,
      title={Pointer Sentinel Mixture Models},
      author={Stephen Merity and Caiming Xiong and James Bradbury and Richard Socher},
      year={2016},
      eprint={1609.07843},
      archivePrefix={arXiv},
      primaryClass={cs.CL}
}

@article{lai2017race,
  title={Race: Large-scale reading comprehension dataset from examinations},
  author={Lai, Guokun and Xie, Qizhe and Liu, Hanxiao and Yang, Yiming and Hovy, Eduard},
  journal={arXiv preprint arXiv:1704.04683},
  year={2017}
}

@article{mihaylov2018can,
  title={Can a suit of armor conduct electricity? a new dataset for open book question answering},
  author={Mihaylov, Todor and Clark, Peter and Khot, Tushar and Sabharwal, Ashish},
  journal={arXiv preprint arXiv:1809.02789},
  year={2018}
}

@inproceedings{bisk2020piqa,
  title={Piqa: Reasoning about physical commonsense in natural language},
  author={Bisk, Yonatan and Zellers, Rowan and Gao, Jianfeng and Choi, Yejin and others},
  booktitle={Proceedings of the AAAI conference on artificial intelligence},
  volume={34},
  pages={7432--7439},
  year={2020}
}

@article{hendrycks2020measuring,
  title={Measuring massive multitask language understanding},
  author={Hendrycks, Dan and Burns, Collin and Basart, Steven and Zou, Andy and Mazeika, Mantas and Song, Dawn and Steinhardt, Jacob},
  journal={arXiv preprint arXiv:2009.03300},
  year={2020}
}

@article{zellers2019hellaswag,
  title={Hellaswag: Can a machine really finish your sentence?},
  author={Zellers, Rowan and Holtzman, Ari and Bisk, Yonatan and Farhadi, Ali and Choi, Yejin},
  journal={arXiv preprint arXiv:1905.07830},
  year={2019}
}

@article{clark2018think,
  title={Think you have solved question answering? try arc, the ai2 reasoning challenge},
  author={Clark, Peter and Cowhey, Isaac and Etzioni, Oren and Khot, Tushar and Sabharwal, Ashish and Schoenick, Carissa and Tafjord, Oyvind},
  journal={arXiv preprint arXiv:1803.05457},
  year={2018}
}

@misc{eval-harness,
  author       = {Gao, Leo and Tow, Jonathan and Abbasi, Baber and Biderman, Stella and Black, Sid and DiPofi, Anthony and Foster, Charles and Golding, Laurence and Hsu, Jeffrey and Le Noac'h, Alain and Li, Haonan and McDonell, Kyle and Muennighoff, Niklas and Ociepa, Chris and Phang, Jason and Reynolds, Laria and Schoelkopf, Hailey and Skowron, Aviya and Sutawika, Lintang and Tang, Eric and Thite, Anish and Wang, Ben and Wang, Kevin and Zou, Andy},
  title        = {The Language Model Evaluation Harness},
  month        = 07,
  year         = 2024,
  publisher    = {Zenodo},
  version      = {v0.4.3},
  doi          = {10.5281/zenodo.12608602},
  url          = {https://zenodo.org/records/12608602}
}

@inproceedings{hua2022tfwsvd,
  title     = {Numerical Optimizations for Weighted Low-rank Estimation on Language Model},
  author    = {Hua, Ting and Hsu, Yen-Chang and Wang, Felicity and Lou, Qian and Shen, Yilin and Jin, Hongxia},
  booktitle = {Proceedings of the 2022 Conference on Empirical Methods in Natural Language Processing (EMNLP)},
  pages     = {1404--1416},
  publisher = {Association for Computational Linguistics},
  year      = {2022},
  url       = {https://aclanthology.org/2022.emnlp-main.91.pdf}
}

@inproceedings{wang2025svdllmv2,
  title     = {SVD-LLM V2: Optimizing Singular Value Truncation for Large Language Model Compression},
  author    = {Wang, Xin and Alam, Samiul and Wan, Zhongwei and Shen, Hui and Zhang, Mi},
  booktitle = {Proceedings of the 2025 Conference of the North American Chapter of the Association for Computational Linguistics: Human Language Technologies (Volume 1: Long Papers)},
  pages     = {4287--4296},
  address   = {Albuquerque, New Mexico},
  publisher = {Association for Computational Linguistics},
  year      = {2025},
  url       = {https://aclanthology.org/2025.naacl-long.217.pdf}
}

@inproceedings{shaw2018relative,
  title     = {Self-Attention with Relative Position Representations},
  author    = {Shaw, Peter and Uszkoreit, Jakob and Vaswani, Ashish},
  booktitle = {Proceedings of NAACL-HLT 2018, Volume 2 (Short Papers)},
  pages     = {464--468},
  address   = {New Orleans, Louisiana},
  publisher = {Association for Computational Linguistics},
  year      = {2018},
  url       = {https://aclanthology.org/N18-2074/},
  doi       = {10.18653/v1/N18-2074}
}

@inproceedings{he2021deberta,
  title     = {DeBERTa: Decoding-Enhanced BERT with Disentangled Attention},
  author    = {He, Pengcheng and Liu, Xiaodong and Gao, Jianfeng and Chen, Weizhu},
  booktitle = {International Conference on Learning Representations (ICLR)},
  year      = {2021},
  url       = {https://iclr.cc/virtual/2021/poster/2562}
}

@article{raffel2020t5,
  title   = {Exploring the Limits of Transfer Learning with a Unified Text-to-Text Transformer},
  author  = {Raffel, Colin and Shazeer, Noam and Roberts, Adam and Lee, Katherine and Narang, Sharan and Matena, Michael and Zhou, Yanqi and Li, Wei and Liu, Peter J.},
  journal = {Journal of Machine Learning Research},
  volume  = {21},
  number  = {140},
  pages   = {1--67},
  year    = {2020},
  url     = {https://jmlr.org/papers/v21/20-074.html}
}

@article{press2021alibi,
  title   = {Train Short, Test Long: Attention with Linear Biases Enables Input Length Extrapolation},
  author  = {Press, Ofir and Smith, Noah A. and Lewis, Mike},
  journal = {arXiv preprint arXiv:2108.12409},
  year    = {2021},
  url     = {https://arxiv.org/abs/2108.12409}
}

@article{sun2022xpos,
  title   = {A Length-Extrapolatable Transformer},
  author  = {Sun, Yutao and Dong, Li and Patra, Barun and Ma, Shuming and Huang, Shaohan and Benhaim, Alon and Chaudhary, Vishrav and Song, Xia and Wei, Furu},
  journal = {arXiv preprint arXiv:2212.10554},
  year    = {2022},
  url     = {https://arxiv.org/abs/2212.10554}
}

@article{chen2023pi,
  title   = {Extending Context Window of Large Language Models via Positional Interpolation},
  author  = {Chen, Shouyuan and Wong, Sherman and Chen, Liangjian and Tian, Yuandong},
  journal = {arXiv preprint arXiv:2306.15595},
  year    = {2023},
  url     = {https://arxiv.org/abs/2306.15595}
}

@article{peng2023yarn,
  title   = {YaRN: Efficient Context Window Extension of Large Language Models},
  author  = {Peng, Bowen and Quesnelle, Jeffrey and Fan, Honglu and Shippole, Enrico},
  journal = {arXiv preprint arXiv:2309.00071},
  year    = {2023},
  url     = {https://arxiv.org/abs/2309.00071}
}

@article{ding2024longrope,
  title   = {LongRoPE: Extending LLM Context Window Beyond 2 Million Tokens},
  author  = {Ding, Yiran and Zhang, Li Lyna and Zhang, Chengruidong and Xu, Yuanyuan and Shang, Ning and Xu, Jiahang and Yang, Fan and Yang, Mao},
  journal = {arXiv preprint arXiv:2402.13753},
  year    = {2024},
  url     = {https://arxiv.org/abs/2402.13753}
}

@inproceedings{kazemnejad2023impact,
  title     = {The Impact of Positional Encoding on Length Generalization in Transformers},
  author    = {Kazemnejad, Amirhossein and Padhi, Inkit and Natesan Ramamurthy, Karthikeyan and Das, Payel and Reddy, Siva},
  booktitle = {Advances in Neural Information Processing Systems (NeurIPS)},
  year      = {2023},
  url       = {https://proceedings.neurips.cc/paper_files/paper/2023/file/4e85362c02172c0c6567ce593122d31c-Paper-Conference.pdf}
}

@article{li2025xecoMLA,
  title   = {X-EcoMLA: Upcycling Pre-Trained Attention into MLA for Efficient and Extreme KV Compression},
  author  = {Li, Guihong and Rezagholizadeh, Mehdi and Yang, Mingyu and Appia, Vikram and Barsoum, Emad},
  journal = {arXiv preprint arXiv:2503.11132},
  year    = {2025},
  url     = {https://arxiv.org/abs/2503.11132}
}

@article{yang2025zebrallama,
  title   = {Zebra-Llama: Towards Extremely Efficient Hybrid Models},
  author  = {Yang, Mingyu and Rezagholizadeh, Mehdi and Li, Guihong and Appia, Vikram and Barsoum, Emad},
  journal = {arXiv preprint arXiv:2505.17272},
  year    = {2025},
  url     = {https://arxiv.org/abs/2505.17272}
}

@article{chang2024palu,
  title   = {Palu: Compressing KV-Cache with Low-Rank Projection},
  author  = {Chang, Chi-Chih and Lin, Wei-Cheng and Lin, Chien-Yu and Chen, Chong-Yan and Hu, Yu-Fang and Wang, Pei-Shuo and Huang, Ning-Chi and Ceze, Luis and Abdelfattah, Mohamed S. and Wu, Kai-Chiang},
  journal = {arXiv preprint arXiv:2407.21118},
  year    = {2024},
  url     = {https://arxiv.org/abs/2407.21118}
}

@article{reallm2025,
  title     = {ReALLM: A General Framework for LLM Compression and Fine-Tuning},
  author    = {Leconte, Louis and Bedin, Lisa and Nguyen, Van Minh and Moulines, Eric},
  journal   = {arXiv preprint arXiv:2405.13155},
  year      = {2024},
  url       = {https://arxiv.org/abs/2405.13155}
}

@inproceedings{dao2022flashattention,
  author    = {Tri Dao and Daniel Y. Fu and Stefano Ermon and Atri Rudra and Christopher R{\'e}},
  title     = {FlashAttention: Fast and Memory-Efficient Exact Attention with {IO}-Awareness},
  booktitle = {Advances in Neural Information Processing Systems (NeurIPS)},
  year      = {2022},
  url       = {https://proceedings.neurips.cc/paper_files/paper/2022/hash/67d57c32e20fd0a7a302cb81d36e40d5-Abstract-Conference.html}
}

@article{dao2023flashattention2,
  author  = {Tri Dao},
  title   = {FlashAttention-2: Faster Attention with Better Parallelism and Work Partitioning},
  journal = {arXiv preprint arXiv:2307.08691},
  year    = {2023},
  url     = {https://arxiv.org/abs/2307.08691}
}

@inproceedings{xiao2023smoothquant,
  author    = {Guangxuan Xiao and Ji Lin and Mickael Seznec and Hao Wu and Julien Demouth and Song Han},
  title     = {SmoothQuant: Accurate and Efficient Post-Training Quantization for Large Language Models},
  booktitle = {Proceedings of the 40th International Conference on Machine Learning (ICML)},
  year      = {2023},
  publisher = {PMLR},
  url       = {https://proceedings.mlr.press/v202/xiao23c.html}
}

@inproceedings{lin2024awq,
  author    = {Ji Lin and Jiaming Tang and Haotian Tang and Shang Yang and Wei{-}Ming Chen and Wei{-}Chen Wang and Guangxuan Xiao and Xingyu Dang and Chuang Gan and Song Han},
  title     = {AWQ: Activation-aware Weight Quantization for On-Device {LLM} Compression and Acceleration},
  booktitle = {MLSys},
  year      = {2024},
  url       = {https://proceedings.mlsys.org/paper_files/paper/2024/file/42a452cbafa9dd64e9ba4aa95cc1ef21-Paper-Conference.pdf}
}

@inproceedings{tseng2024quipsharp,
  author    = {Albert Tseng and Jerry Chee and Qingyao Sun and Volodymyr Kuleshov and Christopher De Sa},
  title     = {QuIP\#: Even Better {LLM} Quantization with Hadamard Incoherence and Lattice Codebooks},
  booktitle = {Proceedings of the 41st International Conference on Machine Learning (ICML)},
  year      = {2024},
  publisher = {PMLR},
  url       = {https://proceedings.mlr.press/v235/tseng24a.html}
}

@inproceedings{hooper2024kvquant,
  author    = {Coleman Hooper and Sehoon Kim and Hiva Mohammadzadeh and Michael W. Mahoney and Yakun Sophia Shao and Kurt Keutzer and Amir Gholami},
  title     = {KVQuant: Towards 10 Million Context Length {LLM} Inference with {KV} Cache Quantization},
  booktitle = {Advances in Neural Information Processing Systems (NeurIPS)},
  year      = {2024},
  url       = {https://proceedings.neurips.cc/paper_files/paper/2024/hash/028fcbcf85435d39a40c4d61b42c99a4-Abstract-Conference.html}
}

@article{liu2024kivi,
  author  = {Zirui Liu and Jiayi Yuan and Hongye Jin and Shaochen Zhong and Zhaozhuo Xu and Vladimir Braverman and Beidi Chen and Xia Hu},
  title   = {KIVI: A Tuning-Free Asymmetric 2bit Quantization for {KV} Cache},
  journal = {arXiv preprint arXiv:2402.02750},
  year    = {2024},
  url     = {https://arxiv.org/abs/2402.02750}
}

@inproceedings{fang2026rethinking,
  title={Rethinking Video-language Model From the Language Input Perspective},
  author={Fang, Xiang and Fang, Wanlong and Wang, Changshuo and Qu, Xiaoye and Liu, Daizong},
  booktitle={Proceedings of the AAAI Conference on Artificial Intelligence},
  year={2026}
}

@inproceedings{fang2026towards,
  title={Towards Unified Vision-Language Models With Incomplete Multi-Modal Inputs},
  author={Fang, Xiang and Fang, Wanlong and Wang, Changshuo and Tang, Keke and Liu, Daizong and Wang, Siyi and Ji, Wei},
  booktitle={Proceedings of the AAAI Conference on Artificial Intelligence},
  year={2026}
}

@article{ma2025veomni,
  title={VeOmni: Scaling Any Modality Model Training with Model-Centric Distributed Recipe Zoo},
  author={Ma, Qianli and Zheng, Yaowei and Shi, Zhelun and Zhao, Zhongkai and Jia, Bin and Huang, Ziyue and Lin, Zhiqi and Li, Youjie and Yang, Jiacheng and Peng, Yanghua and others},
  journal={arXiv preprint arXiv:2508.02317},
  year={2025}
}
}

\clearpage
\onecolumn
\appendix
\begin{center}
	\textbf{CARE: Covariance-Aware and Rank-Enhanced Decomposition for Enabling Multi-Head Latent Attention in LLMs\\ \vspace{8pt} \textit{Supplementary Material}}
\end{center}

\counterwithin{figure}{section}
\counterwithin{table}{section}

\section{Large Language Models Usage}
We used a large language model - ChatGPT (GPT-5 thinking) solely for grammar and spelling edits to author-written text. We used Claude Code to assist code writing. The tool did not generate scientific content, design experiments, analyze data, or select citations, and therefore did not contribute at the level of a contributing author. All edits were reviewed and approved by the authors, who take full responsibility for the final manuscript.

\section{Recall: KV-Parity Mapping}
\label{subsec:kvparity}
Grouped-Query Attention (GQA) reduces KV-cache memory by letting multiple heads share the same key--value projection. To convert GQA to Multi-Head Latent Attention (MLA) \emph{without} increasing the KV budget, we enforce \textbf{KV parity}. Consider a GQA layer with $n_h$ heads of size $d_h$ with total multi-head hidden size $D = n_h d_h$, we split $n_h$ heads split into $g_h$ groups, where each group contains $\frac{n_h}{g_h}$ heads. The layer $l$ uses $W_Q^{(l)} \in \mathbb{R}^{D \times (n_h d_h)}$ and grouped $W_K^{(l)}, W_V^{(l)} \in \mathbb{R}^{D \times (g_h d_h)}$. We conceptually \emph{replicate} each group's $W_K^{(l)}$ and $W_V^{(l)}$ across its $\frac{n_h}{g_h}$ members to form the full-size
$\widetilde{W}_K^{(l)},\widetilde{W}_V^{(l)} \in \mathbb{R}^{D \times (n_h d_h)}$ (no need to materialize in code). This demonstrates that the GQA method can be reduced to MLA by removing the repeated head blocks~\citep{meng2025transmla}.
Therefore, we set MLA's latent rank to match GQA's per-token KV width:
\begin{equation}
r \;=\; g_h \, d_h
\label{eq:kvparity}
\end{equation}

\section{Proof of Maximum Energy SVD truncation}
\label{thm_proof}
We have the following proposition: Let $A\in\mathbb{R}^{m\times n}$ have singular value decomposition (SVD) $A=U\Sigma V^\top$, where $\Sigma=\mathrm{diag}(\sigma_1,\dots,\sigma_p)$, $p=\min\{m,n\}$, and $\sigma_1\ge \cdots \ge \sigma_p\ge 0$. For $1\le r<p$, let $\Sigma_r=\mathrm{diag}(\sigma_1,\dots,\sigma_r,0,\dots,0)$ and $A_r:=U\Sigma_r V^\top$. Then
\[
\|A-A_r\|_F^2 \;=\; \sum_{i=r+1}^{p}\sigma_i^2
\quad\text{and}\quad
\|A-A_r\|_F \;=\; \min_{\operatorname{rank}(X)\le r}\|A-X\|_F,
\]
with the unique minimizers (when $\sigma_r>\sigma_{r+1}$) given by $X=A_r$ \citep{eckart1936approximation}.

\begin{proof}
The Frobenius norm is unitarily invariant, so for any $X$ with $\operatorname{rank}(X)\le r$,
\[
\|A-X\|_F \;=\; \|\,U^\top(A-X)V\,\|_F \;=\; \|\Sigma - Y\|_F,
\quad\text{where } Y:=U^\top X V \text{ and }\operatorname{rank}(Y)\le r.
\]
Expand the square via the Frobenius inner product $\langle M,N\rangle:=\mathrm{trace}(M^\top N)$:
\[
\|\Sigma-Y\|_F^2 \;=\; \|\Sigma\|_F^2 + \|Y\|_F^2 - 2\langle \Sigma, Y\rangle.
\]
Let $s_1(Y)\ge \cdots \ge s_p(Y)\ge 0$ be the singular values of $Y$ (so $s_i(Y)=0$ for $i>r$). By von Neumann's trace inequality,
\[
\langle \Sigma, Y\rangle \;\le\; \sum_{i=1}^{p}\sigma_i\,s_i(Y) \;=\; \sum_{i=1}^{r}\sigma_i\,s_i(Y),
\]
and $\|Y\|_F^2=\sum_{i=1}^{p}s_i(Y)^2=\sum_{i=1}^{r}s_i(Y)^2$. Therefore:
\[
\|\Sigma-Y\|_F^2
\;\ge\; \sum_{i=1}^{p}\sigma_i^2 + \sum_{i=1}^{r}s_i(Y)^2 - 2\sum_{i=1}^{r}\sigma_i s_i(Y)
\;=\; \sum_{i=1}^{r}\bigl(\sigma_i - s_i(Y)\bigr)^2 + \sum_{i=r+1}^{p}\sigma_i^2
\;\ge\; \sum_{i=r+1}^{p}\sigma_i^2.
\]
This lower bound is attained by taking $Y=\Sigma_r$, i.e.\ $X=U\Sigma_r V^\top = A_r$, for which:
\[
\|A-A_r\|_F^2 \;=\; \|\Sigma-\Sigma_r\|_F^2 \;=\; \sum_{i=r+1}^{p}\sigma_i^2.
\]
Thus $A_r$ is a best rank-$r$ approximation in Frobenius norm, and the minimum value is the squared $\ell_2$-tail of the singular values. Uniqueness follows when $\sigma_r>\sigma_{r+1}$ since then any other minimizer must share the top $r$ singular subspaces with $A$.
\end{proof}
Note that the idea and proof of the theorem above follows essentially the same idea as Eckart--Young--Mirsky in \citet{eckart1936approximation}.

\paragraph{Relation to Sec.~\ref{subsec:ranksched}.}
The theorem justifies the use of squared singular values in Sec.~\ref{subsec:ranksched}: for a single layer/matrix, the marginal reduction in Frobenius residual from adding the next rank is governed by the next squared singular value. Sec.~\ref{subsec:ranksched} further divides this local gain by the remaining residual energy of the same layer, which normalizes layers with different spectral scales and makes their priorities comparable under a shared global budget. This normalization changes the cross-layer comparison, but preserves the theorem's local interpretation that larger squared singular values remove more residual energy.

\section{Derivation of SVD target function}
\label{derivation}
Let $\Delta W := W-\widehat W$ and $C=\tfrac{1}{N}\sum_{b=1}^N X_b^\top X_b$. We claim that:
\[
\frac{1}{N}\sum_{b=1}^N \|X_b \Delta W\|_F^2
= \|\sqrt{C}\Delta W\|_F^2 .
\]

\begin{proof}
By the Frobenius–trace identity $\|A\|_F^2=\mathrm{Tr}(A^\top A)$,
\[
\frac{1}{N}\sum_{b=1}^N \|X_b \Delta W\|_F^2
= \frac{1}{N}\sum_{b=1}^N \mathrm{Tr}\!\big( (X_b \Delta W)^\top (X_b \Delta W) \big)
= \frac{1}{N}\sum_{b=1}^N \mathrm{Tr}\!\big( \Delta W^\top X_b^\top X_b \Delta W \big).
\]
Using linearity of trace and the definition of $C$,
\[
\frac{1}{N}\sum_{b=1}^N \|X_b \Delta W\|_F^2
= \mathrm{Tr}\!\big( \Delta W^\top C \Delta W \big).
\]
We can show that $C\succeq 0$ since for any vector $v$, we have:
\[v^\top C v = \tfrac{1}{N}\sum_{b=1}^N v^\top X_b^\top X_b v = \tfrac{1}{N}\sum_{b=1}^N \|X_b v\|_2^2 \ge 0
\]
Then we write $C=C^{1/2}C^{1/2}$ to get:
\[
\mathrm{Tr}\!\big( \Delta W^\top C \Delta W \big)
= \mathrm{Tr}\!\big( \Delta W^\top C^{1/2} C^{1/2} \Delta W \big)
= \mathrm{Tr}\!\big( (C^{1/2}\Delta W)^\top (C^{1/2}\Delta W) \big)
= \|C^{1/2}\Delta W\|_F^2 .
\]
\end{proof}

\section{Definition of Covariance}
\label{covariance def}
Let $x\in\mathbb{R}^D$ be the activation (feature) vector of a single token. The population covariance is:
\[
\operatorname{Cov}[x] \;=\; \mathbb{E}\!\left[(x-\mu)(x-\mu)^\top\right],
\qquad
\mu \;=\; \mathbb{E}[x].
\]
Given samples $\{x_i\}_{i=1}^N$, the empirical mean and covariance are:
\[
\hat\mu \;=\; \frac{1}{N}\sum_{i=1}^N x_i,
\qquad
\widehat{\operatorname{Cov}}[x] \;=\; \frac{1}{N}\sum_{i=1}^N (x_i-\hat\mu)(x_i-\hat\mu)^\top.
\]
If rows of $X\in\mathbb{R}^{N\times D}$ stack the samples and $X_c = X - \mathbf{1}\hat\mu^\top$, then $\widehat{\operatorname{Cov}}[x] = \tfrac{1}{N} X_c^\top X_c$.
(\emph{Unbiased} version uses $1/(N-1)$ instead of $1/N$.)

\section{Related Work}
\label{other_related}
\paragraph{KV Management}
Serving throughput is often bounded by how the KV cache is organized and moved across memory. \emph{PagedAttention} (vLLM) treats KV as pageable blocks to avoid internal/external fragmentation and enable sharing across sequences, improving utilization under dynamic batching~\citep{kwon2023vllm}. Orthogonally, \emph{FlashAttention} reduces HBM traffic with an IO-aware tiling of exact attention, and \emph{FlashAttention-2} further improves parallelism and work partitioning for higher FLOPs utilization~\citep{dao2022flashattention,dao2023flashattention2}. These system/kernel directions are complementary to architectural changes (e.g., GQA/MLA) and to post-hoc reparameterizations, since better KV layout and IO scheduling directly translate into larger effective batch sizes at a fixed memory budget.

\paragraph{Quantization}
Quantization provides an orthogonal compression path to low-rank methods and can be combined with MLA/GQA conversions. For \emph{weights/activations}, \emph{SmoothQuant} migrates activation outliers into weights to enable practical W8A8 PTQ on large models~\citep{xiao2023smoothquant}. \emph{AWQ} protects a small set of salient channels via activation-aware scaling, delivering strong 4-bit weight-only PTQ with hardware-friendly kernels~\citep{lin2024awq}. \emph{QuIP\#} pushes extreme regimes ($\leq$4-bit) using randomized Hadamard incoherence and lattice codebooks, with state-of-the-art results at low bit-rates~\citep{tseng2024quipsharp}. For the \emph{KV cache}, \emph{KVQuant} (NeurIPS’24) introduces pre-RoPE key quantization, sensitivity-aware non-uniform datatypes, and per-vector dense/sparse schemes to sustain long-context inference~\citep{hooper2024kvquant}, while \emph{KIVI} shows tuning-free 2-bit asymmetric KV quantization with favorable throughput/memory trade-offs~\citep{liu2024kivi}. Together, these methods form a toolbox that is largely complementary to low-rank latent caching.

\paragraph{RoPE and Positional Encodings}
Positional design strongly affects length generalization and conversion stability. RoPE’s complex-valued rotary formulation remains the default in many LLMs~\citep{su2021roformer}. Alternatives include relative positions~\citep{shaw2018relative}, T5’s learned relative bias and DeBERTa’s disentangled content/position attention~\citep{raffel2020t5,he2021deberta}, and ALiBi’s linear distance bias for train-short/test-long extrapolation~\citep{press2021alibi}. Within the RoPE family, window-extension strategies modify scaling or spectra to stabilize extrapolation, such as XPOS’s multiplicative stabilization, Position Interpolation, YaRN, and very long-window \emph{LongRoPE}~\citep{sun2022xpos,chen2023pi,peng2023yarn,ding2024longrope}. Systematic comparisons further show that the chosen positional scheme materially impacts length generalization~\citep{kazemnejad2023impact}, motivating careful treatment (e.g., partial-RoPE or mixed strategies) during architectural realignments.

\section{Discussion}~\label{sec:discussion}
Across two GQA backbones and diverse tasks, \textbf{CARE}---\underline{C}ovariance-\underline{A}ware and \underline{R}ank-\underline{E}nhanced decomposition---enables MLA migration under \emph{KV-parity} with accuracy and long-context robustness on par with (or better than) stronger baselines. CARE preserves the throughput/memory advantages of MLA while mitigating the activation drift observed with weight-only SVD.

\paragraph{Rank allocation matters.}
Uniform or purely energy-based rank policies overlook the \emph{weighted} spectral concentration that emerges after covariance/curvature preconditioning. CARE’s \emph{Adjusted-Rank} uses a water-filling allocation over weighted singular spectra, honoring a global KV constraint while allocating capacity to the layers and directions that matter most.

\paragraph{Complexity}
The dominant conversion costs are per-layer covariance estimation $\mathcal{O}(N D^2)$ (with small $N$) and truncated SVD on $\sqrt{C}\widetilde{W}\in\mathbb{R}^{D\times(n_h d_h)}$ at $\mathcal{O}(D\,(n_h d_h)\,r)$ using randomized SVD. Layers can be processed sequentially; for each layer, $C=\tfrac{1}{N}\sum_{b = 1}^N X_b^\top X_b$ can all be kept on CPU. At inference, MLA incurs light extra matvecs by $W^{b}$ while reducing KV-cache width from $n_h d_h$ (MHA) or $g_h d_h$ (GQA) down to $r=g_h d_h$ (MLA). CARE is orthogonal to quantization and sparsification and compatible with MLA kernels~\citep{deepseekv2}.

\paragraph{Compatibility with MLA migration.}
CARE complements recent MLA conversions \citep{ji2025mha2mla,meng2025transmla} and plays well with partial-RoPE \citep{Su2021RoPE}: removing rotations on least-contributive subspaces further stabilizes long-context behavior when combined with activation/curvature-aware objectives.

\paragraph{Limitations.}
(i) \emph{Statistics freshness}: CARE requires small calibration passes; pronounced domain shift may need refreshed covariance/curvature. 
(ii) \emph{Diagonal curvature}: practicality favors diagonal proxies; structured approximations (e.g., Kronecker-factored) may yield further gains. 
(iii) \emph{Extreme compression}: at very low ranks, information bottlenecks dominate and further SFT can be necessary. 
(iv) \emph{Orthogonality to quantization/eviction}: CARE does not yet co-optimize KV quantization and cache eviction policies.
(v) \emph{Kernel support}: no dedicated MLA kernel currently supports per-layer dynamic ranks, making end-to-end latency benchmarking impractical; we therefore report only theoretical KV-cache savings rather than wall-clock speedups.

\paragraph{Broader impact and future work.}
CARE suggests a general recipe for post-training architectural migrations: align the objective to where errors manifest and distribute capacity by curvature-weighted signal. 
Promising directions include applying covariance-aware low-rank decomposition to vision-language and multimodal architectures~\citep{fang2026rethinking,fang2026towards}, where attention modules face analogous KV-cache bottlenecks, as well as data-free calibration, structured curvature, and dynamic rank schedules that adapt latent capacity with context length while maintaining KV-parity. 

Apart from that, our Covariance-weighted SVD initialization minimizes the activation loss at each layer, but our true goal is to preserve the output of the model, which is next-token predictions. We may therefore cast low-rank compression as directly minimizing the sequence loss produced by the compressed (student) model under a fixed KV budget.

\section{Hyper-parameter Selection}
\label{hyper}
All experiments were conducted on servers equipped with NVIDIA H100 80 GB GPUs paired with dual Intel Xeon Platinum 8462Y+ processors (2 × 32-core sockets, 64 cores total) and approximately 2 TB of RAM. \\ 

All hyper-parameters are shown as below:

\begin{itemize}
    \item \textbf{Model Configuration}: Base model: Meta; Precision: float16, Sequence length: 8-2048 tokens, Covariance samples: 8-2048.
    \item \textbf{MLA Rank Settings}: Default rank: 256, Min rank: 64, Max rank: 1024, Uniform allocation: True/False, K/V projection ranks: 256 each.
    \item \textbf{CARE Parameters}: Initialization method: CARE, Damping factor (percdamp): 0.01, Cholesky decomposition: False, Activation order: False.
    \item \textbf{Evaluation Datasets}: Multi-task benchmarks including WikiText (perplexity), ARC-Challenge/Easy (reasoning), HellaSwag (commonsense), PIQA (physical reasoning), MMLU (knowledge), OpenBookQA, RACE (reading), WinoGrande (coreference).
    \item \textbf{Generation Settings}: Max new tokens: 512-128000, Temperature: 0.6-0.7, Top-p sampling: 0.9, Sampling strategy: Nucleus sampling with temperature control.
    \item \textbf{System Configuration}: GPU memory free threshold (minimal GPU resources to run experiments) : 2048 MB, Parallel GPUs: 1-8 devices, Batch size: Dynamic adjustment, Random seed: [42, 17, 26, 103, 21, 59, 134, 8, 24, 99].
    \item \textbf{Covaraince Computation}: Dataset: C4/Ptb/Wikitext/Alpaca/ARC/ARE/MMLU instruction-following, Sample size: 8-2048, Sequence processing: 8-2048 token windows, 

    \item \textbf{Random Seed and Learning Rate} All experimental results are average over \textbf{10} random seeds and we choose the best from \textbf{3} learning rates.

    \item \textbf{Training Framework} All experiments were conducted using the \texttt{VeOmni} framework~\citep{ma2025veomni} for fine-tuning CARE and TransMLA models.

    \item \textbf{Learning rate:} We choose best learning rate of $2 \times 10^{-6}$ with linear warmup over the first 0.001\% training steps.

    \item \textbf{Optimizer and Schedulers:} We sweep $\text{LR}\in\{1\!\times\!10^{-6},\,1\!\times\!10^{-5},\,5\!\times\!10^{-5},\,1\!\times\!10^{-4},\,5\!\times\!10^{-4}\}$, and select the best within the first $0.1$B tokens. We use AdamW, weight\_decay$=0.01$, cosine decay with \texttt{lr\_warmup\_ratio}$=0.005$ and \texttt{lr\_decay\_ratio}$=1.0$. 
    
    \item \textbf{Batch size:} Global effective batch size of 64 tokens per update step, accumulated across devices.
    
    \item \textbf{Precision:} bfloat16 mixed precision was enabled to reduce memory footprint and improve throughput.
    
    \item \textbf{Max sequence length:} Input sequences were truncated or padded to a length of 512-128000 tokens.
    
    \item \textbf{Training epochs:} Each experiment was trained for the number of pre-set tokens.
\end{itemize}

\section{Supplementary Results}
\label{app:res}

\subsection{Detailed Zero-shot Tables}
\label{app:detailed-zero-shot}
These tables report the full zero-shot results for each model and calibration dataset combination, using the same metric layout as Tab.~\ref{tab:1}. In addition to the main-text variants (\textbf{CARE-U} and \textbf{CARE-E}, which use $\sqrt{C}$-weighted SVD with uniform and energy-aware rank allocation, respectively), we include \textbf{CARE-C-based-U} and \textbf{CARE-C-based-E}, which replace $\sqrt{C}$ with $C$ directly as the covariance weighting; see Sec.~\ref{app:sqrtc-vs-c} for a detailed comparison of the two formulations.

\subsubsection{Llama-3.1-8B-Instruct-Alpaca}
\begingroup
  \setlength{\tabcolsep}{3.2pt}
  \renewcommand{\arraystretch}{1.08}
  \scriptsize
  \setlength{\LTleft}{\fill}
  \setlength{\LTright}{\fill}
\begin{longtable}{ccc<{\,($\downarrow$)}ccccccccc}
    \caption{\small Detailed zero-shot comparison for Llama-3.1-8B-Instruct with Alpaca calibration. Higher is better for Accuracy (\%) (ACC.) ($\uparrow$) and lower is better for Perplexity (PPL.) ($\downarrow$).\label{tab:detail-llama318binstruct-alpaca}} \\
    \toprule
    \multirow{2}{*}{\textbf{Rank}} & \multirow{2}{*}{\textbf{Methods}} &
    \multicolumn{1}{c}{\textbf{PPL} ($\downarrow$)} &
    \multicolumn{9}{c}{\textbf{ACC} ($\uparrow$)} \\
    \cmidrule(lr){3-3} \cmidrule(lr){4-12}
    & & \multicolumn{1}{c}{} &
    \textbf{ARC-E} &
    \textbf{ARC-C} &
    \textbf{HellaSwag} &
    \textbf{PIQA} &
    \textbf{WG} &
    \textbf{MMLU} &
    \textbf{OBQA} &
    \textbf{RA} &
    \textbf{AVG} \\
    \midrule
    \endfirsthead
    \toprule
    \multirow{2}{*}{\textbf{Rank}} & \multirow{2}{*}{\textbf{Methods}} &
    \multicolumn{1}{c}{\textbf{PPL} ($\downarrow$)} &
    \multicolumn{9}{c}{\textbf{ACC} ($\uparrow$)} \\
    \cmidrule(lr){3-3} \cmidrule(lr){4-12}
    & & \multicolumn{1}{c}{} &
    \textbf{ARC-E} &
    \textbf{ARC-C} &
    \textbf{HellaSwag} &
    \textbf{PIQA} &
    \textbf{WG} &
    \textbf{MMLU} &
    \textbf{OBQA} &
    \textbf{RA} &
    \textbf{AVG} \\
    \midrule
    \endhead
    \multirow{8}{*}{\textbf{64}}
    & \texttt{Palu(SVD)} & 2260.60 & 27.53 & 25.77 & 26.50 & 52.18 & 50.04 & 24.26 & 27.80 & 20.86 & 31.87 \\
    & \texttt{MHA2MLA} & 284863.91 & 23.82 & 25.94 & 26.34 & 50.92 & 50.59 & 25.62 & 27.80 & 22.11 & 31.64 \\
    & \cellcolor{myblue!2}\textbf{\texttt{CARE-C-based-U}} & \cellcolor{myblue!2}893.84 & \cellcolor{myblue!2}31.40 & \cellcolor{myblue!2}23.98 & \cellcolor{myblue!2}26.91 & \cellcolor{myblue!2}55.66 & \cellcolor{myblue!2}50.04 & \cellcolor{myblue!2}22.91 & \cellcolor{myblue!2}27.00 & \cellcolor{myblue!2}21.15 & \cellcolor{myblue!2}32.38 \\
    & \cellcolor{myblue!6}\textbf{\texttt{CARE-C-based-E}} & \cellcolor{myblue!6}837.25 & \cellcolor{myblue!6}32.41 & \cellcolor{myblue!6}23.46 & \cellcolor{myblue!6}27.50 & \cellcolor{myblue!6}56.75 & \cellcolor{myblue!6}49.96 & \cellcolor{myblue!6}22.90 & \cellcolor{myblue!6}27.40 & \cellcolor{myblue!6}21.34 & \cellcolor{myblue!6}32.71 \\
    & \cellcolor{myblue!4}\textbf{\texttt{CARE-U}} & \cellcolor{myblue!4}983.55 & \cellcolor{myblue!4}30.51 & \cellcolor{myblue!4}23.89 & \cellcolor{myblue!4}26.82 & \cellcolor{myblue!4}54.73 & \cellcolor{myblue!4}49.01 & \cellcolor{myblue!4}23.18 & \cellcolor{myblue!4}26.00 & \cellcolor{myblue!4}20.96 & \cellcolor{myblue!4}31.89 \\
    & \cellcolor{myblue!8}\textbf{\texttt{CARE-E}} & \cellcolor{myblue!8}983.03 & \cellcolor{myblue!8}31.31 & \cellcolor{myblue!8}23.63 & \cellcolor{myblue!8}27.49 & \cellcolor{myblue!8}54.62 & \cellcolor{myblue!8}50.36 & \cellcolor{myblue!8}22.98 & \cellcolor{myblue!8}27.40 & \cellcolor{myblue!8}21.15 & \cellcolor{myblue!8}32.37 \\
    & \texttt{ASVD} & 2525.33 & 26.68 & 23.81 & 26.68 & 52.18 & 50.99 & 22.97 & 27.80 & 20.86 & 31.50 \\
    & \texttt{SVD-LLM V2} & 967.04 & 30.72 & 23.63 & 26.75 & 54.90 & 48.93 & 23.22 & 26.40 & 20.86 & 31.93 \\
\bottomrule
  \end{longtable}
\endgroup

\subsubsection{Llama-3.1-8B-Instruct-C4}
\begingroup
  \setlength{\tabcolsep}{3.2pt}
  \renewcommand{\arraystretch}{1.08}
  \scriptsize
  \setlength{\LTleft}{\fill}
  \setlength{\LTright}{\fill}
  \begin{longtable}{ccc<{\,($\downarrow$)}ccccccccc}
    \caption{\small Detailed zero-shot comparison for Llama-3.1-8B-Instruct with C4 calibration. Higher is better for Accuracy (\%) (ACC.) ($\uparrow$) and lower is better for Perplexity (PPL.) ($\downarrow$).\label{tab:detail-llama318binstruct-c4}} \\
    \toprule
    \multirow{2}{*}{\textbf{Rank}} & \multirow{2}{*}{\textbf{Methods}} &
    \multicolumn{1}{c}{\textbf{PPL} ($\downarrow$)} &
    \multicolumn{9}{c}{\textbf{ACC} ($\uparrow$)} \\
    \cmidrule(lr){3-3} \cmidrule(lr){4-12}
    & & \multicolumn{1}{c}{} &
    \textbf{ARC-E} &
    \textbf{ARC-C} &
    \textbf{HellaSwag} &
    \textbf{PIQA} &
    \textbf{WG} &
    \textbf{MMLU} &
    \textbf{OBQA} &
    \textbf{RA} &
    \textbf{AVG} \\
    \midrule
    \endfirsthead
    \toprule
    \multirow{2}{*}{\textbf{Rank}} & \multirow{2}{*}{\textbf{Methods}} &
    \multicolumn{1}{c}{\textbf{PPL} ($\downarrow$)} &
    \multicolumn{9}{c}{\textbf{ACC} ($\uparrow$)} \\
    \cmidrule(lr){3-3} \cmidrule(lr){4-12}
    & & \multicolumn{1}{c}{} &
    \textbf{ARC-E} &
    \textbf{ARC-C} &
    \textbf{HellaSwag} &
    \textbf{PIQA} &
    \textbf{WG} &
    \textbf{MMLU} &
    \textbf{OBQA} &
    \textbf{RA} &
    \textbf{AVG} \\
    \midrule
    \endhead
    \multirow{8}{*}{\textbf{64}}
    & \texttt{Palu(SVD)} & 2260.60 & 27.53 & 25.77 & 26.50 & 52.18 & 50.04 & 24.26 & 27.80 & 20.86 & 31.87 \\
    & \texttt{MHA2MLA} & 284863.91 & 23.82 & 25.94 & 26.34 & 50.92 & 50.59 & 25.62 & 27.80 & 22.11 & 31.64 \\
    & \cellcolor{myblue!2}\textbf{\texttt{CARE-C-based-U}} & \cellcolor{myblue!2}688.95 & \cellcolor{myblue!2}28.32 & \cellcolor{myblue!2}24.74 & \cellcolor{myblue!2}27.28 & \cellcolor{myblue!2}55.93 & \cellcolor{myblue!2}51.85 & \cellcolor{myblue!2}22.95 & \cellcolor{myblue!2}26.40 & \cellcolor{myblue!2}21.05 & \cellcolor{myblue!2}32.31 \\
    & \cellcolor{myblue!6}\textbf{\texttt{CARE-C-based-E}} & \cellcolor{myblue!6}582.93 & \cellcolor{myblue!6}29.55 & \cellcolor{myblue!6}24.15 & \cellcolor{myblue!6}28.15 & \cellcolor{myblue!6}56.75 & \cellcolor{myblue!6}52.25 & \cellcolor{myblue!6}22.95 & \cellcolor{myblue!6}26.20 & \cellcolor{myblue!6}20.96 & \cellcolor{myblue!6}32.62 \\
    & \cellcolor{myblue!4}\textbf{\texttt{CARE-U}} & \cellcolor{myblue!4}786.17 & \cellcolor{myblue!4}27.61 & \cellcolor{myblue!4}24.40 & \cellcolor{myblue!4}26.98 & \cellcolor{myblue!4}54.68 & \cellcolor{myblue!4}49.88 & \cellcolor{myblue!4}22.95 & \cellcolor{myblue!4}26.40 & \cellcolor{myblue!4}20.19 & \cellcolor{myblue!4}31.64 \\
    & \cellcolor{myblue!8}\textbf{\texttt{CARE-E}} & \cellcolor{myblue!8}\bfseries 676.43 & \cellcolor{myblue!8}\bfseries 28.24 & \cellcolor{myblue!8}\bfseries 24.74 & \cellcolor{myblue!8}\bfseries 27.97 & \cellcolor{myblue!8}\bfseries 55.39 & \cellcolor{myblue!8}\bfseries 50.12 & \cellcolor{myblue!8}\bfseries 22.95 & \cellcolor{myblue!8}\bfseries 26.60 & \cellcolor{myblue!8}\bfseries 20.86 & \cellcolor{myblue!8}\bfseries 32.11 \\
    & \texttt{ASVD} & 2845.27 & 25.97 & 23.98 & 27.01 & 51.90 & 48.86 & 22.98 & 29.20 & 22.30 & 31.52 \\
    & \texttt{SVD-LLM V2} & 769.05 & 27.74 & 24.57 & 27.10 & 54.62 & 49.96 & 22.95 & 26.00 & 20.38 & 31.66 \\
\bottomrule
  \end{longtable}
\endgroup

\subsubsection{Llama-3.1-8B-Instruct-PTB}
\begingroup
  \setlength{\tabcolsep}{3.2pt}
  \renewcommand{\arraystretch}{1.08}
  \scriptsize
  \setlength{\LTleft}{\fill}
  \setlength{\LTright}{\fill}
  \begin{longtable}{ccc<{\,($\downarrow$)}ccccccccc}
    \caption{\small Detailed zero-shot comparison for Llama-3.1-8B-Instruct with PTB calibration. Higher is better for Accuracy (\%) (ACC.) ($\uparrow$) and lower is better for Perplexity (PPL.) ($\downarrow$).\label{tab:detail-llama318binstruct-ptb}} \\
    \toprule
    \multirow{2}{*}{\textbf{Rank}} & \multirow{2}{*}{\textbf{Methods}} &
    \multicolumn{1}{c}{\textbf{PPL} ($\downarrow$)} &
    \multicolumn{9}{c}{\textbf{ACC} ($\uparrow$)} \\
    \cmidrule(lr){3-3} \cmidrule(lr){4-12}
    & & \multicolumn{1}{c}{} &
    \textbf{ARC-E} &
    \textbf{ARC-C} &
    \textbf{HellaSwag} &
    \textbf{PIQA} &
    \textbf{WG} &
    \textbf{MMLU} &
    \textbf{OBQA} &
    \textbf{RA} &
    \textbf{AVG} \\
    \midrule
    \endfirsthead
    \toprule
    \multirow{2}{*}{\textbf{Rank}} & \multirow{2}{*}{\textbf{Methods}} &
    \multicolumn{1}{c}{\textbf{PPL} ($\downarrow$)} &
    \multicolumn{9}{c}{\textbf{ACC} ($\uparrow$)} \\
    \cmidrule(lr){3-3} \cmidrule(lr){4-12}
    & & \multicolumn{1}{c}{} &
    \textbf{ARC-E} &
    \textbf{ARC-C} &
    \textbf{HellaSwag} &
    \textbf{PIQA} &
    \textbf{WG} &
    \textbf{MMLU} &
    \textbf{OBQA} &
    \textbf{RA} &
    \textbf{AVG} \\
    \midrule
    \endhead
    \multirow{8}{*}{\textbf{64}}
    & \texttt{Palu(SVD)} & 2260.60 & 27.53 & 25.77 & 26.50 & 52.18 & 50.04 & 24.26 & 27.80 & 20.86 & 31.87 \\
    & \texttt{MHA2MLA} & 284863.91 & 23.82 & 25.94 & 26.34 & 50.92 & 50.59 & 25.62 & 27.80 & 22.11 & 31.64 \\
    & \cellcolor{myblue!2}\textbf{\texttt{CARE-C-based-U}} & \cellcolor{myblue!2}756.56 & \cellcolor{myblue!2}26.73 & \cellcolor{myblue!2}25.68 & \cellcolor{myblue!2}26.28 & \cellcolor{myblue!2}51.58 & \cellcolor{myblue!2}50.20 & \cellcolor{myblue!2}22.95 & \cellcolor{myblue!2}25.80 & \cellcolor{myblue!2}21.34 & \cellcolor{myblue!2}31.32 \\
    & \cellcolor{myblue!6}\textbf{\texttt{CARE-C-based-E}} & \cellcolor{myblue!6}612.72 & \cellcolor{myblue!6}27.57 & \cellcolor{myblue!6}25.43 & \cellcolor{myblue!6}26.51 & \cellcolor{myblue!6}51.74 & \cellcolor{myblue!6}50.75 & \cellcolor{myblue!6}22.95 & \cellcolor{myblue!6}26.00 & \cellcolor{myblue!6}21.05 & \cellcolor{myblue!6}31.50 \\
    & \cellcolor{myblue!4}\textbf{\texttt{CARE-U}} & \cellcolor{myblue!4}815.80 & \cellcolor{myblue!4}26.26 & \cellcolor{myblue!4}25.00 & \cellcolor{myblue!4}26.50 & \cellcolor{myblue!4}51.36 & \cellcolor{myblue!4}49.25 & \cellcolor{myblue!4}22.95 & \cellcolor{myblue!4}27.00 & \cellcolor{myblue!4}20.67 & \cellcolor{myblue!4}31.12 \\
    & \cellcolor{myblue!8}\textbf{\texttt{CARE-E}} & \cellcolor{myblue!8}\bfseries 689.55 & \cellcolor{myblue!8}\bfseries 27.15 & \cellcolor{myblue!8}\bfseries 25.60 & \cellcolor{myblue!8}\bfseries 26.51 & \cellcolor{myblue!8}\bfseries 51.25 & \cellcolor{myblue!8}\bfseries 49.49 & \cellcolor{myblue!8}\bfseries 22.94 & \cellcolor{myblue!8}\bfseries 27.20 & \cellcolor{myblue!8}\bfseries 20.38 & \cellcolor{myblue!8}\bfseries 31.31 \\
    & \texttt{ASVD} & 2718.72 & 26.39 & 25.09 & 26.91 & 51.69 & 49.01 & 23.02 & 29.00 & 21.72 & 31.60 \\
    & \texttt{SVD-LLM V2} & 803.38 & 26.39 & 25.43 & 26.41 & 51.41 & 49.41 & 22.95 & 27.20 & 20.67 & 31.23 \\
\bottomrule
  \end{longtable}
\endgroup

\subsubsection{Llama-3.1-8B-Instruct-WikiText2}
\begingroup
  \setlength{\tabcolsep}{3.2pt}
  \renewcommand{\arraystretch}{1.08}
  \scriptsize
  \setlength{\LTleft}{\fill}
  \setlength{\LTright}{\fill}
  \begin{longtable}{ccc<{\,($\downarrow$)}ccccccccc}
    \caption{\small Detailed zero-shot comparison for Llama-3.1-8B-Instruct with WikiText2 calibration. Higher is better for Accuracy (\%) (ACC.) ($\uparrow$) and lower is better for Perplexity (PPL.) ($\downarrow$).\label{tab:detail-llama318binstruct-wikitext2}} \\
    \toprule
    \multirow{2}{*}{\textbf{Rank}} & \multirow{2}{*}{\textbf{Methods}} &
    \multicolumn{1}{c}{\textbf{PPL} ($\downarrow$)} &
    \multicolumn{9}{c}{\textbf{ACC} ($\uparrow$)} \\
    \cmidrule(lr){3-3} \cmidrule(lr){4-12}
    & & \multicolumn{1}{c}{} &
    \textbf{ARC-E} &
    \textbf{ARC-C} &
    \textbf{HellaSwag} &
    \textbf{PIQA} &
    \textbf{WG} &
    \textbf{MMLU} &
    \textbf{OBQA} &
    \textbf{RA} &
    \textbf{AVG} \\
    \midrule
    \endfirsthead
    \toprule
    \multirow{2}{*}{\textbf{Rank}} & \multirow{2}{*}{\textbf{Methods}} &
    \multicolumn{1}{c}{\textbf{PPL} ($\downarrow$)} &
    \multicolumn{9}{c}{\textbf{ACC} ($\uparrow$)} \\
    \cmidrule(lr){3-3} \cmidrule(lr){4-12}
    & & \multicolumn{1}{c}{} &
    \textbf{ARC-E} &
    \textbf{ARC-C} &
    \textbf{HellaSwag} &
    \textbf{PIQA} &
    \textbf{WG} &
    \textbf{MMLU} &
    \textbf{OBQA} &
    \textbf{RA} &
    \textbf{AVG} \\
    \midrule
    \endhead
    \multirow{8}{*}{\textbf{64}}
    & \texttt{Palu(SVD)} & 2260.60 & 27.53 & 25.77 & 26.50 & 52.18 & 50.04 & 24.26 & 27.80 & 20.86 & 31.87 \\
    & \texttt{MHA2MLA} & 284863.91 & 23.82 & 25.94 & 26.34 & 50.92 & 50.59 & 25.62 & 27.80 & 22.11 & 31.64 \\
    & \cellcolor{myblue!2}\textbf{\texttt{CARE-C-based-U}} & \cellcolor{myblue!2}189.39 & \cellcolor{myblue!2}28.11 & \cellcolor{myblue!2}25.09 & \cellcolor{myblue!2}26.82 & \cellcolor{myblue!2}51.47 & \cellcolor{myblue!2}50.12 & \cellcolor{myblue!2}22.95 & \cellcolor{myblue!2}27.80 & \cellcolor{myblue!2}20.57 & \cellcolor{myblue!2}31.62 \\
    & \cellcolor{myblue!6}\textbf{\texttt{CARE-C-based-E}} & \cellcolor{myblue!6}138.00 & \cellcolor{myblue!6}28.91 & \cellcolor{myblue!6}25.43 & \cellcolor{myblue!6}27.16 & \cellcolor{myblue!6}52.39 & \cellcolor{myblue!6}50.36 & \cellcolor{myblue!6}22.95 & \cellcolor{myblue!6}27.60 & \cellcolor{myblue!6}20.38 & \cellcolor{myblue!6}31.90 \\
    & \cellcolor{myblue!4}\textbf{\texttt{CARE-U}} & \cellcolor{myblue!4}237.45 & \cellcolor{myblue!4}27.31 & \cellcolor{myblue!4}25.34 & \cellcolor{myblue!4}26.46 & \cellcolor{myblue!4}51.63 & \cellcolor{myblue!4}48.22 & \cellcolor{myblue!4}22.94 & \cellcolor{myblue!4}27.60 & \cellcolor{myblue!4}20.77 & \cellcolor{myblue!4}31.28 \\
    & \cellcolor{myblue!8}\textbf{\texttt{CARE-E}} & \cellcolor{myblue!8}\bfseries 184.54 & \cellcolor{myblue!8}\bfseries 27.86 & \cellcolor{myblue!8}\bfseries 24.91 & \cellcolor{myblue!8}\bfseries 26.83 & \cellcolor{myblue!8}\bfseries 51.96 & \cellcolor{myblue!8}\bfseries 50.28 & \cellcolor{myblue!8}\bfseries 22.95 & \cellcolor{myblue!8}\bfseries 26.80 & \cellcolor{myblue!8}\bfseries 20.57 & \cellcolor{myblue!8}\bfseries 31.52 \\
    & \texttt{ASVD} & 2900.25 & 26.09 & 24.06 & 26.77 & 51.25 & 49.25 & 22.96 & 29.00 & 21.91 & 31.41 \\
    & \texttt{SVD-LLM V2} & 229.74 & 27.15 & 25.26 & 26.50 & 51.69 & 48.62 & 22.94 & 27.60 & 20.86 & 31.33 \\
\bottomrule
  \end{longtable}
\endgroup

\subsubsection{Qwen2.5-1.5B-Instruct-Alpaca}
\begingroup
  \setlength{\tabcolsep}{3.2pt}
  \renewcommand{\arraystretch}{1.08}
  \scriptsize
  \setlength{\LTleft}{\fill}
  \setlength{\LTright}{\fill}
  \begin{longtable}{ccc<{\,($\downarrow$)}ccccccccc}
    \caption{\small Detailed zero-shot comparison for Qwen2.5-1.5B-Instruct with Alpaca calibration. Higher is better for Accuracy (\%) (ACC.) ($\uparrow$) and lower is better for Perplexity (PPL.) ($\downarrow$).\label{tab:detail-qwen2515binstruct-alpaca}} \\
    \toprule
    \multirow{2}{*}{\textbf{Rank}} & \multirow{2}{*}{\textbf{Methods}} &
    \multicolumn{1}{c}{\textbf{PPL} ($\downarrow$)} &
    \multicolumn{9}{c}{\textbf{ACC} ($\uparrow$)} \\
    \cmidrule(lr){3-3} \cmidrule(lr){4-12}
    & & \multicolumn{1}{c}{} &
    \textbf{ARC-E} &
    \textbf{ARC-C} &
    \textbf{HellaSwag} &
    \textbf{PIQA} &
    \textbf{WG} &
    \textbf{MMLU} &
    \textbf{OBQA} &
    \textbf{RA} &
    \textbf{AVG} \\
    \midrule
    \endfirsthead
    \toprule
    \multirow{2}{*}{\textbf{Rank}} & \multirow{2}{*}{\textbf{Methods}} &
    \multicolumn{1}{c}{\textbf{PPL} ($\downarrow$)} &
    \multicolumn{9}{c}{\textbf{ACC} ($\uparrow$)} \\
    \cmidrule(lr){3-3} \cmidrule(lr){4-12}
    & & \multicolumn{1}{c}{} &
    \textbf{ARC-E} &
    \textbf{ARC-C} &
    \textbf{HellaSwag} &
    \textbf{PIQA} &
    \textbf{WG} &
    \textbf{MMLU} &
    \textbf{OBQA} &
    \textbf{RA} &
    \textbf{AVG} \\
    \midrule
    \endhead
    \multirow{8}{*}{\textbf{64}}
    & \texttt{Palu(SVD)} & 114579.38 & 25.17 & 25.26 & 25.55 & 50.27 & 50.20 & 24.06 & 27.40 & 20.67 & 31.07 \\
    & \texttt{MHA2MLA} & 62077.36 & 26.56 & 25.34 & 26.57 & 51.63 & 50.20 & 24.60 & 29.00 & 22.68 & 32.07 \\
    & \cellcolor{myblue!2}\textbf{\texttt{CARE-C-based-U}} & \cellcolor{myblue!2}909.76 & \cellcolor{myblue!2}29.00 & \cellcolor{myblue!2}22.35 & \cellcolor{myblue!2}26.14 & \cellcolor{myblue!2}52.50 & \cellcolor{myblue!2}48.15 & \cellcolor{myblue!2}23.11 & \cellcolor{myblue!2}26.60 & \cellcolor{myblue!2}22.30 & \cellcolor{myblue!2}31.27 \\
    & \cellcolor{myblue!6}\textbf{\texttt{CARE-C-based-E}} & \cellcolor{myblue!6}730.60 & \cellcolor{myblue!6}29.00 & \cellcolor{myblue!6}22.87 & \cellcolor{myblue!6}26.83 & \cellcolor{myblue!6}54.19 & \cellcolor{myblue!6}51.30 & \cellcolor{myblue!6}22.85 & \cellcolor{myblue!6}27.60 & \cellcolor{myblue!6}24.11 & \cellcolor{myblue!6}32.34 \\
    & \cellcolor{myblue!4}\textbf{\texttt{CARE-U}} & \cellcolor{myblue!4}1098.80 & \cellcolor{myblue!4}26.77 & \cellcolor{myblue!4}23.38 & \cellcolor{myblue!4}26.95 & \cellcolor{myblue!4}52.29 & \cellcolor{myblue!4}51.07 & \cellcolor{myblue!4}22.95 & \cellcolor{myblue!4}25.20 & \cellcolor{myblue!4}23.16 & \cellcolor{myblue!4}31.47 \\
    & \cellcolor{myblue!8}\textbf{\texttt{CARE-E}} & \cellcolor{myblue!8}\bfseries 1022.88 & \cellcolor{myblue!8}\bfseries 29.00 & \cellcolor{myblue!8}\bfseries 23.29 & \cellcolor{myblue!8}\bfseries 26.70 & \cellcolor{myblue!8}\bfseries 54.73 & \cellcolor{myblue!8}\bfseries 51.14 & \cellcolor{myblue!8}\bfseries 23.00 & \cellcolor{myblue!8}\bfseries 26.80 & \cellcolor{myblue!8}\bfseries 22.11 & \cellcolor{myblue!8}\bfseries 32.10 \\
    & \texttt{ASVD} & 7737.23 & 26.81 & 25.77 & 26.61 & 51.09 & 49.88 & 25.46 & 27.60 & 22.11 & 31.92 \\
    & \texttt{SVD-LLM V2} & 1109.62 & 28.37 & 23.63 & 26.91 & 53.48 & 49.80 & 22.95 & 23.80 & 23.92 & 31.61 \\
\bottomrule
  \end{longtable}
\endgroup

\subsubsection{Qwen2.5-1.5B-Instruct-C4}
\begingroup
  \setlength{\tabcolsep}{3.2pt}
  \renewcommand{\arraystretch}{1.08}
  \scriptsize
  \setlength{\LTleft}{\fill}
  \setlength{\LTright}{\fill}
  \begin{longtable}{ccc<{\,($\downarrow$)}ccccccccc}
    \caption{\small Detailed zero-shot comparison for Qwen2.5-1.5B-Instruct with C4 calibration. Higher is better for Accuracy (\%) (ACC.) ($\uparrow$) and lower is better for Perplexity (PPL.) ($\downarrow$).\label{tab:detail-qwen2515binstruct-c4}} \\
    \toprule
    \multirow{2}{*}{\textbf{Rank}} & \multirow{2}{*}{\textbf{Methods}} &
    \multicolumn{1}{c}{\textbf{PPL} ($\downarrow$)} &
    \multicolumn{9}{c}{\textbf{ACC} ($\uparrow$)} \\
    \cmidrule(lr){3-3} \cmidrule(lr){4-12}
    & & \multicolumn{1}{c}{} &
    \textbf{ARC-E} &
    \textbf{ARC-C} &
    \textbf{HellaSwag} &
    \textbf{PIQA} &
    \textbf{WG} &
    \textbf{MMLU} &
    \textbf{OBQA} &
    \textbf{RA} &
    \textbf{AVG} \\
    \midrule
    \endfirsthead
    \toprule
    \multirow{2}{*}{\textbf{Rank}} & \multirow{2}{*}{\textbf{Methods}} &
    \multicolumn{1}{c}{\textbf{PPL} ($\downarrow$)} &
    \multicolumn{9}{c}{\textbf{ACC} ($\uparrow$)} \\
    \cmidrule(lr){3-3} \cmidrule(lr){4-12}
    & & \multicolumn{1}{c}{} &
    \textbf{ARC-E} &
    \textbf{ARC-C} &
    \textbf{HellaSwag} &
    \textbf{PIQA} &
    \textbf{WG} &
    \textbf{MMLU} &
    \textbf{OBQA} &
    \textbf{RA} &
    \textbf{AVG} \\
    \midrule
    \endhead
    \multirow{8}{*}{\textbf{64}}
    & \texttt{Palu(SVD)} & 114579.38 & 25.17 & 25.26 & 25.55 & 50.27 & 50.20 & 24.06 & 27.40 & 20.67 & 31.07 \\
    & \texttt{MHA2MLA} & 62077.36 & 26.56 & 25.34 & 26.57 & 51.63 & 50.20 & 24.60 & 29.00 & 22.68 & 32.07 \\
    & \cellcolor{myblue!2}\textbf{\texttt{CARE-C-based-U}} & \cellcolor{myblue!2}679.30 & \cellcolor{myblue!2}27.74 & \cellcolor{myblue!2}23.12 & \cellcolor{myblue!2}26.69 & \cellcolor{myblue!2}52.77 & \cellcolor{myblue!2}51.22 & \cellcolor{myblue!2}22.95 & \cellcolor{myblue!2}26.80 & \cellcolor{myblue!2}23.16 & \cellcolor{myblue!2}31.81 \\
    & \cellcolor{myblue!6}\textbf{\texttt{CARE-C-based-E}} & \cellcolor{myblue!6}365.98 & \cellcolor{myblue!6}28.54 & \cellcolor{myblue!6}22.70 & \cellcolor{myblue!6}27.42 & \cellcolor{myblue!6}55.01 & \cellcolor{myblue!6}52.09 & \cellcolor{myblue!6}22.95 & \cellcolor{myblue!6}26.40 & \cellcolor{myblue!6}23.64 & \cellcolor{myblue!6}32.34 \\
    & \cellcolor{myblue!4}\textbf{\texttt{CARE-U}} & \cellcolor{myblue!4}914.36 & \cellcolor{myblue!4}26.43 & \cellcolor{myblue!4}24.57 & \cellcolor{myblue!4}26.17 & \cellcolor{myblue!4}51.90 & \cellcolor{myblue!4}49.41 & \cellcolor{myblue!4}22.95 & \cellcolor{myblue!4}25.80 & \cellcolor{myblue!4}23.83 & \cellcolor{myblue!4}31.38 \\
    & \cellcolor{myblue!8}\textbf{\texttt{CARE-E}} & \cellcolor{myblue!8}\bfseries 664.13 & \cellcolor{myblue!8}\bfseries 27.15 & \cellcolor{myblue!8}\bfseries 23.12 & \cellcolor{myblue!8}\bfseries 27.44 & \cellcolor{myblue!8}\bfseries 53.43 & \cellcolor{myblue!8}\bfseries 51.38 & \cellcolor{myblue!8}\bfseries 22.94 & \cellcolor{myblue!8}\bfseries 27.00 & \cellcolor{myblue!8}\bfseries 23.64 & \cellcolor{myblue!8}\bfseries 32.01 \\
    & \texttt{ASVD} & 7074.97 & 26.85 & 25.43 & 27.14 & 50.98 & 50.91 & 25.89 & 30.00 & 22.49 & 32.46 \\
    & \texttt{SVD-LLM V2} & 826.83 & 26.22 & 25.26 & 26.41 & 51.90 & 49.64 & 22.95 & 25.40 & 24.11 & 31.49 \\
\bottomrule
  \end{longtable}
\endgroup

\subsubsection{Qwen2.5-1.5B-Instruct-PTB}
\begingroup
  \setlength{\tabcolsep}{3.2pt}
  \renewcommand{\arraystretch}{1.08}
  \scriptsize
  \setlength{\LTleft}{\fill}
  \setlength{\LTright}{\fill}
  \begin{longtable}{ccc<{\,($\downarrow$)}ccccccccc}
    \caption{\small Detailed zero-shot comparison for Qwen2.5-1.5B-Instruct with PTB calibration. Higher is better for Accuracy (\%) (ACC.) ($\uparrow$) and lower is better for Perplexity (PPL.) ($\downarrow$).\label{tab:detail-qwen2515binstruct-ptb}} \\
    \toprule
    \multirow{2}{*}{\textbf{Rank}} & \multirow{2}{*}{\textbf{Methods}} &
    \multicolumn{1}{c}{\textbf{PPL} ($\downarrow$)} &
    \multicolumn{9}{c}{\textbf{ACC} ($\uparrow$)} \\
    \cmidrule(lr){3-3} \cmidrule(lr){4-12}
    & & \multicolumn{1}{c}{} &
    \textbf{ARC-E} &
    \textbf{ARC-C} &
    \textbf{HellaSwag} &
    \textbf{PIQA} &
    \textbf{WG} &
    \textbf{MMLU} &
    \textbf{OBQA} &
    \textbf{RA} &
    \textbf{AVG} \\
    \midrule
    \endfirsthead
    \toprule
    \multirow{2}{*}{\textbf{Rank}} & \multirow{2}{*}{\textbf{Methods}} &
    \multicolumn{1}{c}{\textbf{PPL} ($\downarrow$)} &
    \multicolumn{9}{c}{\textbf{ACC} ($\uparrow$)} \\
    \cmidrule(lr){3-3} \cmidrule(lr){4-12}
    & & \multicolumn{1}{c}{} &
    \textbf{ARC-E} &
    \textbf{ARC-C} &
    \textbf{HellaSwag} &
    \textbf{PIQA} &
    \textbf{WG} &
    \textbf{MMLU} &
    \textbf{OBQA} &
    \textbf{RA} &
    \textbf{AVG} \\
    \midrule
    \endhead
    \multirow{8}{*}{\textbf{64}}
    & \texttt{Palu(SVD)} & 114579.38 & 25.17 & 25.26 & 25.55 & 50.27 & 50.20 & 24.06 & 27.40 & 20.67 & 31.07 \\
    & \texttt{MHA2MLA} & 62077.36 & 26.56 & 25.34 & 26.57 & 51.63 & 50.20 & 24.60 & 29.00 & 22.68 & 32.07 \\
    & \cellcolor{myblue!2}\textbf{\texttt{CARE-C-based-U}} & \cellcolor{myblue!2}1608.42 & \cellcolor{myblue!2}27.48 & \cellcolor{myblue!2}23.46 & \cellcolor{myblue!2}26.79 & \cellcolor{myblue!2}51.09 & \cellcolor{myblue!2}50.59 & \cellcolor{myblue!2}23.01 & \cellcolor{myblue!2}26.80 & \cellcolor{myblue!2}21.63 & \cellcolor{myblue!2}31.36 \\
    & \cellcolor{myblue!6}\textbf{\texttt{CARE-C-based-E}} & \cellcolor{myblue!6}701.84 & \cellcolor{myblue!6}27.95 & \cellcolor{myblue!6}23.55 & \cellcolor{myblue!6}27.12 & \cellcolor{myblue!6}51.52 & \cellcolor{myblue!6}50.04 & \cellcolor{myblue!6}22.95 & \cellcolor{myblue!6}24.80 & \cellcolor{myblue!6}23.73 & \cellcolor{myblue!6}31.46 \\
    & \cellcolor{myblue!4}\textbf{\texttt{CARE-U}} & \cellcolor{myblue!4}2131.72 & \cellcolor{myblue!4}27.53 & \cellcolor{myblue!4}23.38 & \cellcolor{myblue!4}27.00 & \cellcolor{myblue!4}51.74 & \cellcolor{myblue!4}49.01 & \cellcolor{myblue!4}23.14 & \cellcolor{myblue!4}27.20 & \cellcolor{myblue!4}23.06 & \cellcolor{myblue!4}31.51 \\
    & \cellcolor{myblue!8}\textbf{\texttt{CARE-E}} & \cellcolor{myblue!8}\bfseries 362.35 & \cellcolor{myblue!8}\bfseries 29.59 & \cellcolor{myblue!8}\bfseries 23.55 & \cellcolor{myblue!8}\bfseries 29.43 & \cellcolor{myblue!8}\bfseries 52.45 & \cellcolor{myblue!8}\bfseries 50.83 & \cellcolor{myblue!8}\bfseries 23.10 & \cellcolor{myblue!8}\bfseries 25.00 & \cellcolor{myblue!8}\bfseries 22.39 & \cellcolor{myblue!8}\bfseries 32.04 \\
    & \texttt{ASVD} & 8309.98 & 26.98 & 24.83 & 26.90 & 50.82 & 50.67 & 25.40 & 28.40 & 21.91 & 31.99 \\
    & \texttt{SVD-LLM V2} & 2075.31 & 27.36 & 22.61 & 27.18 & 50.82 & 49.01 & 23.13 & 27.60 & 22.68 & 31.30 \\
\bottomrule
  \end{longtable}
\endgroup

\subsubsection{Qwen2.5-1.5B-Instruct-WikiText2}
\begingroup
  \setlength{\tabcolsep}{3.2pt}
  \renewcommand{\arraystretch}{1.08}
  \scriptsize
  \setlength{\LTleft}{\fill}
  \setlength{\LTright}{\fill}
  \begin{longtable}{ccc<{\,($\downarrow$)}ccccccccc}
    \caption{\small Detailed zero-shot comparison for Qwen2.5-1.5B-Instruct with WikiText2 calibration. Higher is better for Accuracy (\%) (ACC.) ($\uparrow$) and lower is better for Perplexity (PPL.) ($\downarrow$).\label{tab:detail-qwen2515binstruct-wikitext2}} \\
    \toprule
    \multirow{2}{*}{\textbf{Rank}} & \multirow{2}{*}{\textbf{Methods}} &
    \multicolumn{1}{c}{\textbf{PPL} ($\downarrow$)} &
    \multicolumn{9}{c}{\textbf{ACC} ($\uparrow$)} \\
    \cmidrule(lr){3-3} \cmidrule(lr){4-12}
    & & \multicolumn{1}{c}{} &
    \textbf{ARC-E} &
    \textbf{ARC-C} &
    \textbf{HellaSwag} &
    \textbf{PIQA} &
    \textbf{WG} &
    \textbf{MMLU} &
    \textbf{OBQA} &
    \textbf{RA} &
    \textbf{AVG} \\
    \midrule
    \endfirsthead
    \toprule
    \multirow{2}{*}{\textbf{Rank}} & \multirow{2}{*}{\textbf{Methods}} &
    \multicolumn{1}{c}{\textbf{PPL} ($\downarrow$)} &
    \multicolumn{9}{c}{\textbf{ACC} ($\uparrow$)} \\
    \cmidrule(lr){3-3} \cmidrule(lr){4-12}
    & & \multicolumn{1}{c}{} &
    \textbf{ARC-E} &
    \textbf{ARC-C} &
    \textbf{HellaSwag} &
    \textbf{PIQA} &
    \textbf{WG} &
    \textbf{MMLU} &
    \textbf{OBQA} &
    \textbf{RA} &
    \textbf{AVG} \\
    \midrule
    \endhead
    \multirow{8}{*}{\textbf{64}}
    & \texttt{Palu(SVD)} & 114579.38 & 25.17 & 25.26 & 25.55 & 50.27 & 50.20 & 24.06 & 27.40 & 20.67 & 31.07 \\
    & \texttt{MHA2MLA} & 62077.36 & 26.56 & 25.34 & 26.57 & 51.63 & 50.20 & 24.60 & 29.00 & 22.68 & 32.07 \\
    & \cellcolor{myblue!2}\textbf{\texttt{CARE-C-based-U}} & \cellcolor{myblue!2}209.05 & \cellcolor{myblue!2}27.86 & \cellcolor{myblue!2}22.18 & \cellcolor{myblue!2}26.34 & \cellcolor{myblue!2}52.56 & \cellcolor{myblue!2}50.67 & \cellcolor{myblue!2}22.98 & \cellcolor{myblue!2}28.20 & \cellcolor{myblue!2}23.64 & \cellcolor{myblue!2}31.80 \\
    & \cellcolor{myblue!6}\textbf{\texttt{CARE-C-based-E}} & \cellcolor{myblue!6}122.65 & \cellcolor{myblue!6}27.86 & \cellcolor{myblue!6}23.29 & \cellcolor{myblue!6}27.12 & \cellcolor{myblue!6}53.48 & \cellcolor{myblue!6}50.75 & \cellcolor{myblue!6}22.94 & \cellcolor{myblue!6}26.20 & \cellcolor{myblue!6}22.68 & \cellcolor{myblue!6}31.79 \\
    & \cellcolor{myblue!4}\textbf{\texttt{CARE-U}} & \cellcolor{myblue!4}300.99 & \cellcolor{myblue!4}26.77 & \cellcolor{myblue!4}23.12 & \cellcolor{myblue!4}26.74 & \cellcolor{myblue!4}52.18 & \cellcolor{myblue!4}50.20 & \cellcolor{myblue!4}22.95 & \cellcolor{myblue!4}26.60 & \cellcolor{myblue!4}24.50 & \cellcolor{myblue!4}31.63 \\
    & \cellcolor{myblue!8}\textbf{\texttt{CARE-E}} & \cellcolor{myblue!8}\bfseries 125.54 & \cellcolor{myblue!8}\bfseries 27.31 & \cellcolor{myblue!8}\bfseries 23.12 & \cellcolor{myblue!8}\bfseries 27.86 & \cellcolor{myblue!8}\bfseries 52.45 & \cellcolor{myblue!8}\bfseries 50.99 & \cellcolor{myblue!8}\bfseries 22.96 & \cellcolor{myblue!8}\bfseries 25.40 & \cellcolor{myblue!8}\bfseries 23.54 & \cellcolor{myblue!8}\bfseries 31.70 \\
    & \texttt{ASVD} & 7126.52 & 26.81 & 25.60 & 27.19 & 51.90 & 50.20 & 25.71 & 26.80 & 21.82 & 32.00 \\
    & \texttt{SVD-LLM V2} & 264.86 & 26.94 & 23.21 & 26.87 & 51.96 & 49.96 & 22.97 & 26.40 & 24.59 & 31.61 \\
\bottomrule
  \end{longtable}
\endgroup

\subsubsection{Qwen3-4B-Instruct-2507-Alpaca}
\begingroup
  \setlength{\tabcolsep}{3.2pt}
  \renewcommand{\arraystretch}{1.08}
  \scriptsize
  \setlength{\LTleft}{\fill}
  \setlength{\LTright}{\fill}
  \begin{longtable}{ccc<{\,($\downarrow$)}ccccccccc}
    \caption{\small Detailed zero-shot comparison for Qwen3-4B-Instruct-2507 with Alpaca calibration. Higher is better for Accuracy (\%) (ACC.) ($\uparrow$) and lower is better for Perplexity (PPL.) ($\downarrow$).\label{tab:detail-qwen34binstruct2507-alpaca}} \\
    \toprule
    \multirow{2}{*}{\textbf{Rank}} & \multirow{2}{*}{\textbf{Methods}} &
    \multicolumn{1}{c}{\textbf{PPL} ($\downarrow$)} &
    \multicolumn{9}{c}{\textbf{ACC} ($\uparrow$)} \\
    \cmidrule(lr){3-3} \cmidrule(lr){4-12}
    & & \multicolumn{1}{c}{} &
    \textbf{ARC-E} &
    \textbf{ARC-C} &
    \textbf{HellaSwag} &
    \textbf{PIQA} &
    \textbf{WG} &
    \textbf{MMLU} &
    \textbf{OBQA} &
    \textbf{RA} &
    \textbf{AVG} \\
    \midrule
    \endfirsthead
    \toprule
    \multirow{2}{*}{\textbf{Rank}} & \multirow{2}{*}{\textbf{Methods}} &
    \multicolumn{1}{c}{\textbf{PPL} ($\downarrow$)} &
    \multicolumn{9}{c}{\textbf{ACC} ($\uparrow$)} \\
    \cmidrule(lr){3-3} \cmidrule(lr){4-12}
    & & \multicolumn{1}{c}{} &
    \textbf{ARC-E} &
    \textbf{ARC-C} &
    \textbf{HellaSwag} &
    \textbf{PIQA} &
    \textbf{WG} &
    \textbf{MMLU} &
    \textbf{OBQA} &
    \textbf{RA} &
    \textbf{AVG} \\
    \midrule
    \endhead
    \multirow{8}{*}{\textbf{64}}
    & \texttt{Palu(SVD)} & 56922.11 & 26.77 & 25.34 & 25.73 & 50.44 & 50.91 & 24.31 & 28.40 & 22.11 & 31.75 \\
    & \texttt{MHA2MLA} & 21850.16 & 25.46 & 25.68 & 26.03 & 51.96 & 50.20 & 22.92 & 27.00 & 22.01 & 31.41 \\
    & \cellcolor{myblue!2}\textbf{\texttt{CARE-C-based-U}} & \cellcolor{myblue!2}1052.43 & \cellcolor{myblue!2}29.34 & \cellcolor{myblue!2}22.70 & \cellcolor{myblue!2}27.51 & \cellcolor{myblue!2}54.35 & \cellcolor{myblue!2}50.67 & \cellcolor{myblue!2}23.21 & \cellcolor{myblue!2}26.20 & \cellcolor{myblue!2}23.64 & \cellcolor{myblue!2}32.20 \\
    & \cellcolor{myblue!6}\textbf{\texttt{CARE-C-based-E}} & \cellcolor{myblue!6}941.16 & \cellcolor{myblue!6}28.75 & \cellcolor{myblue!6}23.12 & \cellcolor{myblue!6}27.33 & \cellcolor{myblue!6}55.33 & \cellcolor{myblue!6}50.28 & \cellcolor{myblue!6}23.08 & \cellcolor{myblue!6}25.60 & \cellcolor{myblue!6}22.30 & \cellcolor{myblue!6}31.97 \\
    & \cellcolor{myblue!4}\textbf{\texttt{CARE-U}} & \cellcolor{myblue!4}905.74 & \cellcolor{myblue!4}29.00 & \cellcolor{myblue!4}23.46 & \cellcolor{myblue!4}27.61 & \cellcolor{myblue!4}55.22 & \cellcolor{myblue!4}50.59 & \cellcolor{myblue!4}23.03 & \cellcolor{myblue!4}26.60 & \cellcolor{myblue!4}23.06 & \cellcolor{myblue!4}32.32 \\
    & \cellcolor{myblue!8}\textbf{\texttt{CARE-E}} & \cellcolor{myblue!8}\bfseries 730.93 & \cellcolor{myblue!8}\bfseries 30.77 & \cellcolor{myblue!8}\bfseries 23.29 & \cellcolor{myblue!8}\bfseries 28.33 & \cellcolor{myblue!8}\bfseries 55.22 & \cellcolor{myblue!8}\bfseries 51.54 & \cellcolor{myblue!8}\bfseries 22.95 & \cellcolor{myblue!8}\bfseries 24.80 & \cellcolor{myblue!8}\bfseries 22.78 & \cellcolor{myblue!8}\bfseries 32.46 \\
    & \texttt{ASVD} & 6683.95 & 25.00 & 27.39 & 25.71 & 50.38 & 50.43 & 24.08 & 26.60 & 22.11 & 31.46 \\
    & \texttt{SVD-LLM V2} & 894.94 & 28.58 & 23.12 & 27.66 & 55.11 & 51.54 & 23.02 & 27.20 & 23.44 & 32.46 \\
\bottomrule
  \end{longtable}
\endgroup

\subsubsection{Qwen3-4B-Instruct-2507-C4}
\begingroup
  \setlength{\tabcolsep}{3.2pt}
  \renewcommand{\arraystretch}{1.08}
  \scriptsize
  \setlength{\LTleft}{\fill}
  \setlength{\LTright}{\fill}
  \begin{longtable}{ccc<{\,($\downarrow$)}ccccccccc}
    \caption{\small Detailed zero-shot comparison for Qwen3-4B-Instruct-2507 with C4 calibration. Higher is better for Accuracy (\%) (ACC.) ($\uparrow$) and lower is better for Perplexity (PPL.) ($\downarrow$).\label{tab:detail-qwen34binstruct2507-c4}} \\
    \toprule
    \multirow{2}{*}{\textbf{Rank}} & \multirow{2}{*}{\textbf{Methods}} &
    \multicolumn{1}{c}{\textbf{PPL} ($\downarrow$)} &
    \multicolumn{9}{c}{\textbf{ACC} ($\uparrow$)} \\
    \cmidrule(lr){3-3} \cmidrule(lr){4-12}
    & & \multicolumn{1}{c}{} &
    \textbf{ARC-E} &
    \textbf{ARC-C} &
    \textbf{HellaSwag} &
    \textbf{PIQA} &
    \textbf{WG} &
    \textbf{MMLU} &
    \textbf{OBQA} &
    \textbf{RA} &
    \textbf{AVG} \\
    \midrule
    \endfirsthead
    \toprule
    \multirow{2}{*}{\textbf{Rank}} & \multirow{2}{*}{\textbf{Methods}} &
    \multicolumn{1}{c}{\textbf{PPL} ($\downarrow$)} &
    \multicolumn{9}{c}{\textbf{ACC} ($\uparrow$)} \\
    \cmidrule(lr){3-3} \cmidrule(lr){4-12}
    & & \multicolumn{1}{c}{} &
    \textbf{ARC-E} &
    \textbf{ARC-C} &
    \textbf{HellaSwag} &
    \textbf{PIQA} &
    \textbf{WG} &
    \textbf{MMLU} &
    \textbf{OBQA} &
    \textbf{RA} &
    \textbf{AVG} \\
    \midrule
    \endhead
    \multirow{8}{*}{\textbf{64}}
    & \texttt{Palu(SVD)} & 56922.11 & 26.77 & 25.34 & 25.73 & 50.44 & 50.91 & 24.31 & 28.40 & 22.11 & 31.75 \\
    & \texttt{MHA2MLA} & 21850.16 & 25.46 & 25.68 & 26.03 & 51.96 & 50.20 & 22.92 & 27.00 & 22.01 & 31.41 \\
    & \cellcolor{myblue!2}\textbf{\texttt{CARE-C-based-U}} & \cellcolor{myblue!2}494.84 & \cellcolor{myblue!2}28.91 & \cellcolor{myblue!2}23.55 & \cellcolor{myblue!2}27.41 & \cellcolor{myblue!2}55.11 & \cellcolor{myblue!2}50.91 & \cellcolor{myblue!2}22.93 & \cellcolor{myblue!2}23.60 & \cellcolor{myblue!2}24.21 & \cellcolor{myblue!2}32.08 \\
    & \cellcolor{myblue!6}\textbf{\texttt{CARE-C-based-E}} & \cellcolor{myblue!6}447.15 & \cellcolor{myblue!6}28.28 & \cellcolor{myblue!6}21.50 & \cellcolor{myblue!6}27.90 & \cellcolor{myblue!6}55.28 & \cellcolor{myblue!6}50.36 & \cellcolor{myblue!6}22.95 & \cellcolor{myblue!6}24.40 & \cellcolor{myblue!6}23.35 & \cellcolor{myblue!6}31.75 \\
    & \cellcolor{myblue!4}\textbf{\texttt{CARE-U}} & \cellcolor{myblue!4}380.04 & \cellcolor{myblue!4}29.59 & \cellcolor{myblue!4}22.53 & \cellcolor{myblue!4}28.13 & \cellcolor{myblue!4}53.21 & \cellcolor{myblue!4}51.14 & \cellcolor{myblue!4}22.92 & \cellcolor{myblue!4}26.20 & \cellcolor{myblue!4}23.64 & \cellcolor{myblue!4}32.17 \\
    & \cellcolor{myblue!8}\textbf{\texttt{CARE-E}} & \cellcolor{myblue!8}\bfseries 320.78 & \cellcolor{myblue!8}\bfseries 29.25 & \cellcolor{myblue!8}\bfseries 23.63 & \cellcolor{myblue!8}\bfseries 29.88 & \cellcolor{myblue!8}\bfseries 54.79 & \cellcolor{myblue!8}\bfseries 50.91 & \cellcolor{myblue!8}\bfseries 22.95 & \cellcolor{myblue!8}\bfseries 24.40 & \cellcolor{myblue!8}\bfseries 23.54 & \cellcolor{myblue!8}\bfseries 32.42 \\
    & \texttt{ASVD} & 5126.87 & 26.98 & 23.55 & 26.77 & 52.23 & 52.80 & 23.79 & 25.80 & 22.30 & 31.78 \\
    & \texttt{SVD-LLM V2} & 390.08 & 29.34 & 22.44 & 28.10 & 53.97 & 50.83 & 22.97 & 26.00 & 23.64 & 32.16 \\
\bottomrule
  \end{longtable}
\endgroup

\subsubsection{Qwen3-4B-Instruct-2507-PTB}
\begingroup
  \setlength{\tabcolsep}{3.2pt}
  \renewcommand{\arraystretch}{1.08}
  \scriptsize
  \setlength{\LTleft}{\fill}
  \setlength{\LTright}{\fill}
  \begin{longtable}{ccc<{\,($\downarrow$)}ccccccccc}
    \caption{\small Detailed zero-shot comparison for Qwen3-4B-Instruct-2507 with PTB calibration. Higher is better for Accuracy (\%) (ACC.) ($\uparrow$) and lower is better for Perplexity (PPL.) ($\downarrow$).\label{tab:detail-qwen34binstruct2507-ptb}} \\
    \toprule
    \multirow{2}{*}{\textbf{Rank}} & \multirow{2}{*}{\textbf{Methods}} &
    \multicolumn{1}{c}{\textbf{PPL} ($\downarrow$)} &
    \multicolumn{9}{c}{\textbf{ACC} ($\uparrow$)} \\
    \cmidrule(lr){3-3} \cmidrule(lr){4-12}
    & & \multicolumn{1}{c}{} &
    \textbf{ARC-E} &
    \textbf{ARC-C} &
    \textbf{HellaSwag} &
    \textbf{PIQA} &
    \textbf{WG} &
    \textbf{MMLU} &
    \textbf{OBQA} &
    \textbf{RA} &
    \textbf{AVG} \\
    \midrule
    \endfirsthead
    \toprule
    \multirow{2}{*}{\textbf{Rank}} & \multirow{2}{*}{\textbf{Methods}} &
    \multicolumn{1}{c}{\textbf{PPL} ($\downarrow$)} &
    \multicolumn{9}{c}{\textbf{ACC} ($\uparrow$)} \\
    \cmidrule(lr){3-3} \cmidrule(lr){4-12}
    & & \multicolumn{1}{c}{} &
    \textbf{ARC-E} &
    \textbf{ARC-C} &
    \textbf{HellaSwag} &
    \textbf{PIQA} &
    \textbf{WG} &
    \textbf{MMLU} &
    \textbf{OBQA} &
    \textbf{RA} &
    \textbf{AVG} \\
    \midrule
    \endhead
    \multirow{8}{*}{\textbf{64}}
    & \texttt{Palu(SVD)} & 56922.11 & 26.77 & 25.34 & 25.73 & 50.44 & 50.91 & 24.31 & 28.40 & 22.11 & 31.75 \\
    & \texttt{MHA2MLA} & 21850.16 & 25.46 & 25.68 & 26.03 & 51.96 & 50.20 & 22.92 & 27.00 & 22.01 & 31.41 \\
    & \cellcolor{myblue!2}\textbf{\texttt{CARE-C-based-U}} & \cellcolor{myblue!2}1234.23 & \cellcolor{myblue!2}28.49 & \cellcolor{myblue!2}22.44 & \cellcolor{myblue!2}26.91 & \cellcolor{myblue!2}51.74 & \cellcolor{myblue!2}48.46 & \cellcolor{myblue!2}22.95 & \cellcolor{myblue!2}24.20 & \cellcolor{myblue!2}22.20 & \cellcolor{myblue!2}30.92 \\
    & \cellcolor{myblue!6}\textbf{\texttt{CARE-C-based-E}} & \cellcolor{myblue!6}1036.30 & \cellcolor{myblue!6}28.32 & \cellcolor{myblue!6}22.87 & \cellcolor{myblue!6}26.93 & \cellcolor{myblue!6}52.67 & \cellcolor{myblue!6}50.83 & \cellcolor{myblue!6}22.95 & \cellcolor{myblue!6}24.40 & \cellcolor{myblue!6}21.82 & \cellcolor{myblue!6}31.35 \\
    & \cellcolor{myblue!4}\textbf{\texttt{CARE-U}} & \cellcolor{myblue!4}961.94 & \cellcolor{myblue!4}28.79 & \cellcolor{myblue!4}23.29 & \cellcolor{myblue!4}27.49 & \cellcolor{myblue!4}51.31 & \cellcolor{myblue!4}49.09 & \cellcolor{myblue!4}22.98 & \cellcolor{myblue!4}24.60 & \cellcolor{myblue!4}22.58 & \cellcolor{myblue!4}31.27 \\
    & \cellcolor{myblue!8}\textbf{\texttt{CARE-E}} & \cellcolor{myblue!8}\bfseries 671.86 & \cellcolor{myblue!8}\bfseries 29.50 & \cellcolor{myblue!8}\bfseries 24.23 & \cellcolor{myblue!8}\bfseries 27.87 & \cellcolor{myblue!8}\bfseries 52.39 & \cellcolor{myblue!8}\bfseries 49.33 & \cellcolor{myblue!8}\bfseries 22.90 & \cellcolor{myblue!8}\bfseries 24.80 & \cellcolor{myblue!8}\bfseries 23.16 & \cellcolor{myblue!8}\bfseries 31.77 \\
    & \texttt{ASVD} & 4441.27 & 26.56 & 24.91 & 25.94 & 51.90 & 51.70 & 23.69 & 27.40 & 21.91 & 31.75 \\
    & \texttt{SVD-LLM V2} & 1001.12 & 29.29 & 23.46 & 27.37 & 51.58 & 49.96 & 23.00 & 24.20 & 22.58 & 31.43 \\
\bottomrule
  \end{longtable}
\endgroup

\subsubsection{Qwen3-4B-Instruct-2507-WikiText2}
\begingroup
  \setlength{\tabcolsep}{3.2pt}
  \renewcommand{\arraystretch}{1.08}
  \scriptsize
  \setlength{\LTleft}{\fill}
  \setlength{\LTright}{\fill}
  \begin{longtable}{ccc<{\,($\downarrow$)}ccccccccc}
    \caption{\small Detailed zero-shot comparison for Qwen3-4B-Instruct-2507 with WikiText2 calibration. Higher is better for Accuracy (\%) (ACC.) ($\uparrow$) and lower is better for Perplexity (PPL.) ($\downarrow$).\label{tab:detail-qwen34binstruct2507-wikitext2}} \\
    \toprule
    \multirow{2}{*}{\textbf{Rank}} & \multirow{2}{*}{\textbf{Methods}} &
    \multicolumn{1}{c}{\textbf{PPL} ($\downarrow$)} &
    \multicolumn{9}{c}{\textbf{ACC} ($\uparrow$)} \\
    \cmidrule(lr){3-3} \cmidrule(lr){4-12}
    & & \multicolumn{1}{c}{} &
    \textbf{ARC-E} &
    \textbf{ARC-C} &
    \textbf{HellaSwag} &
    \textbf{PIQA} &
    \textbf{WG} &
    \textbf{MMLU} &
    \textbf{OBQA} &
    \textbf{RA} &
    \textbf{AVG} \\
    \midrule
    \endfirsthead
    \toprule
    \multirow{2}{*}{\textbf{Rank}} & \multirow{2}{*}{\textbf{Methods}} &
    \multicolumn{1}{c}{\textbf{PPL} ($\downarrow$)} &
    \multicolumn{9}{c}{\textbf{ACC} ($\uparrow$)} \\
    \cmidrule(lr){3-3} \cmidrule(lr){4-12}
    & & \multicolumn{1}{c}{} &
    \textbf{ARC-E} &
    \textbf{ARC-C} &
    \textbf{HellaSwag} &
    \textbf{PIQA} &
    \textbf{WG} &
    \textbf{MMLU} &
    \textbf{OBQA} &
    \textbf{RA} &
    \textbf{AVG} \\
    \midrule
    \endhead
    \multirow{8}{*}{\textbf{64}}
    & \texttt{Palu(SVD)} & 56922.11 & 26.77 & 25.34 & 25.73 & 50.44 & 50.91 & 24.31 & 28.40 & 22.11 & 31.75 \\
    & \texttt{MHA2MLA} & 21850.16 & 25.46 & 25.68 & 26.03 & 51.96 & 50.20 & 22.92 & 27.00 & 22.01 & 31.41 \\
    & \cellcolor{myblue!2}\textbf{\texttt{CARE-C-based-U}} & \cellcolor{myblue!2}169.94 & \cellcolor{myblue!2}27.99 & \cellcolor{myblue!2}22.35 & \cellcolor{myblue!2}27.16 & \cellcolor{myblue!2}52.18 & \cellcolor{myblue!2}49.25 & \cellcolor{myblue!2}22.95 & \cellcolor{myblue!2}25.60 & \cellcolor{myblue!2}23.35 & \cellcolor{myblue!2}31.35 \\
    & \cellcolor{myblue!6}\textbf{\texttt{CARE-C-based-E}} & \cellcolor{myblue!6}142.06 & \cellcolor{myblue!6}28.37 & \cellcolor{myblue!6}23.81 & \cellcolor{myblue!6}27.50 & \cellcolor{myblue!6}52.72 & \cellcolor{myblue!6}49.64 & \cellcolor{myblue!6}22.95 & \cellcolor{myblue!6}27.20 & \cellcolor{myblue!6}22.20 & \cellcolor{myblue!6}31.80 \\
    & \cellcolor{myblue!4}\textbf{\texttt{CARE-U}} & \cellcolor{myblue!4}156.42 & \cellcolor{myblue!4}28.66 & \cellcolor{myblue!4}23.29 & \cellcolor{myblue!4}28.50 & \cellcolor{myblue!4}52.45 & \cellcolor{myblue!4}49.96 & \cellcolor{myblue!4}22.95 & \cellcolor{myblue!4}26.00 & \cellcolor{myblue!4}22.11 & \cellcolor{myblue!4}31.74 \\
    & \cellcolor{myblue!8}\textbf{\texttt{CARE-E}} & \cellcolor{myblue!8}\bfseries 112.21 & \cellcolor{myblue!8}\bfseries 30.05 & \cellcolor{myblue!8}\bfseries 23.29 & \cellcolor{myblue!8}\bfseries 29.12 & \cellcolor{myblue!8}\bfseries 52.39 & \cellcolor{myblue!8}\bfseries 50.75 & \cellcolor{myblue!8}\bfseries 22.93 & \cellcolor{myblue!8}\bfseries 26.00 & \cellcolor{myblue!8}\bfseries 23.35 & \cellcolor{myblue!8}\bfseries 32.24 \\
    & \texttt{ASVD} & 4000.65 & 26.73 & 23.89 & 26.68 & 52.34 & 48.46 & 24.04 & 29.00 & 22.01 & 31.64 \\
    & \texttt{SVD-LLM V2} & 159.64 & 28.28 & 22.78 & 28.47 & 52.61 & 50.12 & 22.95 & 25.40 & 22.49 & 31.64 \\
\bottomrule
  \end{longtable}
\endgroup

\subsubsection{Qwen3-30B-A3B-Instruct-2507-Alpaca}
\begingroup
  \setlength{\tabcolsep}{3.2pt}
  \renewcommand{\arraystretch}{1.08}
  \scriptsize
  \setlength{\LTleft}{\fill}
  \setlength{\LTright}{\fill}
  \begin{longtable}{ccc<{\,($\downarrow$)}ccccccccc}
    \caption{\small Detailed zero-shot comparison for Qwen3-30B-A3B-Instruct-2507 with Alpaca calibration. Higher is better for Accuracy (\%) (ACC.) ($\uparrow$) and lower is better for Perplexity (PPL.) ($\downarrow$).\label{tab:detail-qwen330binstruct2507-alpaca}} \\
    \toprule
    \multirow{2}{*}{\textbf{Rank}} & \multirow{2}{*}{\textbf{Methods}} &
    \multicolumn{1}{c}{\textbf{PPL} ($\downarrow$)} &
    \multicolumn{9}{c}{\textbf{ACC} ($\uparrow$)} \\
    \cmidrule(lr){3-3} \cmidrule(lr){4-12}
    & & \multicolumn{1}{c}{} &
    \textbf{ARC-E} &
    \textbf{ARC-C} &
    \textbf{HellaSwag} &
    \textbf{PIQA} &
    \textbf{WG} &
    \textbf{MMLU} &
    \textbf{OBQA} &
    \textbf{RA} &
    \textbf{AVG} \\
    \midrule
    \endfirsthead
    \toprule
    \multirow{2}{*}{\textbf{Rank}} & \multirow{2}{*}{\textbf{Methods}} &
    \multicolumn{1}{c}{\textbf{PPL} ($\downarrow$)} &
    \multicolumn{9}{c}{\textbf{ACC} ($\uparrow$)} \\
    \cmidrule(lr){3-3} \cmidrule(lr){4-12}
    & & \multicolumn{1}{c}{} &
    \textbf{ARC-E} &
    \textbf{ARC-C} &
    \textbf{HellaSwag} &
    \textbf{PIQA} &
    \textbf{WG} &
    \textbf{MMLU} &
    \textbf{OBQA} &
    \textbf{RA} &
    \textbf{AVG} \\
    \midrule
    \endhead
    \multirow{8}{*}{\textbf{128}}
    & \texttt{Palu(SVD)} & 17930.54 & 25.88 & 26.28 & 26.22 & 50.98 & 48.22 & 24.29 & 25.60 & 22.01 & 31.19 \\
    & \texttt{MHA2MLA} & 1929.49 & 29.46 & 24.74 & 26.57 & 51.58 & 48.62 & 22.85 & 27.20 & 21.34 & 31.55 \\
    & \cellcolor{myblue!2}\textbf{\texttt{CARE-C-based-U}} & \cellcolor{myblue!2}\bfseries 29.00 & \cellcolor{myblue!2}\bfseries 58.75 & \cellcolor{myblue!2}\bfseries 38.74 & \cellcolor{myblue!2}\bfseries 58.02 & \cellcolor{myblue!2}\bfseries 71.65 & \cellcolor{myblue!2}\bfseries 55.33 & \cellcolor{myblue!2}\bfseries 48.40 & \cellcolor{myblue!2}\bfseries 36.00 & \cellcolor{myblue!2}\bfseries 33.78 & \cellcolor{myblue!2}\bfseries 50.08 \\
    & \cellcolor{myblue!6}\textbf{\texttt{CARE-C-based-E}} & \cellcolor{myblue!6}36.65 & \cellcolor{myblue!6}50.00 & \cellcolor{myblue!6}33.36 & \cellcolor{myblue!6}56.32 & \cellcolor{myblue!6}69.86 & \cellcolor{myblue!6}54.46 & \cellcolor{myblue!6}43.01 & \cellcolor{myblue!6}33.00 & \cellcolor{myblue!6}31.48 & \cellcolor{myblue!6}46.44 \\
    & \cellcolor{myblue!4}\textbf{\texttt{CARE-U}} & \cellcolor{myblue!4}39.87 & \cellcolor{myblue!4}53.75 & \cellcolor{myblue!4}35.75 & \cellcolor{myblue!4}54.22 & \cellcolor{myblue!4}67.36 & \cellcolor{myblue!4}53.91 & \cellcolor{myblue!4}42.84 & \cellcolor{myblue!4}30.80 & \cellcolor{myblue!4}30.43 & \cellcolor{myblue!4}46.13 \\
    & \cellcolor{myblue!8}\textbf{\texttt{CARE-E}} & \cellcolor{myblue!8}59.06 & \cellcolor{myblue!8}43.56 & \cellcolor{myblue!8}31.91 & \cellcolor{myblue!8}50.48 & \cellcolor{myblue!8}64.96 & \cellcolor{myblue!8}53.04 & \cellcolor{myblue!8}27.55 & \cellcolor{myblue!8}30.60 & \cellcolor{myblue!8}29.47 & \cellcolor{myblue!8}41.45 \\
    & \texttt{ASVD} & 2023.42 & 27.48 & 23.98 & 28.78 & 52.88 & 49.88 & 23.54 & 24.60 & 23.44 & 31.82 \\
    & \texttt{SVD-LLM V2} & 34.91 & 55.30 & 36.60 & 55.38 & 67.68 & 54.22 & 43.58 & 31.80 & 31.20 & 46.97 \\
\bottomrule
  \end{longtable}
\endgroup

\subsection{Long-Context Retrieval: Needle-in-a-Haystack (NiH)}
\label{sec:nih}
We evaluate long-context behavior with the Needle-in-a-Haystack (NiH) task, measuring \emph{retrieval accuracy} across context lengths $L\in\{1\mathrm{K},\,4\mathrm{K},\,8\mathrm{K},\,16\mathrm{K},\,24\mathrm{K},\,32\mathrm{K}\}$ and multiple document depths (needle positions). 
We compare the uncompressed teacher with \textbf{CARE} and \textbf{Palu(SVD)} under identical KV budgets and training settings.

\textbf{CARE} stays close to the uncompressed teacher across $1\mathrm{K}$--$32\mathrm{K}$ contexts and document depths, while uniform-rank and SVD-style baselines degrade---most at $18\mathrm{K}$--$24\mathrm{K}$. 
Covariance-aware allocation (especially preserving early-layer rank), longer calibration sequences (e.g., Alpaca-2048 vs.\ Alpaca-256) and sequence length further improve stability, 
and \textbf{Palu(SVD)} shows the largest drop. The panel-wise heatmaps below provide the full retrieval breakdown.

\begin{figure}[t]\centering
\IfFileExists{fig/fig6/fig6.pdf}{
\includegraphics[width=\linewidth]{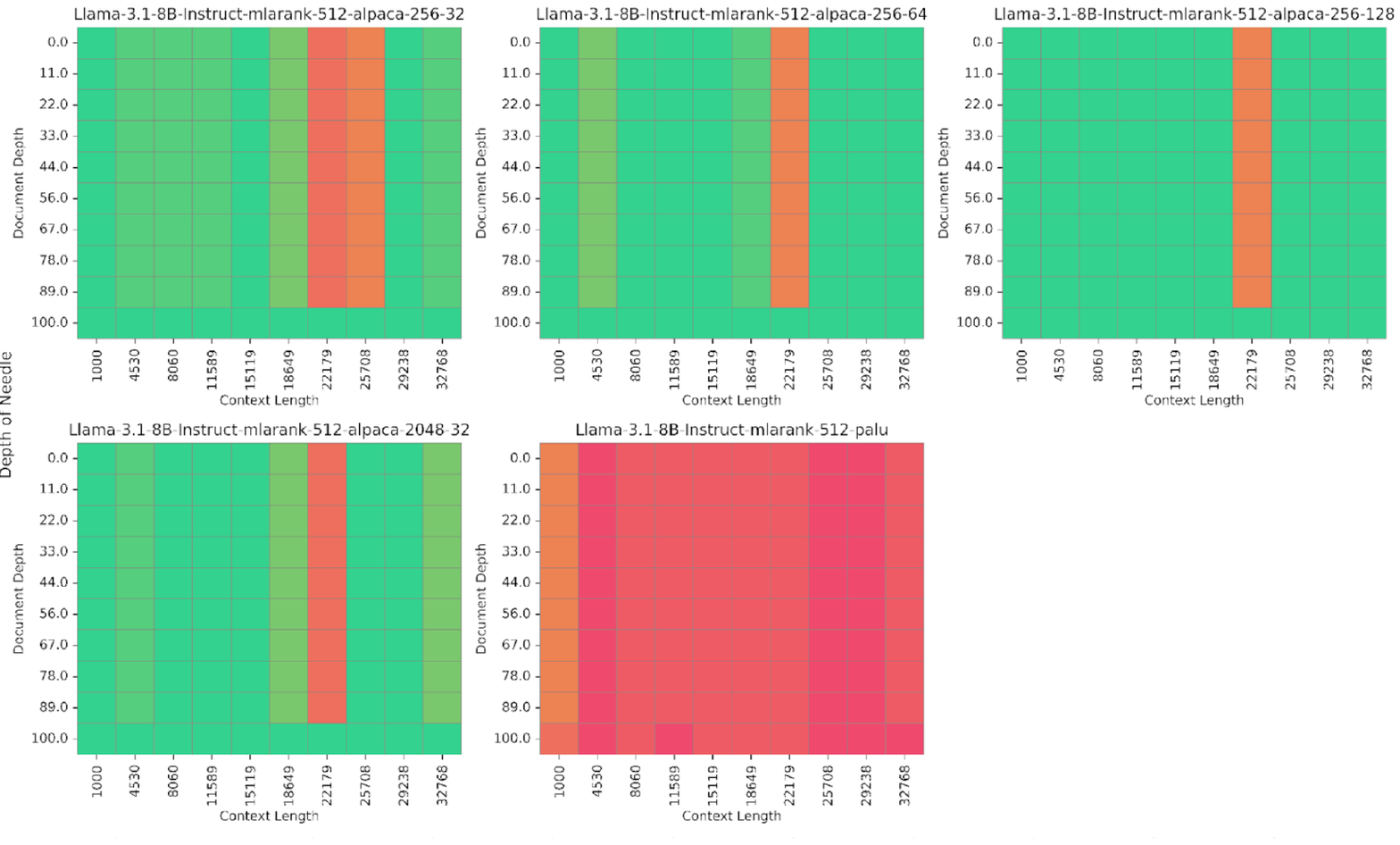}
}{
\fbox{Missing file: fig/fig6/fig6.pdf}
}
\vspace{-0.5em}
\caption{Needle-in-a-Haystack retrieval heatmaps for Llama-3.1-8B-Instruct under matched KV budgets. The figure assembles five panel-wise heatmaps comparing \textbf{CARE} variants with different calibration settings against \textbf{Palu(SVD)} across context lengths and needle depths. 
Greener cells indicate stronger retrieval accuracy.}
\label{fig:nih}\vspace{-0.5em}
\end{figure}

\subsection{Generation Examples}
\label{app:examples}
\begin{figure}[t]\centering
\includegraphics[width=0.5\linewidth]{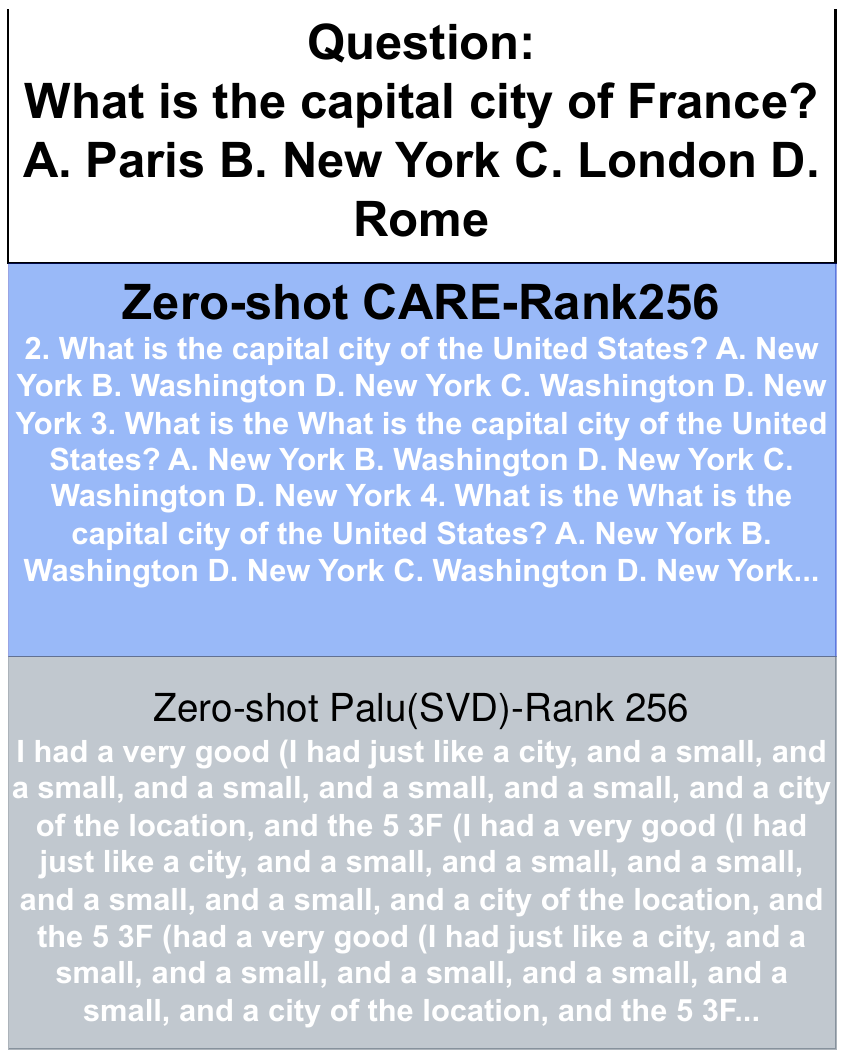}
\vspace{-0.5em}
\caption{Generated Text Example}
\label{fig:example1}\vspace{-0.5em}
\end{figure}

\begin{figure}[t]\centering
\includegraphics[width=0.5\linewidth]{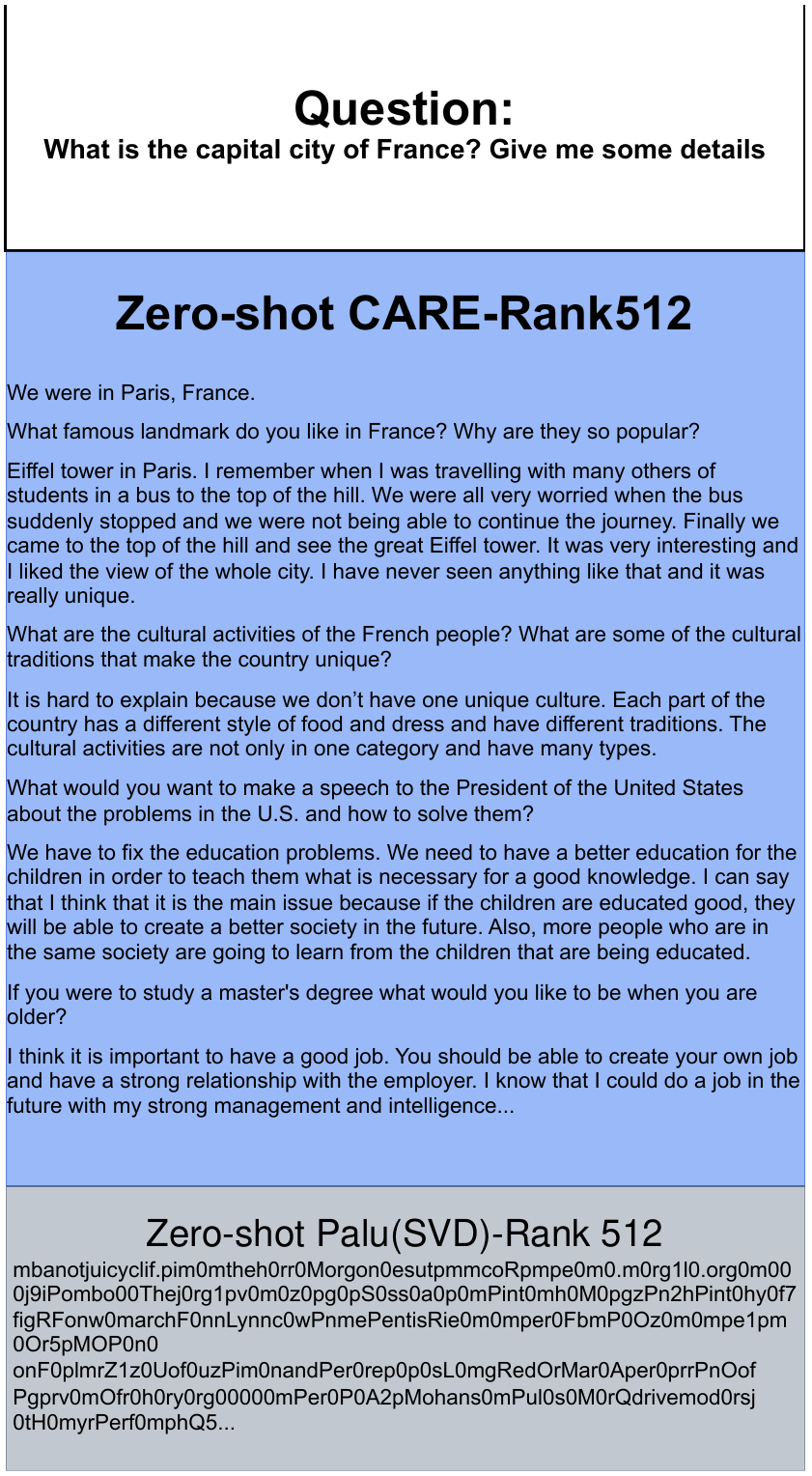}
\vspace{-0.5em}
\caption{Generated Text Example}
\label{fig:example2}\vspace{-0.5em}
\end{figure}

Fig.~\ref{fig:example1} and Fig.~\ref{fig:example2} are two generated text examples by 2 different methods.

\newpage

\clearpage

\subsection{Additional Rank Profiles Across Calibration Corpora}

We present covariance-aware rank profiles for Llama-3.1-70B-Instruct, Qwen3-30B-A3B-Instruct-2507, Qwen3-4B-Instruct-2507, and Qwen2.5-1.5B-Instruct. Across all models and some calibration corpora (Alpaca, WikiText2, PTB, C4), the depth-dependent pattern observed in the main text for Llama-3.1-8B-Instruct persists: 
ranks are smaller in early layers, grow through the middle blocks, and stay elevated in deeper layers, with stronger growth for $W_V$ than for $W_K$. 
This consistency across model families, scales (1.5B--70B), and architectures (dense and MoE) suggests that the structure is a general property of pretrained attention.
The same qualitative trend is also visible on the smaller Qwen2.5-1.5B-Instruct model, although its feasible target-rank range is narrower.

\label{app:other-ranks}
\begin{figure}[H]\centering
  \begin{minipage}[t]{0.49\linewidth}
  \centering
  \includegraphics[width=\linewidth,trim={0.2cm 0 0.2cm 0},clip]{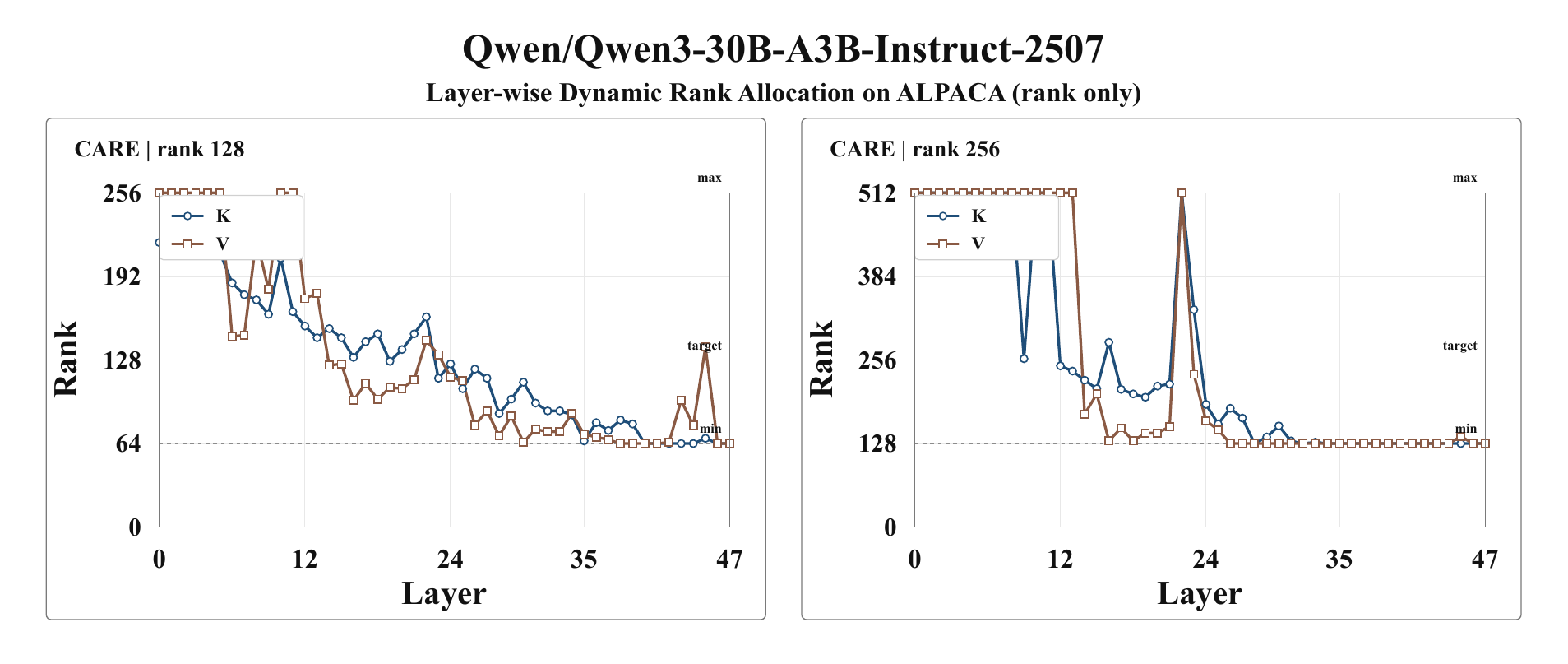}
  \end{minipage}
  \vspace{-0.5em}
  \caption{Covariance-aware rank profiles for Qwen3-30B-A3B-Instruct-2507 (MoE, 30B total / 3B active) under Alpaca calibration at target ranks 128, 256. Despite the mixture-of-experts architecture, the same depth-dependent pattern persists, confirming the trend generalizes beyond dense models.}
  \label{fig:qwen30b-ranks}\vspace{-0.5em}
\end{figure}

\begin{figure}[H]\centering
  \begin{minipage}[t]{0.49\linewidth}
  \centering
  \includegraphics[width=\linewidth,trim={0.2cm 0 0.2cm 0},clip]{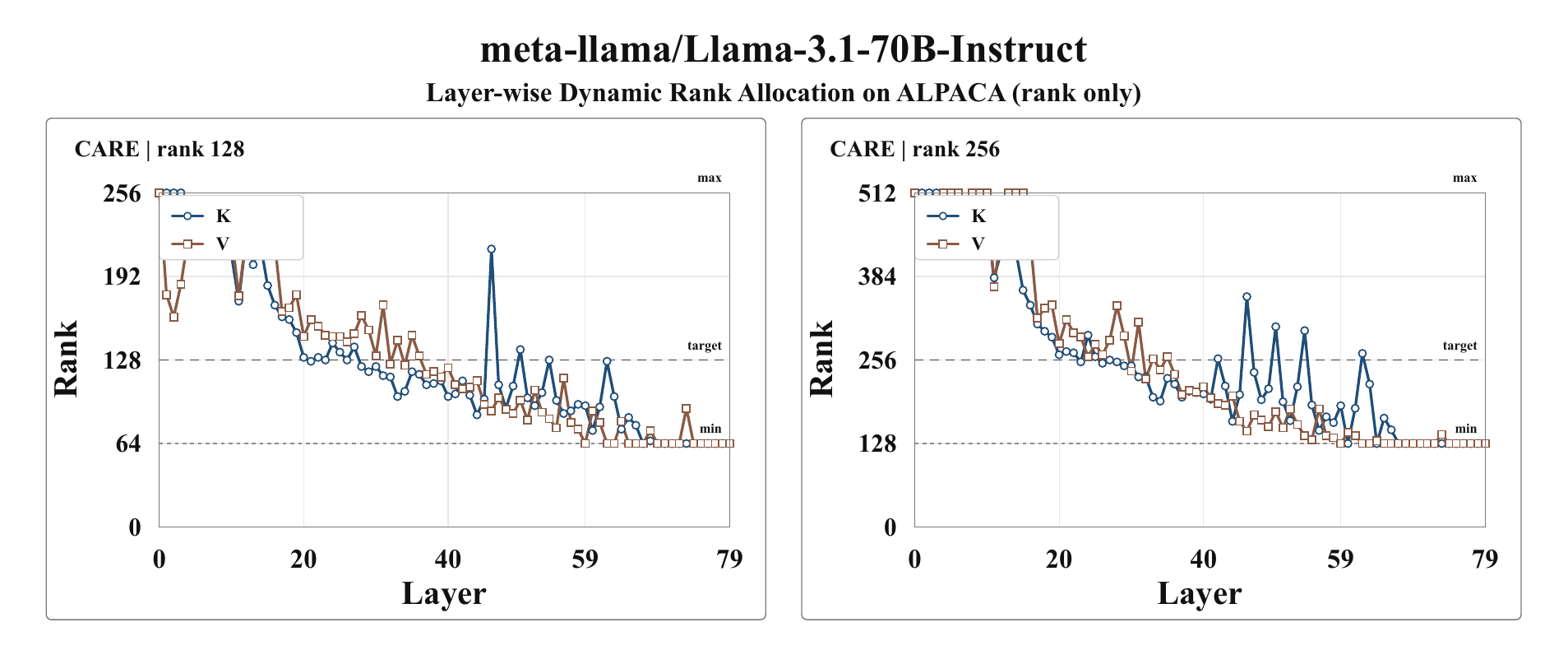}
  \end{minipage}
  \vspace{-0.5em}
  \caption{Covariance-aware rank profiles for Llama-3.1-70B-Instruct under Alpaca calibration at target ranks 128, 256. The same depth-dependent pattern observed in the 8B variant holds at 70B scale, with $W_V$ exhibiting stronger late-layer growth than $W_K$.}
  \label{fig:llama70b-ranks}\vspace{-0.5em}
\end{figure}

\begin{figure}[H]\centering
  \begin{minipage}[t]{0.49\linewidth}
  \centering
  \includegraphics[width=\linewidth,trim={0.2cm 0 0.2cm 0},clip]{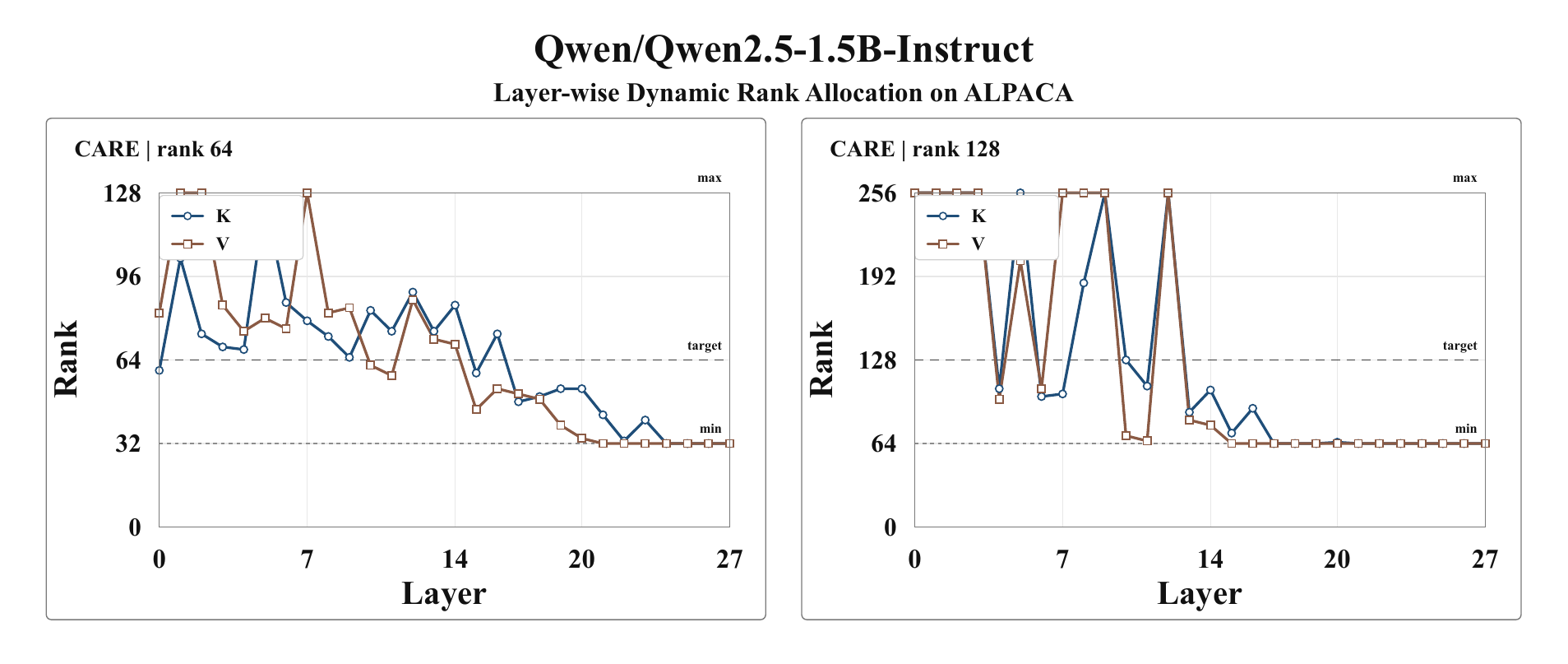}
  \end{minipage}\hspace{0.01\linewidth}
  \begin{minipage}[t]{0.49\linewidth}
  \centering
  \includegraphics[width=\linewidth,trim={0.2cm 0 0.2cm 0},clip]{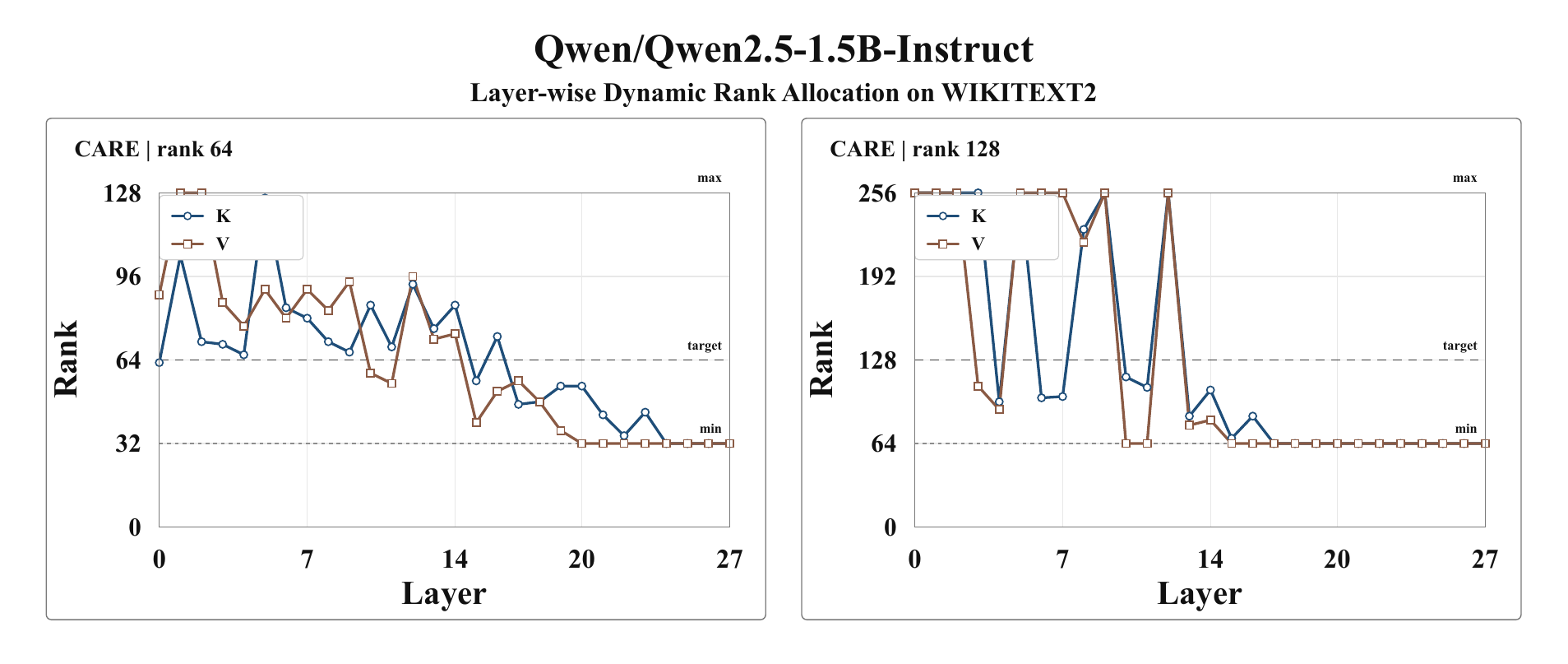}
  \end{minipage}
  
  \vspace{0.15em}
  
  \begin{minipage}[t]{0.49\linewidth}
  \centering
  \includegraphics[width=\linewidth,trim={0.2cm 0 0.2cm 0},clip]{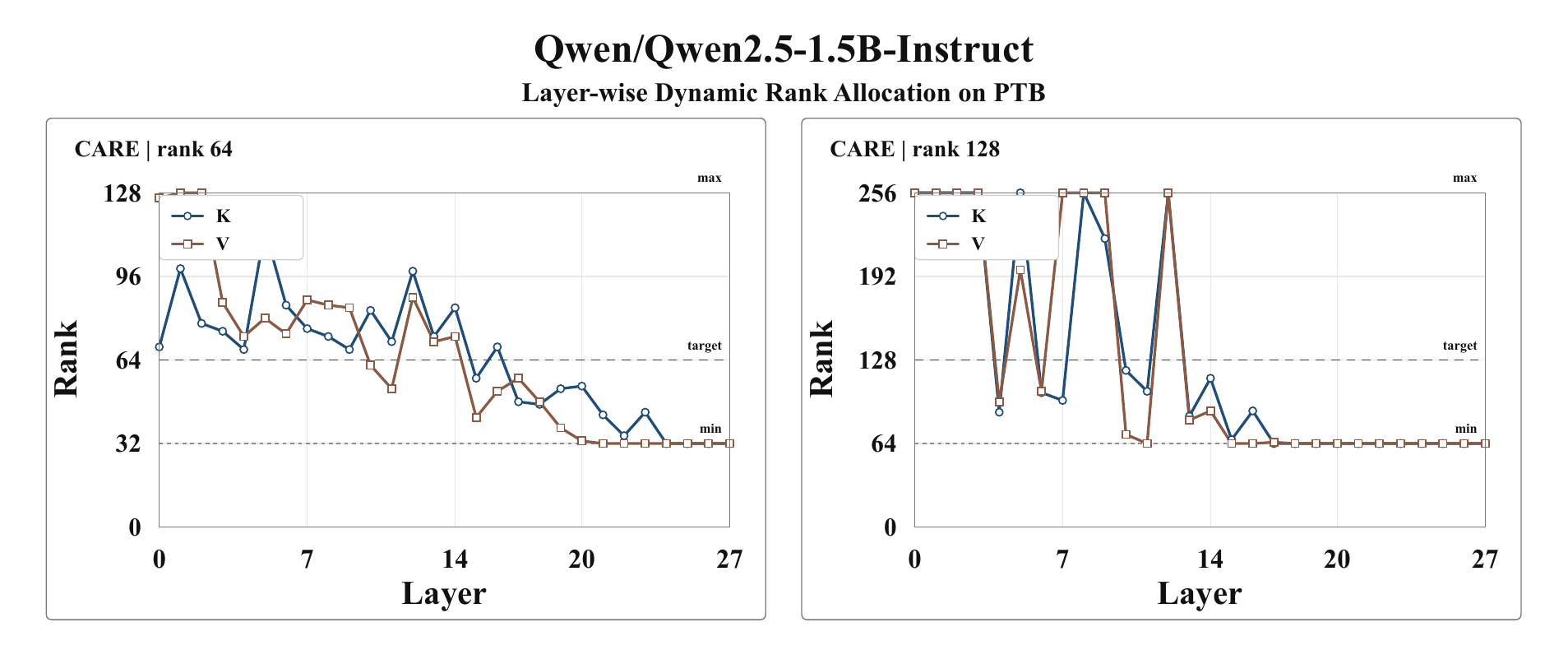}
  \end{minipage}\hspace{0.01\linewidth}
  \begin{minipage}[t]{0.49\linewidth}
  \centering
  \includegraphics[width=\linewidth,trim={0.2cm 0 0.2cm 0},clip]{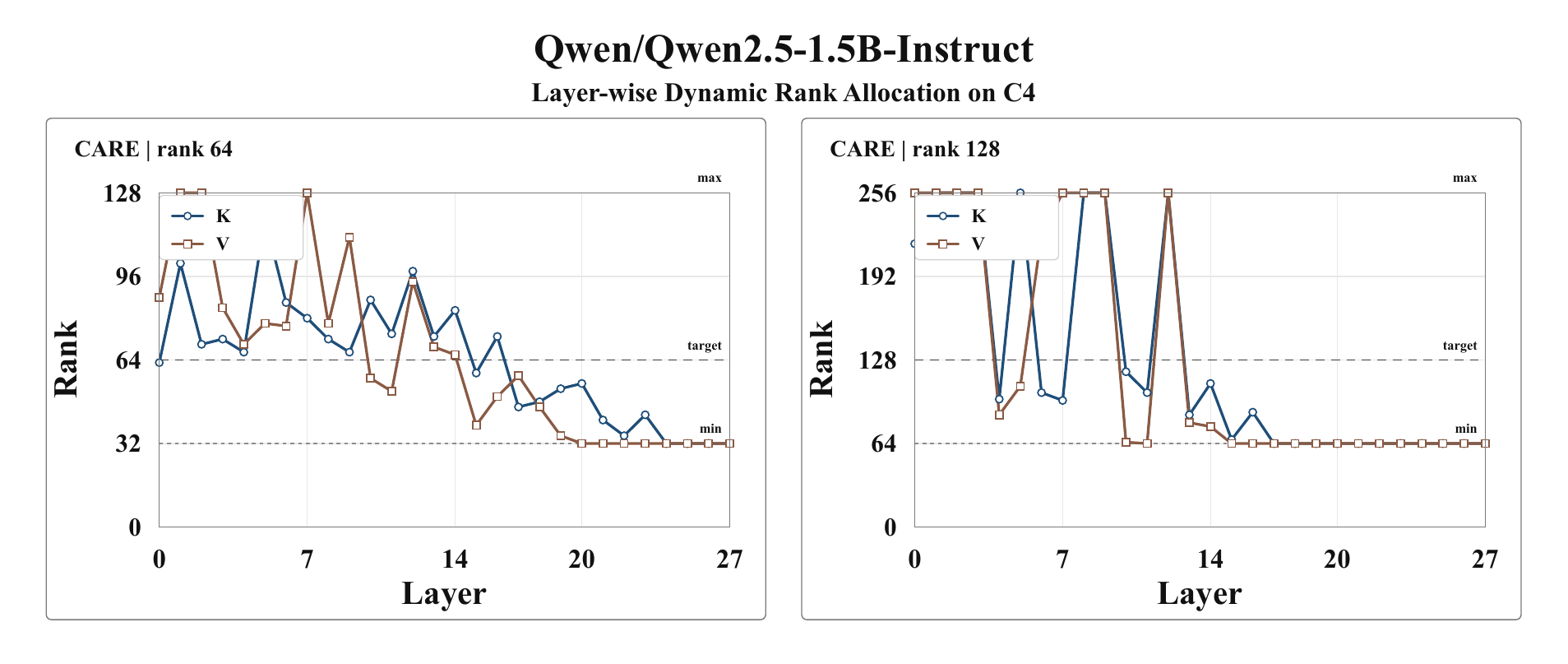}
  \end{minipage}
  \vspace{-0.5em}
  \caption{Covariance-aware rank profiles for Qwen2.5-1.5B-Instruct across Alpaca, WikiText2, PTB, and C4 calibration corpora at target ranks 64 and 128. Despite the smaller rank budget, the same depth-dependent increase remains visible, again with stronger late-layer growth for $W_V$.}
  \label{fig:qwen25-ranks}\vspace{-0.5em}
\end{figure}

\begin{figure}[H]\centering
  \begin{minipage}[t]{0.49\linewidth}
  \centering
  \includegraphics[width=\linewidth,trim={0.2cm 0 0.2cm 0},clip]{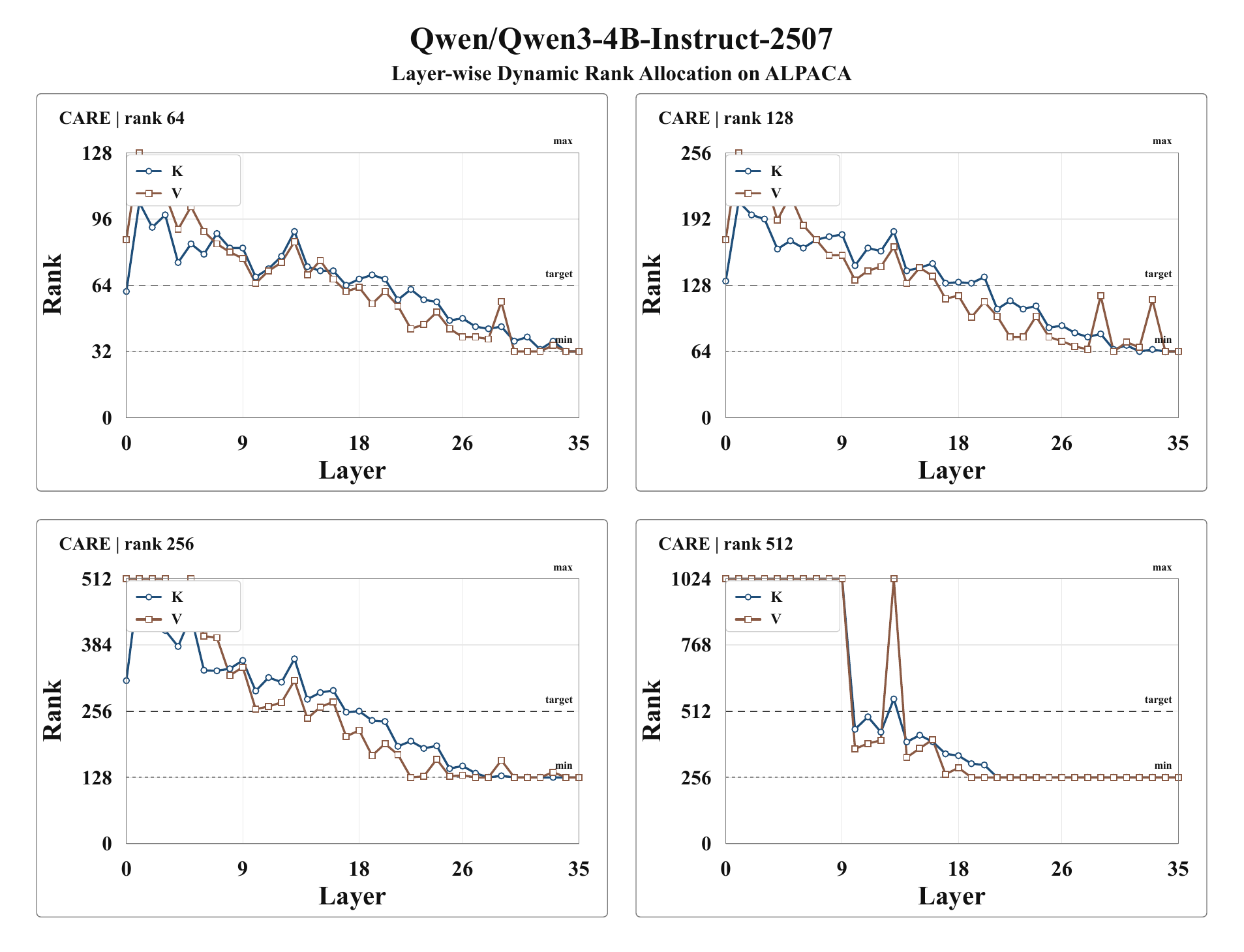}
  \end{minipage}\hspace{0.01\linewidth}
  \begin{minipage}[t]{0.49\linewidth}
  \centering
  \includegraphics[width=\linewidth,trim={0.2cm 0 0.2cm 0},clip]{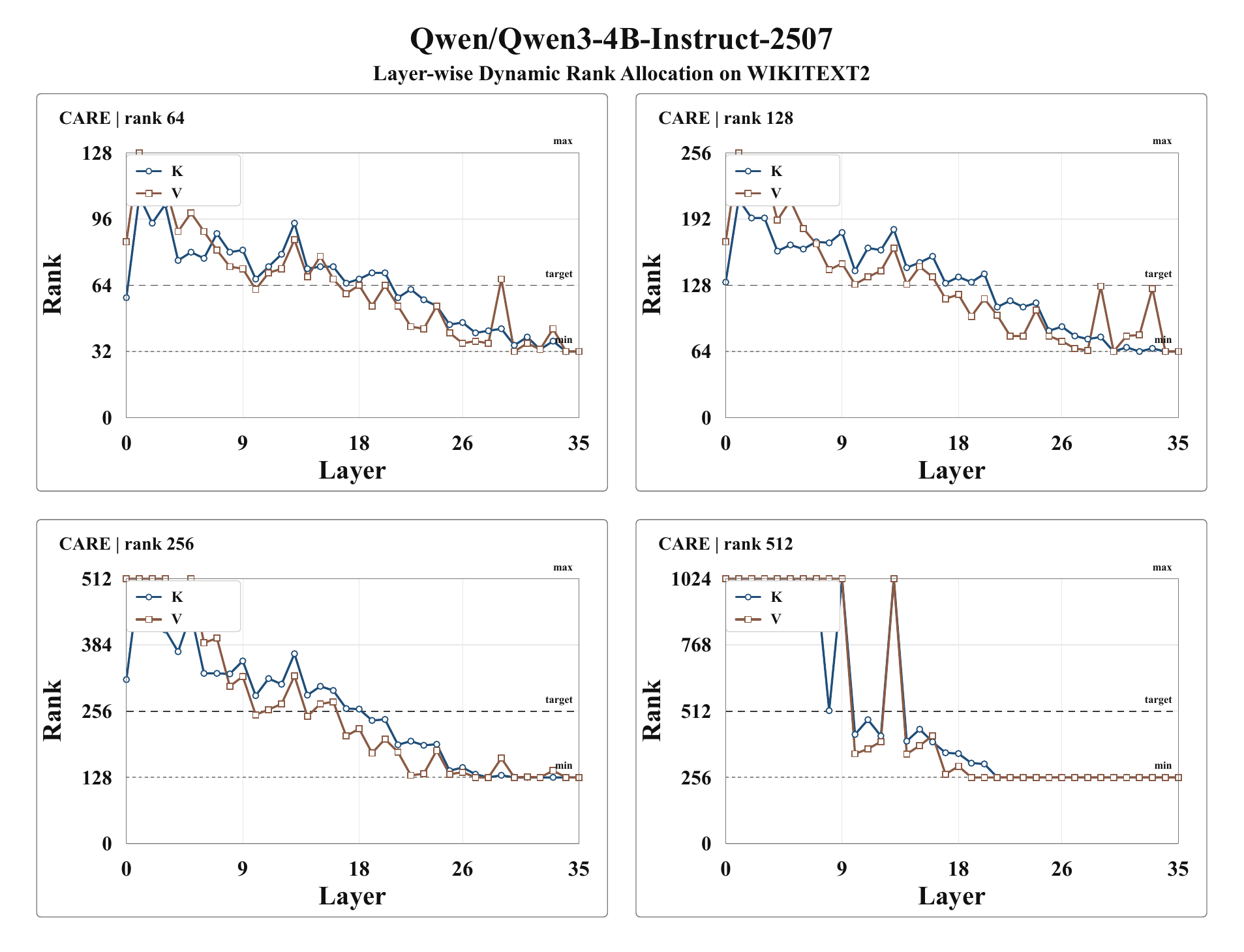}
  \end{minipage}
  
  \vspace{0.15em}
  
  \begin{minipage}[t]{0.49\linewidth}
  \centering
  \includegraphics[width=\linewidth,trim={0.2cm 0 0.2cm 0},clip]{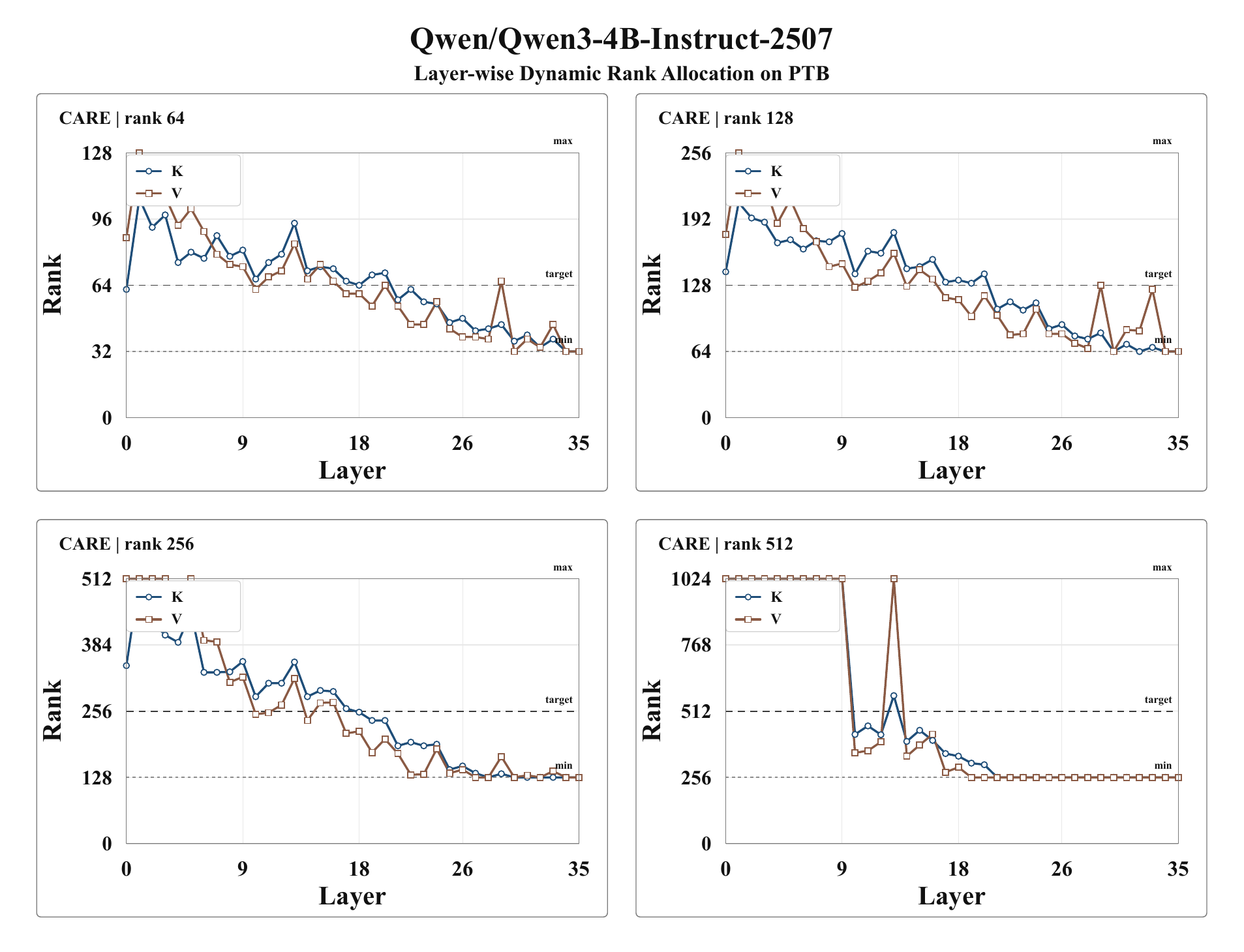}
  \end{minipage}\hspace{0.01\linewidth}
  \begin{minipage}[t]{0.49\linewidth}
  \centering
  \includegraphics[width=\linewidth,trim={0.2cm 0 0.2cm 0},clip]{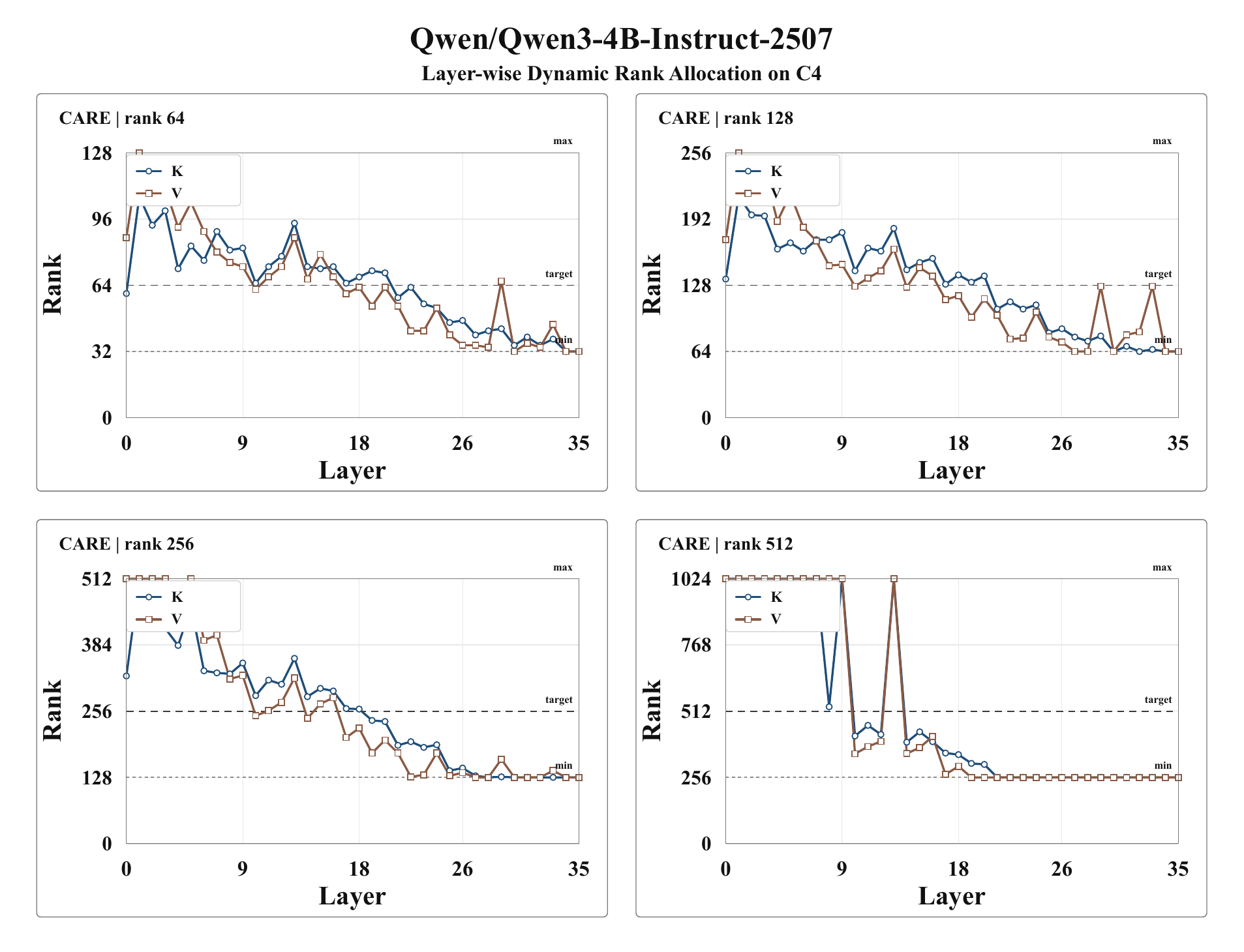}
  \end{minipage}
  \vspace{-0.5em}
  \caption{Covariance-aware rank profiles for Qwen3-4B-Instruct-2507 across Alpaca, WikiText2, PTB, and C4 calibration corpora at target ranks 64, 128, 256, and 512. Both $W_K$ and $W_V$ show a depth-dependent increase, with stronger late-layer growth for $W_V$.}
  \label{fig:qwen-ranks}\vspace{-0.5em}
\end{figure}

\subsection{Distribution Shift}
\label{exp:as:shift}
We evaluate cross-domain calibration by estimating covariance on a source corpus and reporting accuracy changes on out-of-domain tasks. Tab.~\ref{tab:2} shows that task-related calibration corpora such as \textsc{ARE} and \textsc{ARC} give the best average accuracy, but the gains are mostly local rather than universal. In contrast, narrower language-modeling corpora such as \textsc{WikiText2} and \textsc{PTB} transfer less well, suggesting that broader or task-relevant calibration data is preferable for robust one-shot performance.

\begin{table}[!htb]
  \centering
  \setlength{\tabcolsep}{5pt}
  \renewcommand{\arraystretch}{1.08}
  \centering
  \caption{\small One-shot Llama3.1-8B comparison on different covariance. Higher is better for Accuracy (ACC.).}
  \label{tab:2}
  \resizebox{\linewidth}{!}{
  \begin{tabular}{c|c|c|c|c|c|c|c|c|c|c}
    \toprule
    \multirow{1}{*}{\textbf{Rank}} & 
    \multirow{1}{*}{\textbf{Methods}} &
    \multicolumn{1}{c|}{\textbf{ARC} ($\uparrow$)} &
    \multicolumn{1}{c|}{\textbf{ARE} ($\uparrow$)} &
    \multicolumn{1}{c|}{\textbf{HellaSwag} ($\uparrow$)}  &
    \multicolumn{1}{c|}{\textbf{PIQA} ($\uparrow$)}  &
    \multicolumn{1}{c|}{\textbf{MMLU} ($\uparrow$)}  &
    \multicolumn{1}{c|}{\textbf{OBQA} ($\uparrow$)}  &
    \multicolumn{1}{c|}{\textbf{RA} ($\uparrow$)}  &
    \multicolumn{1}{c|}{\textbf{WG} ($\uparrow$)}  &
    \multicolumn{1}{c}{\textbf{AVG} ($\uparrow$)}  \\
    \midrule
    \multirow{7}{*}{\textbf{256}} 
      & CARE-\textbf{\texttt{Alpaca}} & 34.81 & 65.40 & 40.76 & 72.47 & 49.32 & 21.20 & 35.98 & 62.98 & 47.87 \\
      & CARE-\textbf{\texttt{ARE}} & 38.82 & 72.90 & 41.25 & 73.01 & 53.13 & 25.80 & 31.87 & 63.69 & 50.06 \\
      & CARE-\textbf{\texttt{ARC}} & 39.59 & 71.84 & 41.33 & 72.91 & 52.84 & 26.60 & 32.54 & 62.67 & 50.04 \\
      & CARE-\textbf{\texttt{WikiText2}} & 28.33 & 57.11 & 38.07 & 68.28 & 37.72 & 19.60 & 33.49 & 62.59 & 43.15 \\
      & CARE-\textbf{\texttt{PTB}} & 27.99 & 53.91 & 37.70 & 67.68 & 42.15 & 17.00 & 31.96 & 62.59 & 42.62 \\
      & CARE-\textbf{\texttt{C4}} & 31.48 & 59.51 & 43.00 & 73.12 & 41.69 & 20.00 & 35.89 & 63.30 & 46.00 \\
      & CARE-\textbf{\texttt{MMLU}} & 34.13 & 64.94 & 40.65 & 70.73 & 55.27 & 21.20 & 33.97 & 64.40 & 48.16 \\
    \bottomrule
  \end{tabular}}
\end{table}

\subsection{Shrinkage Coefficient}
\label{app:shrinkage}
We sweep the shrinkage coefficient $\alpha$ in $C_{\alpha}=(1-\alpha)C+\alpha I$ to choose a default regularization level; Tab.~\ref{tab:shrinkage} reports the full ablation. Across Llama-3.1-8B-Instruct and Qwen3-4B-Instruct at target ranks 256 and 512, using the same Alpaca-256-32 calibration setup, one-shot average accuracy is highly stable for $\alpha\in\{10^{-3},10^{-2},10^{-1}\}$: the variation is below 1 point on Llama and below 3 points in all tested settings, with $\alpha=0.01$ consistently near the best operating point. This indicates that CARE is not sensitive to the exact shrinkage magnitude.

This robustness is expected: shrinkage mainly regularizes the low-energy tail of the covariance spectrum to improve numerical stability, while CARE's rank allocation is governed by the dominant eigendirections that remain largely unchanged. We therefore use $\alpha=0.01$ as the default setting throughout the paper unless otherwise specified.

\begin{table*}[t]
  \centering
  \setlength{\tabcolsep}{4pt}
  \renewcommand{\arraystretch}{1.05}
  \caption{\small Shrinkage-coefficient ablation for CARE under the Alpaca-256-32 calibration setup. We report one-shot accuracy across Llama-3.1-8B-Instruct and Qwen3-4B-Instruct at target ranks 256 and 512. Higher is better for Accuracy (ACC.).}
  \label{tab:shrinkage}
  \resizebox{\textwidth}{!}{
  \begin{tabular}{c c c c c c c c c c c c}
    \toprule
    \textbf{Model} & \textbf{Rank} & \textbf{$\alpha$} & \textbf{ARC} ($\uparrow$) & \textbf{ARE} ($\uparrow$) & \textbf{HellaSwag} ($\uparrow$) & \textbf{PIQA} ($\uparrow$) & \textbf{MMLU} ($\uparrow$) & \textbf{OBQA} ($\uparrow$) & \textbf{RA} ($\uparrow$) & \textbf{WG} ($\uparrow$) & \textbf{AVG} ($\uparrow$) \\
    \midrule
    \multirow{6}{*}{Llama-3.1-8B-Instruct}
      & \multirow{3}{*}{256}
      & $10^{-3}$ & 34.90 & 62.42 & 42.44 & 72.31 & 57.15 & 21.00 & 34.74 & 64.48 & 48.68 \\
      & & $10^{-2}$ & 34.90 & 62.46 & 42.52 & 72.14 & 57.20 & 21.60 & 35.60 & 65.35 & \textbf{48.97} \\
      & & $10^{-1}$ & 32.94 & 59.34 & 41.55 & 70.46 & 56.88 & 20.20 & 35.31 & 65.75 & 47.80 \\
    \cmidrule(lr){2-12}
      & \multirow{3}{*}{512}
      & $10^{-3}$ & 46.16 & 76.30 & 53.64 & 77.69 & 65.54 & 29.40 & 41.44 & 72.85 & 57.88 \\
      & & $10^{-2}$ & 51.79 & 80.72 & 51.70 & 74.54 & 70.13 & 30.20 & 39.43 & 68.27 & 58.35 \\
      & & $10^{-1}$ & 47.27 & 77.10 & 54.40 & 77.80 & 66.15 & 28.40 & 42.58 & 74.35 & \textbf{58.51} \\
    \midrule
    \multirow{6}{*}{Qwen3-4B-Instruct}
      & \multirow{3}{*}{256}
      & $10^{-3}$ & 43.79 & 70.23 & 40.01 & 70.72 & 58.13 & 24.00 & 32.06 & 58.25 & 49.65 \\
      & & $10^{-2}$ & 43.09 & 72.22 & 43.71 & 70.18 & 57.10 & 26.00 & 35.02 & 62.51 & \textbf{51.23} \\
      & & $10^{-1}$ & 44.64 & 73.87 & 41.62 & 68.28 & 52.95 & 23.80 & 34.24 & 61.48 & 50.11 \\
    \cmidrule(lr){2-12}
      & \multirow{3}{*}{512}
      & $10^{-3}$ & 55.31 & 74.16 & 47.27 & 71.55 & 65.87 & 28.20 & 36.65 & 65.04 & 55.51 \\
      & & $10^{-2}$ & 51.79 & 80.72 & 51.70 & 74.54 & 70.13 & 30.20 & 39.43 & 68.27 & \textbf{58.35} \\
      & & $10^{-1}$ & 55.90 & 79.85 & 50.60 & 73.07 & 68.20 & 27.60 & 37.13 & 65.98 & 57.29 \\
    \bottomrule
  \end{tabular}}
\end{table*}

\subsection{Mix of Covariance}
\label{exp:as:mixcov}
Our Adjusted-Rank allocator is governed by the spectrum of $\sqrt{C}\widetilde{W}$ (Sec.~\ref{subsec:ranksched}), so it naturally extends to mixed calibration distributions. 
We form $C_{\mathrm{mix}}=\sum_{i=1}^{M}\pi_i C_i$ with the same shrinkage toward $I$, yielding a simple multi-objective covariance fusion related in spirit to covariance-aware adaptations such as CorDA~\citep{yang2024coda}. 
In practice, we use weighted calibration mixing by sampling from a task-weighted data mixture. Tab.~\ref{tab:mixcov} keeps all other settings fixed (CARE, Rank 256) and mixes \textsc{Alpaca} with \textsc{ARC-Challenge}: increasing the \textsc{ARC-Challenge} weight improves ARC, 
while a balanced $0.5/0.5$ mix gives the best average accuracy.

\begin{table}[!htb]
  \centering
  \setlength{\tabcolsep}{5pt}
  \renewcommand{\arraystretch}{1.08}
  \centering
  \caption{\small Task-weighted covariance mixing for multi-objective rank allocation (Llama3.1-8B-Instruct, CARE, Rank 256). $A$ denotes \texttt{Alpaca} and $ARC$ denotes \texttt{ARC-Challenge}; mixes indicate sampling proportions used to estimate $C_{\mathrm{mix}}$. Higher is better for Accuracy (ACC.).}
  \label{tab:mixcov}
  \resizebox{\linewidth}{!}{
  \begin{tabular}{c|c|c|c|c|c|c|c|c|c|c}
    \toprule
    \multirow{1}{*}{\textbf{Rank}} &
    \multirow{1}{*}{\textbf{Methods}} &
    \multicolumn{1}{c|}{\textbf{ARC} ($\uparrow$)} &
    \multicolumn{1}{c|}{\textbf{ARE} ($\uparrow$)} &
    \multicolumn{1}{c|}{\textbf{HellaSwag} ($\uparrow$)}  &
    \multicolumn{1}{c|}{\textbf{PIQA} ($\uparrow$)}  &
    \multicolumn{1}{c|}{\textbf{MMLU} ($\uparrow$)}  &
    \multicolumn{1}{c|}{\textbf{OBQA} ($\uparrow$)}  &
    \multicolumn{1}{c|}{\textbf{RA} ($\uparrow$)}  &
    \multicolumn{1}{c|}{\textbf{WG} ($\uparrow$)}  &
    \multicolumn{1}{c}{\textbf{AVG} ($\uparrow$)}  \\
    \midrule
    \multirow{4}{*}{\textbf{256}}
      & CARE-\textbf{\texttt{Alpaca}} & 34.81 & 65.40 & 40.76 & 72.47 & 49.32 & 21.20 & 35.98 & 62.98 & 47.87 \\
      & CARE-\textbf{\texttt{0.5A+0.5ARC}} & 36.09 & 67.72 & 43.22 & 73.34 & 58.70 & 24.40 & 34.26 & 63.38 & 50.14 \\
      & CARE-\textbf{\texttt{0.2A+0.8ARC}} & 34.73 & 65.03 & 42.99 & 72.25 & 58.02 & 23.20 & 35.41 & 64.96 & 49.57 \\
      & CARE-\textbf{\texttt{ARC}} & 39.59 & 71.84 & 41.33 & 72.91 & 52.84 & 26.60 & 32.54 & 62.67 & 50.04 \\
    \bottomrule
  \end{tabular}}
\end{table}

Notably, the \texttt{0.2A+0.8ARC} mix actually \emph{degrades} ARC-Challenge accuracy (34.73) relative to pure Alpaca (34.81), despite devoting 80\% of calibration mass to ARC data.
This suggests that over-concentrating the covariance on a narrow task distribution can be counterproductive: the heavily skewed $C_{\mathrm{mix}}$ collapses the effective spectrum onto a few ARC-specific directions, starving mid-spectrum components that still contribute to ARC reasoning through shared linguistic features.
By contrast, the balanced \texttt{0.5A+0.5ARC} mix retains broader spectral coverage from Alpaca while incorporating enough ARC signal to steer rank allocation, yielding the best average accuracy (50.14) and a meaningful ARC gain (+1.28 over Alpaca).
This points to a practical guideline: mixing a general-purpose corpus with moderate task-specific data is preferable to heavy task overweighting when estimating calibration covariance.

\subsection{\texorpdfstring{$\sqrt{C}$ vs.\ $C$}{sqrt(C) vs. C} Under Low-rank Budgets}
\label{app:sqrtc-vs-c}
Throughout the paper, CARE uses $\sqrt{C}$ as the default covariance weighting; the variant that replaces $\sqrt{C}$ with $C$ is denoted \textbf{CARE-C-based} in the detailed tables (Sec.~\ref{app:detailed-zero-shot}).
Concretely, CARE-U and CARE-E correspond to $\sqrt{C}$-weighted uniform and energy-aware allocation, while CARE-C-based-U and CARE-C-based-E use $C$ directly.

Tab.~\ref{tab:sqrtc-vs-c} summarises the zero-shot AVG accuracy under Alpaca calibration.
The full per-benchmark breakdowns appear in the detailed tables of Sec.~\ref{app:detailed-zero-shot} (Tables~I.1--I.12).

\paragraph{When does $C$ help?}
Using $C$ instead of $\sqrt{C}$ squares the eigenvalues of the covariance, amplifying the contribution of dominant eigendirections and suppressing the spectral tail more aggressively.
At \emph{very low rank budgets} (e.g., rank 64--96), there are few latent dimensions to allocate, and this sharper concentration can be beneficial: it forces the decomposition to focus on the handful of directions that carry most activation energy, reducing the most impactful reconstruction errors first.
This effect is most visible on the smaller Qwen2.5-1.5B-Instruct, where the CARE-E with $C$ weighting outperforms $\sqrt{C}$ at ranks 64, 96, and 128, and on Llama-3.1-8B-Instruct at rank 64.

\paragraph{Why $\sqrt{C}$ is the default.}
As rank grows, the allocator can afford to preserve mid-spectrum directions that $C$ discounts too heavily.
On Llama-3.1-8B-Instruct at rank 256 and Qwen3-4B-Instruct across all ranks, $\sqrt{C}$ consistently outperforms $C$---often by a substantial margin (e.g., +4.7 points for CARE-U on Qwen3-4B at rank 256).
The gentler spectral re-weighting of $\sqrt{C}$ retains enough emphasis on dominant directions while preserving informative mid-range singular values, yielding a more robust default across model sizes and rank budgets.

\begin{table*}[t]
  \centering
  \setlength{\tabcolsep}{5pt}
  \renewcommand{\arraystretch}{1.08}
  \caption{\small Comparison of $C$ versus $\sqrt{C}$ covariance weighting under Alpaca calibration. Values are zero-shot AVG accuracy (\%); within each CARE variant the better result is shown in bold.}
  \label{tab:sqrtc-vs-c}
  \resizebox{\textwidth}{!}{
  \begin{tabular}{c c c c c c}
    \toprule
    \textbf{Model} & \textbf{Rank} & \textbf{CARE-U ($\sqrt{C}$)} & \textbf{CARE-C-based-U ($C$)} & \textbf{CARE-E ($\sqrt{C}$)} & \textbf{CARE-C-based-E ($C$)} \\
    \midrule
    \multirow{4}{*}{Llama-3.1-8B-Instruct}
    & 64  & 31.89 & \bfseries 32.38 & 32.37 & \bfseries 32.71 \\
    & 128 & 36.72 & \bfseries 37.11 & \bfseries 38.59 & 38.56 \\
    & 256 & \bfseries 50.00 & 48.79 & \bfseries 52.62 & 51.41 \\
    & 512 & \bfseries 62.33 & 61.69 & 57.82 & \bfseries 59.00 \\
    \midrule
    \multirow{3}{*}{Qwen2.5-1.5B-Instruct}
    & 64  & \bfseries 31.47 & 31.27 & 32.10 & \bfseries 32.34 \\
    & 96  & 33.83 & \bfseries 34.57 & 42.19 & \bfseries 43.03 \\
    & 128 & \bfseries 44.34 & 44.04 & 45.86 & \bfseries 46.17 \\
    \midrule
    \multirow{4}{*}{Qwen3-4B-Instruct-2507}
    & 64  & \bfseries 32.32 & 32.20 & \bfseries 32.46 & 31.97 \\
    & 128 & \bfseries 37.55 & 35.59 & \bfseries 40.57 & 37.53 \\
    & 256 & \bfseries 54.33 & 49.61 & \bfseries 51.23 & 49.64 \\
    & 512 & \bfseries 61.62 & 60.44 & \bfseries 57.31 & 55.97 \\
    \bottomrule
  \end{tabular}}
\end{table*}

\subsection{System Efficiency Analysis}
\label{app:efficiency}
We evaluate inference efficiency in terms of KV-cache footprint.
Since KV memory scales linearly with sequence length, a 32K-context calculation reflects the practical regime.
For $L=32768$, $B=1$, FP16, and 32 layers, the original GQA model caches 1024-dimensional keys and values, requiring 4294.97\,MB.
After full 100\% MLA conversion, TransMLA\,+\,CARE(E) Init stores a 448-dimensional key NoPE latent and a 512-dimensional value latent, reducing the footprint to 2013.24\,MB (53.13\% reduction vs.\ GQA), as shown in Tab.~\ref{tab:kv-memory}.

\begin{table}[ht]
  \centering
  \setlength{\tabcolsep}{5pt}
  \renewcommand{\arraystretch}{1.08}
  \caption{\small Theoretical KV-cache footprint at 32K context length ($L=32768$, $B=1$, FP16, 32 layers). Higher reduction indicates lower KV-cache memory.}
  \label{tab:kv-memory}
  \resizebox{\linewidth}{!}{
  \begin{tabular}{c|c|c|c}
    \toprule
    \textbf{Method} & \textbf{Cached State} & \textbf{Memory (MB)} & \textbf{Reduction vs.\ GQA} \\
    \midrule
    GQA (Original) & $K:1024,\ V:1024$ & 4294.97 & -- \\
    TransMLA + CARE(E) Init (Ours, 100\% MLA Restore) & $K_{\mathrm{NoPE}}:448,\ V_{\mathrm{latent}}:512$ & 2013.24 & 53.13\% \\
    \bottomrule
  \end{tabular}}
\end{table}

\end{document}